\documentclass{article}
\PassOptionsToPackage{numbers, compress}{natbib}
\usepackage[main,final]{neurips_2026}

\usepackage{times}

\usepackage{nicefrac}       %
\usepackage{microtype}      %

\usepackage[utf8]{inputenc}
\usepackage[T1]{fontenc}
\usepackage{microtype}
\usepackage{graphicx}
\usepackage{subfigure}
\usepackage{booktabs}
\usepackage{amsmath}
\usepackage[showonlyrefs]{mathtools}
\usepackage[colorlinks=true,citecolor=blue,linkcolor=blue]{hyperref}

\usepackage{url}
\usepackage{nicefrac}
\usepackage{listings}
\usepackage{mathtools}
\usepackage{amsfonts}
\usepackage{amssymb}
\usepackage{amsmath}
\usepackage{amsthm}
\usepackage{docmute}
\usepackage[acronym, nomain]{glossaries}
\usepackage{indentfirst}
\usepackage{bm}
\usepackage{here}
\usepackage{booktabs}       %
\usepackage{cancel}
\usepackage{multirow}
\usepackage{wrapfig}
\usepackage{placeins}
\usepackage{needspace}
\usepackage{lipsum}
\usepackage{dsfont}
\usepackage{thm-restate}

\newtheorem{lemma}{Lemma}
\newtheorem{theorem}{Theorem}
\newtheorem{proposition}{Proposition}
\newtheorem{corollary}{Corollary}
\newtheorem{assumption}{Assumption}

\newtheorem{remark}{Remark}

\DeclarePairedDelimiterX{\KL}[2]{\mathrm{KL}[}{]}{#1\;\delimsize\|\;#2}

\DeclarePairedDelimiterX\braket[2]{\langle}{\rangle}{#1 \delimsize\vert #2}

\newcommand{\calx}{\mathcal{X}}

\newcommand{\Ex}{\mathbb{E}}
\newcommand{\calo}{\mathcal{O}}

\newcommand{\calz}{\mathcal{Z}}

\newcommand{\caly}{\mathcal{Y}}
\newcommand{\cald}{\mathcal{D}}

\newcommand{\calf}{\mathcal{F}}

\usepackage{multirow}

\usepackage{xcolor}
\usepackage{mathrsfs}
\usepackage{enumitem}

\newcommand{\R}{\mathbb{R}}

\newcommand{\cC}{\mathcal{C}}

\newcommand{\cG}{\mathcal{G}}
\newcommand{\cH}{\mathcal{H}}

\newcommand{\calv}{\mathcal{V}}

\newcommand{\Law}{\mathrm{Law}}
\newcommand{\CVaR}{\mathrm{CVaR}}
\newcommand{\DRO}{\mathrm{DRO}}

\newcommand{\OCE}{\mathrm{OCE}}
\newcommand{\MC}{\mathrm{MC}}

\newcommand{\UC}{\mathrm{UC}}
\newcommand{\one}{\mathds{1}}

\renewcommand{\KL}{\mathrm{KL}}

\providecommand{\UC}{\mathrm{UC}}
\usepackage{amsmath}
\usepackage{mathtools}

\mathtoolsset{showonlyrefs=true}

\title{Information-Theoretic Generalization for Set-Input Optimization-Valued Objectives}
\author{%
  Futoshi Futami\textsuperscript{1,2,3}\thanks{Equal contribution.}\qquad Masahiro Fujisawa\textsuperscript{1,3,4}\footnotemark[1]\\
  \textsuperscript{1}The University of Osaka\enspace
       \textsuperscript{2}The University of Tokyo\enspace
       \textsuperscript{3}RIKEN AIP\enspace
       \textsuperscript{4}Toyota Motor Corporation\\
  \texttt{futami.futoshi.es@osaka-u.ac.jp}\qquad\texttt{fujisawa@ist.osaka-u.ac.jp}%
}

\begin{document}
\maketitle

\begin{abstract}
Modern machine learning is increasingly evaluated using criteria such as multicalibration, CVaR, and Cressie--Read DRO, which target subgroup reliability, tail risk, and distributional robustness rather than averages of fixed per-example losses.
Existing information-theoretic generalization bounds for pointwise losses do not directly apply, because these empirical criteria depend on the entire batch and involve a post-hoc auxiliary optimization over witnesses, thresholds, or reweightings.
We formalize this structure as \emph{set-input optimization-valued objectives}.
In the balanced supersample setting,  we bound the expected optimized validation--training gap  in terms of an objective-dependent sub-Gaussian parameter and the square root of the prediction-based conditional mutual information (fCMI)  between the learned supersample prediction  table and the selector vector.
Combining this with fixed-predictor population--validation comparisons yields unified algorithm-dependent generalization bounds.
We verify  sub-Gaussianity of the selector gap using stability and convex $L_2$-Lipschitz certificates, obtaining concrete bounds for multicalibration, CVaR, and Cressie--Read DRO, with a stability-based single-index refinement.
\end{abstract}

\section{Introduction}
\label{sec:intro}
As machine learning systems are used in increasingly consequential settings, the question is not only how to optimize a model, but also what kind of performance guarantee should generalize. Classical statistical learning theory is organized around a simple template: each example incurs a fixed pointwise loss, and performance is measured by averaging this loss over the data distribution. Algorithm-dependent information-theoretic (IT) generalization bounds provide a complementary view of this template: MI~\cite{Xu2017}, CMI~\cite{steinke20a}, eCMI~\cite{hellstrom2022a,wang23}, and prediction-based fCMI~\cite{harutyunyan2021} relate generalization to how much the learned hypothesis, prediction, or loss reveals about the selected training sample. Because these bounds depend on the learning algorithm rather than only on a uniform hypothesis-class complexity, they can capture stability and implicit regularization; recent supersample refinements further sharpen this picture~\citep{wang23}.

Many modern evaluation criteria, however, are not averages of fixed per-example losses. In reliable prediction and uncertainty quantification, multicalibration asks whether predictive probabilities remain meaningful across many subpopulations and score regions, not only on average~\citep{johnson18a,gopalan22alow,blasiok2023unify}. In risk-sensitive learning, optimized-certainty-equivalent risks and conditional value-at-risk (CVaR) emphasize tail behavior rather than mean loss alone~\citep{bentalteboulle2007,rockafellar2000,lee2020risksensitive}. In robust learning, distributionally robust optimization (DRO) evaluates performance under adversarial reweightings of the empirical loss distribution~\citep{sagawa2020distributionally,an2021robust,Duchi2021}; related transport-based criteria depend on alignments of empirical distributions~\citep{cuturi2013sinkhorn}. These criteria are central when average loss can look benign while subgroup, tail, or worst-case failures remain severe.

What makes these objectives hard for existing IT analysis is not merely that they are non-pointwise. Recent work has already extended IT generalization beyond pointwise averages to fixed higher-order losses, including pairwise and triplet objectives~\citep{pmlr-v235-dong24a}; however, our setting is different. Here the empirical metric is \emph{set-input}: it depends on the entire batch of predictions and labels. It is \emph{optimization-valued}: after seeing the batch, the metric selects an auxiliary object by solving a downstream optimization problem, such as the most violated subgroup witness, a tail threshold, or an adversarial reweighting. Because this auxiliary choice is data-dependent, a one-point replacement can change not only one observation but also the optimizer that defines the metric. For multicalibration, the worst witness on the training batch need not coincide with the worst witness on the validation batch; for CVaR, the optimizing threshold can shift in the same way. This helps explain why existing analyses are family-specific, e.g., for ECE~\citep{futami2024information}, OCE/CVaR~\citep{lee2020risksensitive}, or robust optimization~\citep{an2021robust}.

We develop a unified  information-theoretic analysis of \emph{set-input optimization-valued objectives} in the supersample setting (Section~\ref{sec:general_set_input}). From each pair, a selector independently chooses one observation for training with equal probability, while the other is held out for validation. The central object is the  selector gap (Eq.~\eqref{eq:selector_gap_main}), defined as the optimized value on the  validation sample minus that on the training sample. This gap directly compares the values obtained after re-optimizing the auxiliary variable separately on the two samples, without fixing a common optimizer. Complementing the selector vector exchanges the two samples and reverses the sign of the gap; thus, the selector gap is antisymmetric. This antisymmetry lets us control the expected optimized validation--training gap through two  quantities (Theorem~\ref{thm:validation_training_gap}): the fCMI between the learned supersample prediction table and the selector vector, which measures algorithmic  dependence on the selector vector, and an objective-dependent sub-Gaussian parameter for the selector gap under the uniform selector law, which reflects the stability or geometry of the optimized  objective. To obtain a complete population--training  bound, we combine this result with an objective-specific fixed-predictor bound on the population--validation gap. The validation--training bound accounts for auxiliary optimization without an explicit entropy, cardinality,  Rademacher, or covering-number term for the auxiliary  class.

We give  two sufficient conditions for the selector gap to be sub-Gaussian under the uniform selector law. The first is a one-point stability certificate: if the objective changes little when one example is replaced, then the selector gap concentrates by bounded differences; this route also yields a single-index refinement that decomposes the validation--training gap into  one-point replacement increments and one CMI term per index (Section~\ref{sec:stability_certificate}; Theorem~\ref{thm:wang_general}). The second is a convex \(L_2\)-Lipschitz certificate, which captures sharper geometry when the objective is a convex function of a coordinatewise score vector (Section~\ref{sec:l2_certificate}). Together, these certificates give concrete bounds for multicalibration, CVaR, and Cressie--Read DRO, including the \(L_2\)-geometric improvement for DRO (Corollaries~\ref{cor:mc_stability}--\ref{cor:dro_l2}). We also give population--training decompositions using fixed-predictor MC, CVaR, and DRO comparisons (Section~\ref{sec:population_validation_terms}) and experiments illustrating computable local-information diagnostics (Section~\ref{sec:experiments}).

\section{Preliminaries}
\label{sec:prelim}
We fix notation used throughout. 
We use uppercase letters for random variables (RVs) and lowercase letters for their realizations. 
For $a\in\mathbb{N}$, write $[a]\coloneqq\{1,\dots,a\}$. 
Let $\mathcal{X}$ be the input space, $\mathcal{Y}$ the label space, and $\mathcal{Z}=\mathcal{X}\times\mathcal{Y}$. 
Let $\mathcal V$ be the prediction space, and let $\mathcal{D}$ be an unknown distribution on $\mathcal{Z}$. 
We denote mutual information (MI) and CMI by $I(\cdot;\cdot)$ and $I(\cdot;\cdot\mid\cdot)$, respectively. 
All expectations are over the random variables under discussion.

\subsection{IT generalization bound based on eCMI}
\label{sec:ordinary_it}
For ordinary pointwise losses, we use the standard supersample construction~\cite{steinke20a,hellstrom2022a}. 
Let $\widetilde{Z}\in\mathcal{Z}^{n\times 2}$ be an array of $2n$ i.i.d.~draws from $\mathcal{D}$, with entries $\widetilde{Z}_{i,j}=(\widetilde{X}_{i,j},\widetilde{Y}_{i,j})$ ($i\in[n],\ j\in\{0,1\}$).
We denote the $i$-th row by $\widetilde{Z}_i=(\widetilde{Z}_{i,0},\widetilde{Z}_{i,1})$.
Let the selector vector $U=(U_1,\ldots,U_n)\sim\mathrm{Unif}(\{0,1\}^n)$ be independent of $\widetilde{Z}$. Each component $U_i$ is called a selector; write $\overline{U}_i=1-U_i$. 
For a deterministic selector vector $u\in\{0,1\}^n$, we similarly write $\overline{u}_i=1-u_i$. 
The selected sample $\widetilde{Z}_U\coloneqq(\widetilde{Z}_{i,U_i})_{i=1}^n$ is used as the training sample, while the complementary sample $\widetilde{Z}_{\overline{U}}\coloneqq(\widetilde{Z}_{i,\overline{U}_i})_{i=1}^n$ is used as the held-out validation (or test) sample. Write $S:=\widetilde{Z}_U$, $S':=\widetilde{Z}_{\overline{U}}$. We use $n$-subscripts for batch size.
A learning algorithm receives $S$ and independent randomness $R\in\mathcal R$, outputs $W=A(S,R)$, and returns a predictor $f_W\in\calf:\mathcal X\to\mathcal V$.
We evaluate $f_W$ on all $2n$ supersample inputs and write $f_W(\widetilde{X})\coloneqq\bigl(f_W(\widetilde{X}_{i,j})\bigr)_{i\in[n],\,j\in\{0,1\}}\in\mathcal{V}^{n\times 2}$.

For comparison with the standard setting, let $\ell:\mathcal{V}\times\mathcal{Y}\to[0,1]$ be  a bounded pointwise loss function. For a predictor $f$ and a sample $s=((x_i,y_i))_{i=1}^n$,  define the population risk $L(f)\coloneqq\mathbb{E}_{Z\sim\mathcal D}\ell(f(X),Y)$ and the empirical risk $\widehat{L}(f;s)\coloneqq n^{-1}\sum_{i=1}^n \ell(f(x_i),y_i)$. When no confusion is possible,  write the loss value $\ell(f,z)$ as shorthand for $\ell(f(x),y)$ with $z=(x,y)$, and write the $i$-th supersample loss row as $\ell(f_W,\widetilde Z_i)\coloneqq(\ell(f_W,\widetilde Z_{i,0}),\ell(f_W,\widetilde Z_{i,1}))$.
The standard  evaluated CMI (eCMI) and functional CMI (fCMI) generalization bound~\citep{harutyunyan2021,hellstrom2022a} gives
\begin{align}
|\Ex\big[\widehat L(f_W;S')-\widehat L(f_W;S)\big]|
\le \frac{1}{n}\sum_{i=1}^n
\sqrt{2I(\ell(f_W,\widetilde{Z}_i);U_i\mid \widetilde{Z})}\le 
\sqrt{\frac{2}{n}I(f_W(\widetilde X);U\mid \widetilde{Z})}.
\label{eq:standard_ecmi}
\end{align}
 After fixing $f_W$, flipping one selector swaps only the  corresponding training and validation loss values.  The next subsection considers batch-level objectives without this row-wise structure.

\subsection{Motivating examples beyond ordinary losses}
\label{sec:set_input_prelim}
 Beyond the pointwise empirical risks of Section~\ref{sec:ordinary_it}, we consider three criteria. They become important precisely when mean performance is not sufficiently expressive: multicalibration~\citep{johnson18a,gopalan22alow,blasiok2023unify} captures subgroup and score-conditional reliability, CVaR~\citep{rockafellar2000,lee2020risksensitive} captures tail-sensitive risk, and distributionally robust optimization (DRO)~\citep{rahimian2019dro,sagawa2020distributionally} captures worst-case performance under adversarial reweighting. Throughout, \(S=((X_i,Y_i))_{i=1}^n\in\mathcal{Z}^n\) and \(Z=(X,Y)\sim\mathcal{D}\).

\paragraph{Multicalibration.}
Multicalibration (MC) controls calibration over subpopulations and score-space tests~\citep{johnson18a,gopalan22alow,blasiok2023unify}. Here \(\mathcal{Y}=\{0,1\}\) and \(\mathcal{V}=[0,1]\). Let \(\Pi_{\mathrm{MC}}\coloneqq \mathcal C\times\mathcal H\times\{\pm1\}\), where \(\mathcal{C}\subseteq\{c:\mathcal{X}\to[0,1]\}\), \(\mathcal{H}\subseteq\{h:[0,1]\to[-1,1]\}\), and write \(\pi=(c,h,\eta)\). The population and empirical values are
\begin{align}
\Phi_{\MC}(f_w;\cald)
&\coloneqq 
\sup_{(c,h,\eta)\in\Pi_{\MC}}
\eta\,\Ex\big[c(X)h(f_w(X))(Y-f_w(X))\big],
\label{eq:mc_population_prelim}
\\[-0.35ex]
\Phi_{\MC}(f_w;S)
&\coloneqq 
\sup_{(c,h,\eta)\in\Pi_{\MC}}
\eta\frac{1}{n}\sum_{i=1}^n c(X_i)h(f_w(X_i))(Y_i-f_w(X_i)).
\label{eq:mc_objective_prelim}
\end{align}
The empirical value selects the subgroup--score witness and sign after seeing the batch, so the selected witness may change when the evaluation batch changes.

\paragraph{OCE risks and CVaR.}
OCE risks are tail-sensitive criteria~\citep{bentalteboulle2007,rockafellar2000,lee2020risksensitive}; for bounded loss, \(\Phi_{\OCE,\psi}(f_w;\cald)=\inf_\lambda\{\lambda+\Ex\psi(\ell(f_w(X),Y)-\lambda)\}\). For CVaR at level \(\alpha\in(0,1)\), write \(\bar\alpha\coloneqq1-\alpha\); this is the case \(\psi_\alpha(t)=\bar\alpha^{-1}[t]_+\). The population and empirical CVaR values are
\begin{align}
\Phi_{\CVaR,\alpha}(f_w;\cald)
&\coloneqq 
\inf_{\tau\in[0,1]}\big\{\tau+\frac{1}{\bar\alpha}\Ex(\ell(f_w(X),Y)-\tau)_+\big\},
\label{eq:cvar_population}
\\[-0.35ex]
\Phi_{\CVaR,\alpha}(f_w;S)
&\coloneqq 
\inf_{\tau\in[0,1]}\big\{\tau+\frac{1}{\bar\alpha n}\sum_{i=1}^n (\ell(f_w,Z_i)-\tau)_+\big\}.
\label{eq:cvar_empirical}
\end{align}
The empirical CVaR threshold is chosen from the observed loss distribution, so the objective is not an average of a fixed per-example loss. More general OCE examples are in Appendix~\ref{app:reliability_oce}.

\paragraph{Cressie--Read DRO.}
DRO allows adversarial reweighting within a divergence ball~\citep{Duchi2021}. We use the Cressie--Read \(f\)-divergence generator \(f_c(t)=(t^c-ct+c-1)/\{c(c-1)\}\), with \(c=2\) giving Pearson \(\chi^2\) and limits \(c\to1,0\) giving KL and empirical-likelihood divergences. In what follows, we restrict to \(c>1\) and write
\(c^\ast:=c/(c-1)\).  Given a radius \(\epsilon\ge0\), and density ratio \(\omega=dQ/d\mathcal{D}\), 
\begin{align}
\Phi_{\DRO,c,\epsilon}(f_w;\cald)
&\coloneqq
\sup_{\omega\ge0}\Bigl\{\Ex_{\cald}[\omega(Z)\ell(f_w(X),Y)]
:
\Ex_{\cald}\omega=1,\ \Ex_{\cald}f_c(\omega)\le\epsilon\Bigr\}.
\label{eq:cr_dro_population_prelim}
\end{align}
On a finite evaluation batch, the ambiguity set becomes
\begin{align}
\Phi_{\DRO,c,\epsilon}(f_w;S)
\coloneqq \!
\sup_{q\in\R_+^n}
\big\{
\frac1n\sum_{i=1}^n q_i\ell(f_w(X_i),Y_i)
:
\frac1n\sum_{i=1}^n q_i=1,
\frac1n\sum_{i=1}^n q_i^c\le 1+c(c-1)\epsilon
\big\}.
\label{eq:cr_dro_empirical_prelim}
\end{align}
Appendix~\ref{app:reliability_fdro} shows the density-ratio and dual forms used later. 

For notational clarity, we write \(\Phi_{\star,n}(f;S)\equiv\Phi_{\star}(f;S)\) when emphasizing that an objective is empirical, and, for fixed objective parameters, \(\Delta_{\star}\coloneqq\Phi_{\star,n}(f_W;S')-\Phi_{\star,n}(f_W;S)\); for example, \(\Delta_{\DRO}=\Phi_{\DRO,c,\epsilon,n}(f_W;S')-\Phi_{\DRO,c,\epsilon,n}(f_W;S)\).

\paragraph{Why ordinary eCMI does not directly apply.}
For ordinary losses, once a predictor \(f\) is fixed, every row of the supersample table is evaluated by the same pointwise loss \(\ell(f,\cdot)\). Flipping one selector swaps a single training example with its validation counterpart, and only that row changes. MC does not have this row-wise form: its value first looks at the whole batch and then selects a witness \(\widehat\pi=(\widehat c,\widehat{h},\widehat\eta)\in\Pi_{\MC}\). In the supersample notation, let \(\widehat{\pi}_{\mathrm{tr}}\) and \(\widehat{\pi}_{\mathrm{val}}\) be witnesses selected for \(\Phi_{\MC,n}(f_W;\widetilde Z_U)\) and \(\Phi_{\MC,n}(f_W;\widetilde Z_{\overline U})\), respectively. A naive eCMI argument could freeze \(\widehat{\pi}_{\mathrm{tr}}\) and compare the same signed MC average on the training and validation batches. The target gap, however, is \(\Phi_{\MC,n}(f_W;\widetilde Z_{\overline U})-\Phi_{\MC,n}(f_W;\widetilde Z_U)\), where the validation batch may choose the different witness \(\widehat{\pi}_{\mathrm{val}}\). Bounding this mismatch by ranging uniformly over all witnesses gives a correct but uniform-convergence-type argument over \(\Pi_{\MC}\), rather than the signed algorithm-dependent cancellation exploited by eCMI.   The same issue for CVaR thresholds and DRO weights motivates this framework.

\section{Main Results}
\label{sec:main}
Sections~\ref{sec:general_set_input}--\ref{sec:l2_certificate} control the optimized validation--training term
$\Delta_\Phi\coloneqq\Phi_n(f_W;S')-\Phi_n(f_W;S)$.
Section~\ref{sec:general_set_input} proves the fCMI bound  using an objective-dependent sub-Gaussian parameter. Sections~\ref{sec:stability_certificate}--\ref{sec:l2_certificate} bound this parameter using stability and convex \(L_2\)-Lipschitz continuity. Throughout these subsections, write $I_{\mathrm{fCMI}}\coloneqq I(f_W(\widetilde X);U\mid\widetilde Z)$. Section~\ref{sec:population_validation_terms} then adds objective-specific population--validation comparisons to obtain population--training statements.

\subsection{Information-theoretic bound for set-input optimization-valued objectives}
\label{sec:general_set_input}
The examples in Section~\ref{sec:set_input_prelim} reveal two structural departures from ordinary pointwise losses. First, the empirical criterion is \emph{set-input}: it is a function of the whole batch of predictions and labels, rather than a sum of fixed row-wise losses. Second, it is \emph{optimization-valued}: after the batch is observed, an auxiliary object is chosen by solving a batch-level optimization problem. We call objectives with both features \textbf{set-input optimization-valued objectives}.

Formally, let \(\Pi\) be a nonempty auxiliary space. A candidate \(\pi\in\Pi\) specifies a measurable score \(\rho_\pi:\calv\times\calz\to\R\) and a post-aggregation map \(\psi_\pi:\R\to\R\). The score input is \((v,z)=(f_w(x),(x,y))\), so the objective value on a batch is determined by the pairs \((f_w(x_i),z_i)\). Given a predictor \(f_w\in\calf\) and a sample \(s=((x_i,y_i))_{i=1}^n\in\calz^n\), define
\begin{align}
\Phi_n(f_w;s)
\coloneqq 
\operatorname{Opt}_{\pi\in\Pi}
\psi_\pi\!\left(
\frac1n\sum_{i=1}^n
\rho_\pi(f_w(x_i),z_i)
\right),
\qquad
\operatorname{Opt}\in\{\sup,\inf\}.
\label{eq:set_input_avg_def}
\end{align}
The population target is \(\Phi(f_w;\cald)\coloneqq\operatorname{Opt}_{\pi\in\Pi}\psi_\pi(\Ex_{Z\sim\cald}\rho_\pi(f_w(X),Z))\). Throughout this section, optimized values are assumed finite and measurable; separable versions for uncountable \(\Pi\) are in Appendix~\ref{app:measurability_set_input}. MC uses \(\pi=(c,h,\eta)\), CVaR uses \(\pi=\tau\), and Cressie--Read DRO enters Eq.~\eqref{eq:set_input_avg_def} through the dual form in Appendix~\ref{app:cr-dual-average}; its sharper \(L_2\) certificate later uses the primal reweighting form. Conditional on \((f_W(\widetilde X),\widetilde Z)\), all values \(\{\Phi_n(f_W;\widetilde Z_u):u\in\{0,1\}^n\}\) are fixed. Thus the analysis controls the optimized mismatch directly.

Given a fixed predictor \(f_w\), a supersample \(\widetilde z\), and a selector vector \(u\), define the  selector gap
\begin{align}
\Gamma_{f_w,\widetilde z}(u)
\coloneqq 
\Phi_n(f_w;\widetilde z_{\overline u})-
\Phi_n(f_w;\widetilde z_u),
\qquad u\in\{0,1\}^n.
\label{eq:selector_gap_main}
\end{align}
The selector gap is antisymmetric: replacing \(u\) by \(\overline u\) exchanges the two samples and reverses the sign of the gap. Pairing every selector vector with its complement under the uniform selector law gives  \(\Ex_{U\sim\mathrm{Unif}(\{0,1\}^n)}\Gamma_{f_w,\widetilde z}(U)=0\); see Lemma~\ref{lem:uniform_selector_pairing_app} for the proof. We impose the following sub-Gaussian control on this selector gap.
\begin{assumption}[Sub-Gaussianity]
\label{ass:selector_gap_subgaussian}
There is a deterministic constant \(L_{\Gamma,n}<\infty\), uniform over fixed predictors \(f_w\) and supersample realizations \(\widetilde z\), such that, for \(Q_U=\mathrm{Unif}(\{0,1\}^n)\),
\begin{align}
\log \Ex_{Q_U}
\exp\{\lambda\Gamma_{f_w,\widetilde{z}}(U)\}
\le
\frac{\lambda^2L_{\Gamma,n}^2}{2},
\qquad
\forall \lambda\in\R.
\label{eq:selector_gap_mgf_main}
\end{align}
\end{assumption}

Assumption~\ref{ass:selector_gap_subgaussian} is fixed-predictor, fixed-supersample concentration under \(Q_U\); it summarizes objective geometry, not the actual conditional selector law \(P_{U\mid f_W(\widetilde X),\widetilde Z}\), which is charged through fCMI. Assumption~\ref{ass:selector_gap_subgaussian} is verified below through either the stability certificate of Section~\ref{sec:stability_certificate} or the convex \(L_2\)-Lipschitz certificate of Section~\ref{sec:l2_certificate}; Table~\ref{tab:gap_summary} summarizes the resulting bounds on $L_{\Gamma,n}$.
\begin{theorem}[Validation--Training bound]
\label{thm:validation_training_gap}
Under Assumption~\ref{ass:selector_gap_subgaussian}, we have
\begin{align}
|\Ex\Delta_\Phi|
&\le
L_{\Gamma,n}\sqrt{2I_{\mathrm{fCMI}}},
\label{eq:signed_val_train_gap}
\\[-0.25ex]
\Ex|\Delta_\Phi|
&\le
L_{\Gamma,n}\sqrt{2(I_{\mathrm{fCMI}}+\log 2)}.
\label{eq:absolute_val_train_gap}
\end{align}
\end{theorem}
\paragraph{Interpretation.}
Theorem~\ref{thm:validation_training_gap} factorizes the bound on the optimized validation--training gap into a learning-dependent term and an objective-dependent term. To see this, condition on \(G=(f_W(\widetilde X),\widetilde Z)\). Conditional on \(G\), all optimized values $\{\Phi_n(f_W;\widetilde Z_u):u\in\{0,1\}^n\}$ are fixed, and the auxiliary optimizer has already been absorbed into the deterministic  function $u\mapsto \Gamma_{f_W,\widetilde Z}(u)$. The Donsker--Varadhan step therefore compares the actual conditional selector law \(P_{U\mid G}\) with the uniform reference law \(Q_U\), paying only \(D(P_{U\mid G}\|Q_U)\). Averaging over \(G\) gives \(I(f_W(\widetilde X);U\mid\widetilde Z)=I_{\mathrm{fCMI}}\). Thus fCMI measures how much the learned supersample prediction table depends on which half of each pair was used for training.

The other factor, \(L_{\Gamma,n}\), captures the geometry of the optimized objective under the reference selector law. Auxiliary optimization is not free; its effect enters through the moment-generating function in Assumption~\ref{ass:selector_gap_subgaussian}. What is avoided in the validation--training term is an explicit uniform-convergence penalty over \(\Pi\), such as an entropy, cardinality, covering-number, or Rademacher complexity term. This is the desired behavior for the validation--training term: if the learned predictor is independent of the selector vector, then $|\Ex\Delta_\Phi|=0$ by symmetry, regardless of the size of $\Pi$. Eq.~\eqref{eq:signed_val_train_gap} reflects this cancellation, since the right-hand side vanishes when \(I_{\mathrm{fCMI}}=0\). 

Sections~\ref{sec:stability_certificate}--\ref{sec:l2_certificate} are devoted to certifying \(L_{\Gamma,n}\). Different certificates expose different structural properties of the same set-input objective. The stability route (Section~\ref{sec:stability_certificate}) controls coordinate-wise replacement sensitivity, corresponding to an \(L_\infty\)-type view of the optimizer. The convex \(L_2\)-Lipschitz route (Section~\ref{sec:l2_certificate}) instead uses Euclidean geometry of the score or weight vector, which is sharper when the optimizer spreads mass across the batch. Consequently, the same algorithmic quantity \(I_{\mathrm{fCMI}}\) pairs with different  sub-Gaussian parameters depending on the objective: MC is controlled by bounded linear witnesses, CVaR reflects the tail level \(\bar\alpha\), and Cressie--Read DRO distinguishes the \(L_\infty\) stability scale from the sharper \(L_2\) reweighting scale. Table~\ref{tab:gap_summary} summarizes these choices.

\suppressfloats[t]
\begin{table}[t]
\centering
\small
\caption{Upper bounds on $L_{\Gamma,n}$ under stability and convex $L_2$-Lipschitz certificates.}
\begin{tabular}{cccc}
\toprule
Objective
& Uniform stability (Assumption~\ref{ass:one_point_stability})
& Convex $L_2$-Lipschitz (Assumption~\ref{ass:convex_l2_score_vector_main})
& Preferred
\\
\midrule \midrule 
MC
& $2n^{-1/2}$
& $4n^{-1/2}$ 
& Stability\\
CVaR
& $(\bar\alpha\sqrt n)^{-1}$
& \(2(\bar\alpha n)^{-1/2}\)
& Best of two\\
CR DRO
& \(\{1+c(c-1)\epsilon\}^{\tfrac1c}n^{\tfrac1c-\tfrac12}\) 
& \(2\{1+c(c-1)\epsilon\}^{\tfrac1c}n^{-\tfrac{1}{c^\ast\vee2}}\)
& Convex 
\\
\bottomrule
\end{tabular}
\label{tab:gap_summary}
\end{table}

\subsection{Stability certificate and single-index refinement}
\label{sec:stability_certificate}
Our first certificate is uniform stability,  a sufficient condition for Assumption~\ref{ass:selector_gap_subgaussian}:
\begin{assumption}[Uniform stability]
\label{ass:one_point_stability}
There exists a possibly \(n\)-dependent constant \(\beta_n>0\) such that, for every fixed predictor \(f_w\) and every pair of samples \(s,s'\in\calz^n\) that differ in at most one coordinate,
\begin{align}
\bigl|
\Phi_n(f_w;s)-\Phi_n(f_w;s')
\bigr|
\le
\beta_n.
\label{eq:one_point_stability}
\end{align}
\end{assumption}

\begin{lemma}[Stability certificate]
\label{lem:stability_certificate_main}
Under Assumption~\ref{ass:one_point_stability}, Assumption~\ref{ass:selector_gap_subgaussian} holds with
$L_{\Gamma,n}\le \beta_n \sqrt n$ and
\begin{align}
|\Ex\Delta_\Phi|
\le
\beta_n\sqrt{2nI_{\mathrm{fCMI}}}.
\label{eq:stability_vt_general_main}
\end{align}
\end{lemma}
In the standard eCMI bound in Eq.~\eqref{eq:standard_ecmi}, the information term is row-wise. For set-input objectives, the following theorem combines  the fCMI bound using the full vector \(U\) with a single-index bound.
\begin{theorem}[Single-index stability bound]
\label{thm:wang_general}
Define, for each $t\in[n]$ and $b\in\{0,1\}$, $S^{(t,b)}\coloneqq (\widetilde{Z}_{1,U_1},\ldots,\widetilde{Z}_{t-1,U_{t-1}},\widetilde{Z}_{t,b},\widetilde{Z}_{t+1,\overline{U}_{t+1}},\ldots,\widetilde{Z}_{n,\overline{U}_n})$. 
Set \(D_t\coloneqq\Phi_n(f_W;S^{(t,0)})-\Phi_n(f_W;S^{(t,1)})\) and \(\varepsilon_t\coloneqq2U_t-1\). If Assumption~\ref{ass:one_point_stability} holds, then \(|D_t|\le \beta_n\), each \(D_t\) is measurable with respect to \((f_W(\widetilde{X}),\widetilde{Z},U_{-t})\), and
\begin{align}
\Phi_n(f_W;S')-
\Phi_n(f_W;S)
= \sum_{t=1}^n \varepsilon_t D_t.
\label{eq:wang_telescoping_identity}
\end{align}
For the local pieces, write \(I_{\mathrm{loc},t}\coloneqq I(D_t;U_t\mid\widetilde Z)\). Then
\begin{align}
|\Ex\Delta_\Phi|
&\le
\beta_n
\min\Bigl\{
\sqrt{2nI_{\mathrm{fCMI}}},
\sum_{t=1}^n\sqrt{2I_{\mathrm{loc},t}}
\Bigr\}.
\label{eq:wang_min_bound}
\end{align}
\end{theorem}

 Eq.~\eqref{eq:wang_min_bound} uses the sharper of two routes. The global term, \(I_{\mathrm{fCMI}}\), uses  information from the full prediction table; VC or growth-function controls apply only after the table is binary or otherwise finite-valued, with the binary case given by Lemma~\ref{lem:vc_fcmi_growth}. The local term, \(I_{\mathrm{loc},t}\), uses scalar information for the pairs $(D_t,U_t)$ and is often easier to estimate directly. 
For pointwise empirical risks, the single-index decomposition recovers the row-wise structure underlying standard eCMI bounds and their fast-rate refinements; see Appendices~\ref{app:single_input_regime} and~\ref{app:shifted_sign_attempt_added}.

We now apply the stability framework to the MC, CVaR, and DRO objectives introduced in Section~\ref{sec:set_input_prelim}. In each case, the task is to evaluate the corresponding stability coefficient. The global information \(I_{\mathrm{fCMI}}\) is common across objectives, while the local information is objective-specific: for \(\star\in\{\MC,\CVaR,\DRO\}\), write \(I_{\mathrm{loc},t}^{\star}\coloneqq I(D_t^{\star};U_t\mid\widetilde Z)\).

\begin{corollary}[Multicalibration]
\label{cor:mc_stability}
For the MC objective~\eqref{eq:mc_objective_prelim}, Assumption~\ref{ass:one_point_stability} holds with \(\beta_n=\beta_{\MC,n}\coloneqq2/n\). Writing \(D_t^{\MC}\) for the local increment in Theorem~\ref{thm:wang_general} specialized to MC, Theorem~\ref{thm:wang_general} gives
\begin{align}
|\Ex\Delta_{\MC}|
&\le
\frac{2}{n}
\min\Bigl\{
\sqrt{2nI_{\mathrm{fCMI}}}, 
\sum_{t=1}^n\sqrt{2I_{\mathrm{loc},t}^{\MC}}
\Bigr\}.
\label{eq:mc_wang_bound_main}
\end{align}
\end{corollary}
We next show that \(I_{\mathrm{fCMI}}\) can be controlled by the number of distinct prediction tables. For MC, since $\calv$ is real-valued, we focus on the pseudo-dimension~\citep{Mohri}, $\operatorname{Pdim}(\mathcal F)$.
\begin{corollary}[Binned MC with real-valued scores]
\label{cor:binned_mc_pdim_main}
Let \(I_1,\ldots,I_B\) be a fixed partition of \([0,1]\), take the MC score-test class \(\cH_B\coloneqq\{v\mapsto\one\{v\in I_b\}:b\in[B]\}\), and let \(\Phi_{\MC,B}\) be the corresponding binned objective from Appendix~\ref{app:binned-mc-pdim}. Assume that the pseudo-dimension \(\operatorname{Pdim}(\mathcal F)=d_0<\infty\). Then, with \(\Delta_{\MC,B}\coloneqq\Phi_{\MC,B}(f_W;S')-\Phi_{\MC,B}(f_W;S)\),
\begin{align}
|\Ex\Delta_{\MC,B}|
\lesssim
\sqrt{\frac{d_0\log\{en(B+n)/d_0\}}{n}}
+\frac1n.
\label{eq:binned_mc_pdim_main}
\end{align}
\end{corollary}

The same stability argument applies to CVaR and DRO. To keep the main text concise, we report the stability coefficient \(\beta_n\) and the fCMI/local-information bounds; complete statements and proofs are in Appendix~\ref{app:section3_proofs}. 
The stability/local-information bounds in Corollaries~\ref{cor:mc_stability}, \ref{cor:cvar_stability}, and~\ref{cor:dro_stability_suboptimal} require no complexity assumption. VC/growth rates are used only after the prediction table is binary or otherwise finite-valued; Lemma~\ref{lem:vc_fcmi_growth} gives the binary case, and real-valued binned scores require the discretized pseudo-dimension control of Corollary~\ref{cor:binned_mc_pdim_main} and Appendix~\ref{app:binned-mc-pdim}.
\begin{corollary}[CVaR]
\label{cor:cvar_stability}
For the CVaR objective~\eqref{eq:cvar_empirical}, Assumption~\ref{ass:one_point_stability} holds with \(\beta_n=\beta_{\CVaR,n}\coloneqq(\bar\alpha n)^{-1}\). Writing \(D_t^{\CVaR}\) for the corresponding local increment, Theorem~\ref{thm:wang_general} gives
\begin{align}
|\Ex\Delta_{\CVaR}|
&\le
\frac{1}{\bar\alpha n}
\min\Bigl\{
\sqrt{2nI_{\mathrm{fCMI}}}, 
\sum_{t=1}^n\sqrt{2I_{\mathrm{loc},t}^{\CVaR}}
\Bigr\}.
\label{eq:cvar_stability_bounds_main}
\end{align}
\end{corollary}

\begin{corollary}[CR DRO]
\label{cor:dro_stability_suboptimal}
Let \(c>1\), \(c^\ast=c/(c-1)\), and \(\kappa_{c,\epsilon}\coloneqq \{1+c(c-1)\epsilon\}^{1/c}\).  For the empirical CR DRO objective~\eqref{eq:cr_dro_empirical_prelim}, Assumption~\ref{ass:one_point_stability} holds with \(\beta_n=\beta_{\DRO,n}\coloneqq\kappa_{c,\epsilon}n^{1/c-1}\).  Writing \(D_t^{\DRO}\) for the corresponding local increment, Theorem~\ref{thm:wang_general} gives
\begin{align}
|\Ex\Delta_{\DRO}|
&\le
\kappa_{c,\epsilon}n^{1/c-1}
\min\Bigl\{
\sqrt{2nI_{\mathrm{fCMI}}}, 
\sum_{t=1}^n\sqrt{2I_{\mathrm{loc},t}^{\DRO}}
\Bigr\}.
\label{eq:dro_stability_bounds_main}
\end{align}
\end{corollary}
For binary prediction models from a class of VC dimension \(d\), Lemma~\ref{lem:vc_fcmi_growth} gives \(I_{\mathrm{fCMI}}\lesssim d\log(en/d)\). In that discrete-table case, the global route scales as \(\mathcal O(\bar\alpha^{-1}\sqrt{d\log(en/d)/n})\) for CVaR and \(\mathcal O(\kappa_{c,\epsilon}n^{1/c-1/2}\sqrt{d\log(en/d)})\) for stability-based DRO. This stability-based DRO bound is generally suboptimal: \citet{Duchi2021} showed that the relevant Cressie--Read plug-in rate scales as $n^{-1/(c^\ast\vee2)}$. The stability route is governed by the coordinate-wise \(\ell_\infty\) sensitivity of the adversarial weights, whereas the sharper CR DRO bound is governed by their \(\ell_2\) geometry. This leads to the convex \(L_2\)-Lipschitz certificate in Section~\ref{sec:l2_certificate}.

\subsection{Convex \(L_2\)-Lipschitz certificate}
\label{sec:l2_certificate}
The stability route is sharp when a one-point replacement has small effect, but it can be too crude for objectives whose optimizer spreads mass across the whole batch.  We therefore introduce a geometric certificate based on convex \(L_2\)-concentration. 
\begin{assumption}[Convex \(L_2\)-Lipschitz score-vector assumption]
\label{ass:convex_l2_score_vector_main}
For every fixed predictor \(f_w\) and sample $s=(z_1,\dots,z_n)$, the empirical objective in Eq.~\eqref{eq:set_input_avg_def} can be written as
 $\Phi_n(f_w;s)
=
\tilde\psi\bigl(R_{f_w,n}(s)\bigr)$, 
where \(R_{f_w,n}:\calz^n\to[0,1]^n\) is coordinate-wise, meaning that its \(i\)-th coordinate depends only on \(z_i\), and \(\tilde\psi:[0,1]^n\to\R\) is convex.  Moreover, \(\tilde\psi\) is \(L_{\psi,n}\)-Lipschitz with respect to the Euclidean norm, meaning that for any $r,r'\in [0,1]^n$, $|\tilde\psi(r)-\tilde\psi(r')|\leq L_{\psi,n}\|r-r'\|_2$.
\end{assumption}
Under this convex \(L_2\)-Lipschitz assumption,  Assumption~\ref{ass:selector_gap_subgaussian} follows directly.  Here the relevant scale is \(\sqrt{2I_{\mathrm{fCMI}}}\), whereas the stability route uses \(\sqrt{2nI_{\mathrm{fCMI}}}\).
\begin{lemma}[Convex \(L_2\)-Lipschitz certificate]
\label{lem:convex_l2_certificate_main}
Under Assumption~\ref{ass:convex_l2_score_vector_main}, Assumption~\ref{ass:selector_gap_subgaussian} holds with
$
L_{\Gamma,n}
\le
2L_{\psi,n}
$ and \(|\Ex\Delta_\Phi|\le2L_{\psi,n}\sqrt{2I_{\mathrm{fCMI}}}\).
\end{lemma}

For MC, the stability bound in Corollary~\ref{cor:mc_stability} is sharper, but the same geometry is transparent: MC is a supremum of linear witness averages,  and has a convex extension obtained by interpolating the witness scores; Lemma~\ref{lem:mc_selector_interpolation_certificate} gives the MC entry in Table~\ref{tab:gap_summary}.

\begin{corollary}[CVaR via the convex \(L_2\) certificate]
\label{cor:cvar_l2}
Using the risk-envelope form of Appendix~\ref{app:reliability_oce}, the CVaR objective~\eqref{eq:cvar_empirical} satisfies Assumption~\ref{ass:convex_l2_score_vector_main} with \(L_{\psi,n}=(\bar\alpha n)^{-1/2}\).
Combining this with Corollary~\ref{cor:cvar_stability}, we have
\begin{align}
L_{\Gamma,n}^{\CVaR}
&\le
\min\{\bar\alpha^{-1}n^{-1/2},\,2(\bar\alpha n)^{-1/2}\},
\label{eq:cvar_lgamma_main}
\end{align}
and 
\begin{align}
|\Ex\Delta_{\CVaR}|
&\le
\min\{\bar\alpha^{-1}n^{-1/2},\,2(\bar\alpha n)^{-1/2}\}\sqrt{2I_{\mathrm{fCMI}}}.
\label{eq:cvar_l2_vt_bound_main}
\end{align}
\end{corollary}
Thus, depending on the tail level, the convex \(L_2\) route may improve the constant relative to the stability route.  More importantly, for CR DRO it improves the rate itself.

\begin{corollary}[CR DRO]
\label{cor:dro_l2}
Let \(c>1\), \(c^\ast=c/(c-1)\), and \(\kappa_{c,\epsilon}\coloneqq \{1+c(c-1)\epsilon\}^{1/c}\).  For the empirical CR DRO objective~\eqref{eq:cr_dro_empirical_prelim}, Assumption~\ref{ass:convex_l2_score_vector_main} holds with $L_{\psi,n}^{\DRO}=\kappa_{c,\epsilon}n^{-1/(c^\ast\vee2)}$. Consequently,
\begin{align}
|\Ex\Delta_{\DRO}|
&\le
2\kappa_{c,\epsilon}n^{-1/(c^\ast\vee2)}
\sqrt{2I_{\mathrm{fCMI}}}.
\label{eq:dro_vt_bound_main}
\end{align}
\end{corollary}
Compared with the stability result in
Corollary~\ref{cor:dro_stability_suboptimal}, the convex \(L_2\) certificate
replaces coordinate-wise \(\ell_\infty\) sensitivity by the Euclidean geometry
of the adversarial weight vector. In particular, for binary prediction tables
generated by a class of VC dimension \(d\), Lemma~\ref{lem:vc_fcmi_growth}
gives \(I_{\mathrm{fCMI}}\lesssim d\log(en/d)\), and hence
$|\Ex\Delta_{\DRO}|
=
\mathcal O\!\left(
\kappa_{c,\epsilon}
n^{-1/(c^\ast\vee 2)}
\sqrt{d\log(en/d)}
\right)$.
 Thus the bound on \(L_{\Gamma,n}\) recovers the exponent of \citet{Duchi2021}, up to logarithmic
factors, while the additional fCMI factor accounts for algorithmic
dependence on the training selector vector.

\subsection{Population--training generalization bound}
\label{sec:population_validation_terms}
 We connect the preceding validation--training bounds to the population target through
\begin{align}
\Phi(f_W;\cald)-\Phi_n(f_W;S)
=
\bigl\{\Phi(f_W;\cald)-\Phi_n(f_W;S')\bigr\}
+
\bigl\{\Phi_n(f_W;S')-\Phi_n(f_W;S)\bigr\}.
\label{eq:population_validation_split}
\end{align}
The second term is controlled by Theorem~\ref{thm:validation_training_gap}. Here each row contains two i.i.d.\ draws and \(U\) only orders them, so \((S,S')\sim\cald^n\otimes\cald^n\); hence \(S'\perp\!\!\!\perp(S,R)\) and \(S'\perp\!\!\!\perp f_W=A(S,R)\). The first term is a fixed-predictor population--validation comparison. This split separates the fCMI-controlled validation--training term from the objective-specific approximation of the population metric by an independent batch, possibly involving empirical-process terms over witnesses, thresholds, or weights.
\paragraph{Absolute comparisons.} Given $S'=(Z'_i)_{i=1}^n$, the fixed-predictor term is controlled by the corresponding empirical process.  In the average-type form of Eq.~\eqref{eq:set_input_avg_def}, write \(\mu_\pi(f)\coloneqq \Ex_{Z\sim\cald}\rho_\pi(f(X),Z)\) and \(\widehat\mu_{\pi,n}(f;S')\coloneqq n^{-1}\sum_{i=1}^n\rho_\pi(f(X'_i),Z'_i)\). Then
\begin{align}
|\Phi(f_W;\cald)-\Phi_n(f_W;S')|
&\leq
\sup_{\pi\in\Pi}
\left|\psi_\pi\bigl(\mu_\pi(f_W)\bigr)-
\psi_\pi\bigl(\widehat\mu_{\pi,n}(f_W;S')\bigr)\right|.
\label{eq:fixed_predictor_emp_process_main}
\end{align}
This term is controlled by objective-specific fixed-predictor uniform-convergence or plug-in arguments rather than by MI/CMI.  For fixed \(f\), write \(\UC_n^\Phi(f)\coloneqq\Ex_{S'\sim\cald^n}|\Phi(f;\cald)-\Phi_n(f;S')|\).  For MC fixed-predictor comparison, put \(\cC_{\pm}\coloneqq\{\eta c:c\in\cC,\eta\in\{\pm1\}\}\). Given \(S=(Z_i)_{i=1}^n=((X_i,Y_i))_{i=1}^n\), for any class \(\mathcal G\) define \(\widehat{\mathfrak R}_{S}(\mathcal G)\coloneqq\Ex_{\sigma}[\sup_{g\in\mathcal G}n^{-1}\sum_{i=1}^n\sigma_i g(Z_i)]\) and \(\mathfrak R_{\cald,n}(\mathcal G)\coloneqq\Ex_{S\sim\cald^n}\widehat{\mathfrak R}_{S}(\mathcal G)\). Write \(N(\delta,\cH,\|\cdot\|_\infty)\) for the score-test sup-norm covering number.
\begin{corollary}[Multicalibration]
\label{cor:mc_population_validation}
For the MC objective~\eqref{eq:mc_objective_prelim}, 
\begin{align}
&\Ex[\UC_n^{\MC}(f_W)]\lesssim 
\mathfrak R_{\cald,n}(\cC_{\pm})
+
\inf_{\delta\ge 0}
\left\{
\delta+\int_{\delta}^{1}
\sqrt{\frac{\log N(\varepsilon,\cH,\|\cdot\|_\infty)}{n}}\,d\varepsilon
\right\},
\label{eq:uniform_gen_main}\\
&\Ex\bigl|
\Phi_{\MC}(f_W;\cald)-\Phi_{\MC,n}(f_W;S)
\bigr|
\le
\Ex[\UC_n^{\MC}(f_W)]
+
2\sqrt{\frac{2\bigl(I_{\mathrm{fCMI}}+\log 2\bigr)}{n}}.
\label{eq:mc_abs_cor}
\end{align}
\end{corollary}
For finite \(\cC\) and \(\cH\), including standard binned MC, \(\Ex[\UC_n^{\MC}(f_W)]=\mathcal{O}(\sqrt{\log(2|\cC||\cH|)/n})\). When \(\cH=\operatorname{Lip}_1([0,1],[-1,1])\), Appendix~\ref{app:mc_smooth_ce_refinement} gives a direct entropy-integral proof of \(\Ex[\UC_n^{\MC}(f_W)]\lesssim n^{-1/2}+\sqrt{\log(2|\cC|)/n}\) for finite \(\cC\). %
Combining this with Corollary~\ref{cor:binned_mc_pdim_main}, we have a corresponding sample-complexity of
$\Ex\bigl|\Phi_{\MC}(f_W;\cald)-\Phi_{\MC,n}(f_W;S)\bigr|$ for the binned MC.

\paragraph{One-sided comparisons.} For supremum-type objectives witten with a common feasible set and convex aggregation maps, Jensen's inequality gives \(\Phi(f;\cald)\le\Ex_{S'\sim\cald^n}[\Phi_n(f;S')\mid f]\). Thus an upper bound on the expected population--training gap follows from the validation--training term alone. MC satisfies this comparison; combining it with Theorem~\ref{thm:wang_general} gives
\begin{align}
\Ex\Phi_{\MC}(f_W;\cald)
&\le
\Ex\Phi_{\MC,n}(f_W;S)
+
\frac{2}{n}
\min\Bigl\{
\sqrt{2nI_{\mathrm{fCMI}}},
\sum_{t=1}^n\sqrt{2I_{\mathrm{loc},t}^{\MC}}
\Bigr\}.
\label{eq_onesided_MC}
\end{align}

Corollaries~\ref{cor:cvar_population_validation} and~\ref{cor:dro_population_validation} in Appendix~\ref{app:proof_population_validation_section} give the fixed-predictor plug-in terms
$\Ex[\UC_n^{\CVaR}(f_W)]=\calo(n^{-1/2})$ for CVaR and
$\Ex[\UC_n^{\DRO}(f_W)]=\widetilde{\mathcal O}(n^{-1/(c^\ast\vee2)})$ for CR DRO.
Both are applied after conditioning on $f_W$, so no additional uniform entropy over $\mathcal{F}$ is paid here.
  Appendix~\ref{app:comparison_summary} compares the full population--training bounds with existing results in Table~\ref{tab:full_objective_comparison}.

\section{Related Work}
\label{sec:related}

Information-theoretic generalization bounds relate expected generalization to the information that a learning algorithm reveals about the selected sample~\citep{Xu2017,steinke20a,hellstrom2022a,harutyunyan2021}.  Supersample and evaluated-CMI analyses  use the symmetry under \(u\mapsto\bar u\) for ordinary pointwise losses, where fixing the predictor also fixes the row-wise loss table. Refined single-index bounds further localize this information to individual selectors~\citep{wang23}.  Our setting differs because the empirical criterion first observes the whole batch and then solves an auxiliary optimization problem.  Theorem~\ref{thm:wang_general} extends the single-index viewpoint by replacing fixed per-example losses with  the increments of the optimized objective along the one-point replacement path. Existing work~\citep{wang23} suggests that shifted-Rademacher arguments can yield fast-rate improvements for standard eCMI bounds.  In the set-input regime, however, the telescoping identity~\eqref{eq:wang_telescoping_identity} does not generally lead to an analogous shifted decomposition.   Appendix~\ref{app:additional_refinements} discusses Bernstein bounds and compares $I_\mathrm{fCMI}$ and $I_{\mathrm{loc}}$.

Optimization-valued criteria also arise in machine learning in a different form: objectives in which the optimization is attached to a single input rather than to an evaluation set. Examples include predict-then-optimize, decision-focused learning, and data-driven optimization~\citep{wilder2019melding,mandi2022dfl,elmachtoub2022spo,iwata2025projection}. PAC-Bayesian guarantees for learned optimization algorithms have also recently appeared~\citep{sucker2024pacbayes}. These objectives are different from the set-input objectives studied here: under appropriate assumptions, their optimized value can be treated as a per-example loss. See Proposition~\ref{prop:single_input_ecmi}.

Risk-sensitive generalization bounds have been developed for CVaR and related OCE criteria~\citep{lee2020risksensitive}; the classical optimization formulation of CVaR goes back to \citet{rockafellar2000}. Information-theoretic analyses of calibration have also studied  the statistical bias of binning ECE~\citep{futami2024information}. Our framework instead treats MC, CVaR, and Cressie--Read DRO as set-input optimized objectives:  once Assumption~\ref{ass:selector_gap_subgaussian} has been verified, the validation--training term follows  by applying the standard fCMI inequality to the centered difference of the separately optimized values.

 Unlike set-input objectives, single-input optimization metrics solve an optimization problem for each data point. In the single-input regime, the downstream optimization is solved \emph{row by row}. Predict-then-optimize and other instance-wise learning-to-optimize problems fit this template: for each input row, the model predicts instance-specific quantities and a separate downstream optimization problem is solved using only that row \cite{elmachtoub2022spo,wilder2019melding,mandi2022dfl,iwata2025projection}.  Once the predictor is fixed, the optimized value behaves like an ordinary per-example loss and admits standard IT analysis; see Appendix~\ref{app:single_input_regime}.

\section{Numerical Bound Evaluation}
\label{sec:experiments}

 We numerically evaluate the proposed bounds and examine their usefulness.\footnote{Reproducibility code for all experiments in this paper is available at \url{https://github.com/msfuji0211/supersample-set-input-bound}.}

\subsection{Linear logistic regression on Flights200K}
\label{sec:flights-experiment}

We apply linear logistic regression to Flights200K~\citep{vegaFlights200K} to numerically evaluate the MC bound in Corollary~\ref{cor:mc_stability}. Experiments for the CVaR and CR DRO bounds in Corollaries~\ref{cor:cvar_l2} and~\ref{cor:dro_l2} under the same settings are reported in Appendices~\ref{app:flights-cvar} and~\ref{app:flights-crdro}, respectively.
We train an $L_2$-regularized linear logistic regression model and add independent Gaussian noise to the fitted parameters, including the intercept, as $W=\widehat\theta(S)+\sigma\xi$, with $\xi\sim N(0,I_8)$ and $\sigma\in\{0.05,0.10\}$.
For each sampled pair index, we retrain under both values of the selector while holding the other selectors fixed.
The resulting Gaussian parameter distributions allow the conditional mutual information between the parameters and the selector to be evaluated by one-dimensional integration.
Combining this with data processing gives an estimate of the single-index bound in Eq.~\eqref{eq:mc_wang_bound_main}.
We compare it with the absolute value of the average validation--training MC difference.
 Appendix~\ref{app:flights-settings} gives the logistic-regression settings, subgroup/bin construction, and Monte Carlo estimation procedure.

\begin{figure}[t]
    \centering
    \includegraphics[width=\linewidth]{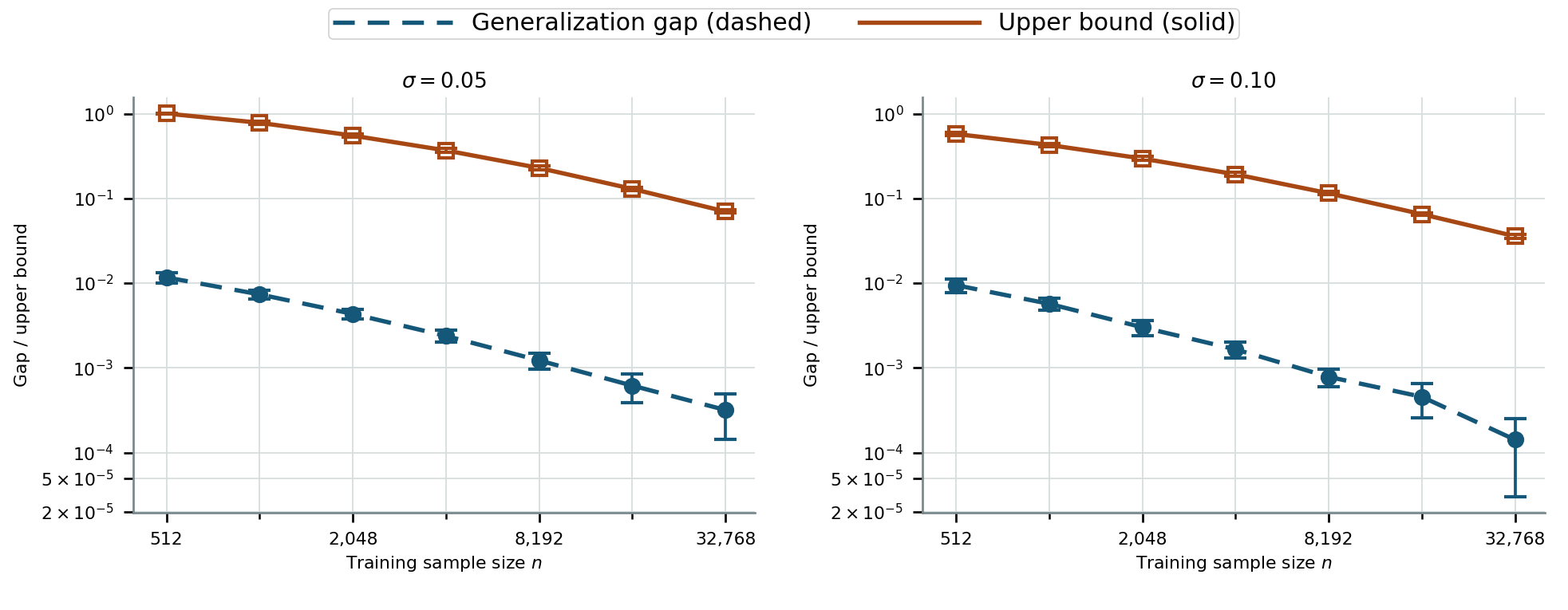}
    \caption{  MC gap (dashed) and Gaussian bound (solid) on Flights200K: $\sigma=0.05$ (left), $0.10$ (right). Points and error bars show the mean $\pm$ one sample standard deviation of independent estimates.}
    \label{fig:flights-bound}
\end{figure}

Figure~\ref{fig:flights-bound} shows the mean and one sample standard deviation from independent Monte Carlo estimates.
For both noise levels, the gap and bound estimates decrease with the sample size, and the bound is approximately linear on the log--log scale.
The bound estimates exceed the gap estimates in every setting and remain conservative, but  fall below the trivial bound of one for larger samples.

\Needspace{4\baselineskip}
\subsection{Large-scale image classification on CelebA}
\label{sec:celeba-experiment}

\begin{figure}[t]
    \centering
    \includegraphics[width=1.\linewidth]{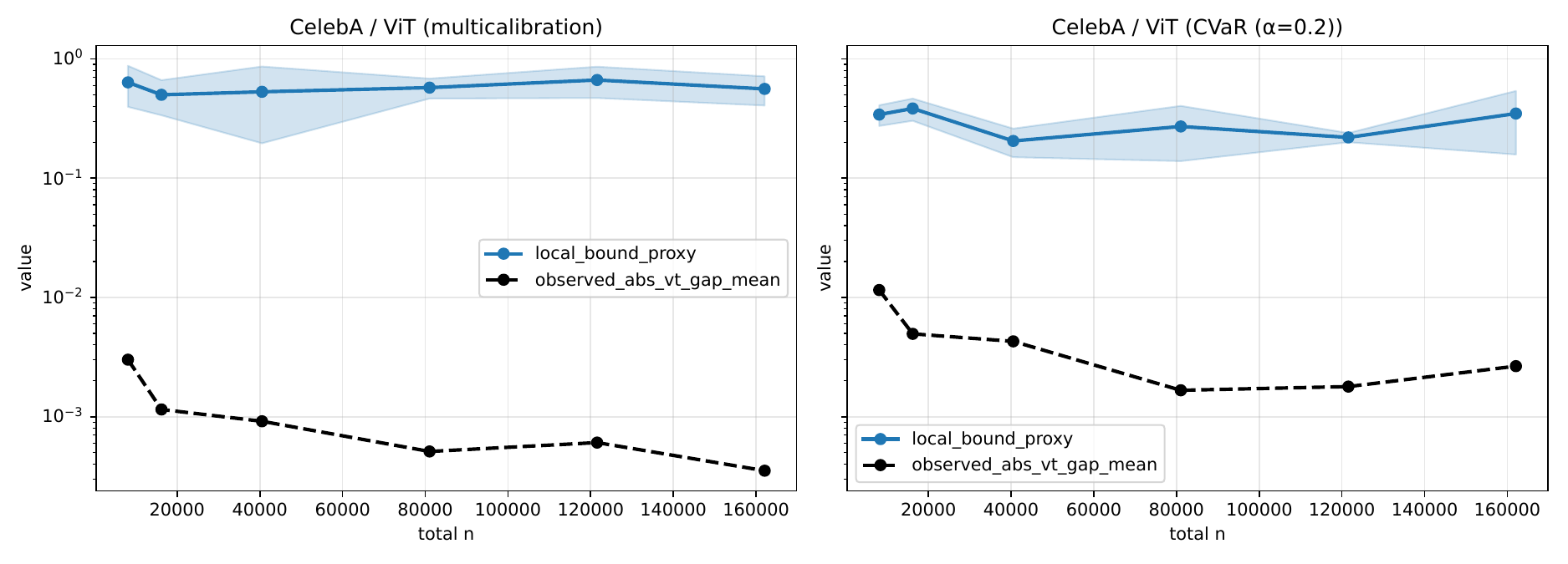}
    \caption{ Observed absolute validation--training gap and estimated local-information proxy on \texttt{CelebA}/\texttt{ViT}: MC (left) and CVaR ($\alpha=0.2$; right). Shaded bands show variation across supersample seeds. The proxy remains above the observed gap in both panels.}
    \label{fig:bound}
\end{figure}

We next evaluate the bounds for a larger model on image data, using \texttt{ViT} on \texttt{CelebA} for MC and CVaR ($\alpha=0.2$). The MC protocol follows \citet{hansen2024multicalibration}; Appendix~\ref{app:celeba-settings} gives the settings. As in Section~\ref{sec:flights-experiment}, we limit computation by Monte Carlo sampling over supersamples, data-pair indices, and training randomness. We sample ten pairs per supersample and repeat training seven times for each choice within a pair, holding the other selectors fixed. Unlike the Gaussian-perturbed logistic model, the conditional distribution of ViT's trained parameters is not explicit, so the Gaussian information formula is unavailable. Following the nearest-neighbor estimation approach of \citet{futami2024information}, we instead estimate the mutual information between local increments and their corresponding selectors from repeated training. This information additionally conditions on the other selectors. Finite-sample information estimates can be biased, and their aggregation need not yield an upper bound; we therefore call the result a proxy rather than a bound.

Figure~\ref{fig:bound} compares this estimate with the observed absolute validation--training gap. For both objectives, the estimate is much larger than the observed gap throughout the range of sample sizes.

\paragraph{Discussion.}
In logistic regression, Gaussian perturbations make the conditional parameter distributions explicit, allowing the information upper bound to be evaluated accurately by one-dimensional integration. For ViT, nearest-neighbor information estimation from repeated training gives a proxy much larger than the observed gap. Evaluating the proposed bounds for larger models therefore requires more accurate and computationally efficient information estimators.

\section{Conclusion and Limitations}
\label{sec:discussion}
We introduced a unified IT framework for \emph{set-input optimization-valued objectives},
a class that includes MC, CVaR, and Cressie--Read DRO. The main message is a separation: fCMI isolates algorithmic dependence on the supersample selector vector,  while the objective-dependent sub-Gaussian parameter reflects the geometry of the optimized objective. We bound this parameter through one-point stability, which also yields a single-index refinement, and through convex \(L_2\)-Lipschitz geometry, which gives the sharper DRO rate. Our theory has several limitations. First, our theory isolates statistical generalization and does not analyze the optimization error of the outer learning problem. Although the single-index refinement isolates one selector at a time, it can still depend on the full prediction table rather than a single row. 
The centering argument uses equal-size training and validation samples; the finite-class restriction in the smaller-validation-set extension instead comes from a union bound and can be relaxed under suitable conditions (Appendix~\ref{app:small_validation_subset_added}).
Future work includes  sharper bounds on the sub-Gaussian parameter, tighter fixed-predictor comparisons, fast-rate refinements, and optimization-aware analyses for other batch-level criteria.  

\clearpage

\begin{ack}
 FF was supported by JST BOOST Program Grant Number JPMJBY24G8. FF and MF were supported by JSPS KAKENHI Grant Number 26K02974, Japan. MF was supported by JSPS KAKENHI Grant Number JP25K21286, Japan. This work was supported by Japan Science and Technology Agency (JST) as part of Adopting Sustainable Partnerships for Innovative Research Ecosystem (ASPIRE), Grant Number JPMJAP25B1.
\end{ack}

\bibliographystyle{plainnat}
\bibliography{main}

\clearpage
\appendix

\section*{Notation}
\begingroup
\small
\setlength{\tabcolsep}{4pt}
\renewcommand{\arraystretch}{1.16}
\noindent
\begin{tabular}{@{}p{0.19\linewidth}p{0.57\linewidth}p{0.19\linewidth}@{}}
\toprule
\textbf{Symbol} & \textbf{Meaning} & \textbf{Defined in} \tabularnewline
\midrule
\multicolumn{3}{@{}l}{\textbf{Theoretical quantities}} \tabularnewline
\(\mathcal D,n,\widetilde Z\) & \raggedright Data distribution, size of each half, and supersample of \(n\) pairs of i.i.d. observations. & \raggedright Section~\ref{sec:ordinary_it} \tabularnewline
\(U,\overline U\) & \raggedright Selector vector and its coordinatewise complement. & \raggedright Section~\ref{sec:ordinary_it} \tabularnewline
\(Q_U\) & \raggedright Uniform reference distribution on \(\{0,1\}^n\). & \raggedright Assumption~\ref{ass:selector_gap_subgaussian} \tabularnewline
\(S,S'\) & \raggedright Training sample \(\widetilde Z_U\) and validation sample \(\widetilde Z_{\overline U}\). & \raggedright Section~\ref{sec:ordinary_it} \tabularnewline
\(R,W,f_W\) & \raggedright Independent learner randomness, output \(W=A(S,R)\), and learned predictor. & \raggedright Section~\ref{sec:ordinary_it} \tabularnewline
\(f_W(\widetilde X)\) & \raggedright Predictions on all \(2n\) supersample inputs. & \raggedright Section~\ref{sec:ordinary_it} \tabularnewline
\(\Pi,\pi\) & \raggedright Auxiliary optimization class and a candidate in that class. & \raggedright Section~\ref{sec:general_set_input} \tabularnewline
\(\Phi_n,\Phi\) & \raggedright Optimized empirical objective and its population counterpart. & \raggedright Eq.~\eqref{eq:set_input_avg_def} \tabularnewline
\(\Gamma_{f,\widetilde z}(u)\) & \raggedright Selector gap for fixed predictor and supersample. & \raggedright Eq.~\eqref{eq:selector_gap_main} \tabularnewline
\(\Delta_\Phi\) & \raggedright Validation--training gap \(\Phi_n(f_W;S')-\Phi_n(f_W;S)\) for the learned predictor. & \raggedright Section~\ref{sec:main} \tabularnewline
\(I_{\mathrm{fCMI}}\) & \raggedright Global information \(I(f_W(\widetilde X);U\mid\widetilde Z)\). & \raggedright Section~\ref{sec:main} \tabularnewline
\(S^{(t,b)},D_t\) & \raggedright Mixed-path sample and local increment. & \raggedright Theorem~\ref{thm:wang_general} \tabularnewline
\(I_{\mathrm{loc},t}\) & \raggedright Local information \(I(D_t;U_t\mid\widetilde Z)\). & \raggedright Theorem~\ref{thm:wang_general} \tabularnewline
\midrule
\multicolumn{3}{@{}l}{\textbf{Objective-dependent quantities}} \tabularnewline
\(L_{\Gamma,n}\) & \raggedright Sub-Gaussian parameter for the selector gap. & \raggedright Assumption~\ref{ass:selector_gap_subgaussian} \tabularnewline
\(\beta_n\) & \raggedright One-point stability coefficient. & \raggedright Assumption~\ref{ass:one_point_stability} \tabularnewline
\(L_{\psi,n}\) & \raggedright Euclidean Lipschitz constant of \(\tilde\psi\). & \raggedright Assumption~\ref{ass:convex_l2_score_vector_main} \tabularnewline
\(\UC_n^\Phi(f)\) & \raggedright Fixed-predictor population--validation error. & \raggedright Section~\ref{sec:population_validation_terms} \tabularnewline
\midrule
\multicolumn{3}{@{}l}{\textbf{Experimental estimates}} \tabularnewline
\(\widehat L_n,\widehat B_{G,n}\) & \raggedright Flights200K gap estimate and Gaussian-bound estimate, respectively. & \raggedright Appendix~\ref{app:flights-settings} \tabularnewline
\(\widehat{\mathcal B}_{\Phi}\) & \raggedright CelebA information proxy averaged over sampled coordinates and supersample configurations. & \raggedright Appendix~\ref{app:celeba-settings} \tabularnewline
\(\widehat I_{j,t}^{\mathrm{LOO}}\) & \raggedright Nearest-neighbor MI estimate with supersample configuration \(j\) and \(U_{-t}\) fixed. & \raggedright Appendix~\ref{app:celeba-settings} \tabularnewline
\bottomrule
\end{tabular}
\endgroup

Throughout the Appendix, we write \(D(P\|Q)\) for the Kullback--Leibler divergence between two probability measures \(P\) and \(Q\).

\section{Detailed explanation about examples of set-input metrics}
\label{app:reliability_examples}

This appendix records the reliability- and risk-oriented objectives used in the paper.  We keep only the definitions and background needed for the later proofs.  Throughout, $Z=(X,Y)\sim\cald$, $S=(Z_i)_{i=1}^n$, and $V_i=f_w(X_i)$.

\subsection{Multicalibration}
\label{app:reliability_multicalibration}

Calibration asks that a score predictor $f_w:\calx\to[0,1]$ be interpretable as a probability, ideally satisfying $\Ex[Y\mid f_w(X)]=f_w(X)$ almost surely~\citep{foster1998asymptotic}.  Multicalibration strengthens this requirement by testing calibration simultaneously over many subgroups and score witnesses~\citep{johnson18a,gopalan22alow,blasiok2023unify}.  This is useful when reliability on overlapping or computationally identifiable population slices matters, for example in fairness-aware risk scoring and downstream decision support.

For a subgroup class $\cC\subseteq\{c:\calx\to[0,1]\}$ and a score-test class $\cH\subseteq\{h:[0,1]\to[-1,1]\}$, set $\Pi_{\MC}\coloneqq\cC\times\cH\times\{\pm1\}$.  The population and empirical multicalibration metrics are
\begin{align}
\Phi_{\MC}(f_w;\cald)
&\coloneqq
\sup_{(c,h,\eta)\in\Pi_{\MC}}
\eta\,\Ex\big[c(X)h(f_w(X))(Y-f_w(X))\big],
\label{eq:app_mc_population_v3}
\\
\Phi_{\MC}(f_w;S)
&\coloneqq
\sup_{(c,h,\eta)\in\Pi_{\MC}}
\frac{\eta}{n}\sum_{i=1}^n c(X_i)h(V_i)(Y_i-V_i).
\label{eq:app_mc_empirical_v3}
\end{align}
The empirical optimizer is the witness triple $(c,h,\eta)$ chosen after observing the whole batch.

Typical subgroup classes are indicators $\cC_{\cG}=\{x\mapsto\one\{x\in G\}:G\in\cG\}$, where $\cG$ may be a finite protected-group collection, a VC or low-degree class, or an oracle-efficient benchmark family~\citep{johnson18a,gopalan22alow,noarov23a}.  Common witness classes include bin indicators $v\mapsto\one\{v\in I_j\}$, bounded measurable witnesses, Lipschitz witnesses, and RKHS unit balls; these choices recover bin-style, smooth, or kernel calibration distances~\citep{foster2018smooth,blasiok2023unify,hartline_et_al_2025}.  Algorithmically, multicalibration has been studied through iterative post-processing, weak-agnostic-learning reductions, boosting-style procedures, and game-dynamics or privacy-based approaches~\citep{johnson18a,gopalan22alow,harris23aMul,noarov23a,Haghtalab2023}.

\subsection{OCE risks and CVaR}
\label{app:reliability_oce}

Optimized-certainty-equivalent (OCE) risks model tail sensitivity or disutility-sensitive preferences~\citep{bentalteboulle2007,rockafellar2000}.  They are used when a small mean loss may still hide rare but severe failures, for example in finance, medical decision support, safe learning, and risk-sensitive reinforcement learning~\citep{lee2020risksensitive,wang2025ocerl}.

Let $L_w\coloneqq \ell(f_w(X),Y)\in[0,1]$ and let $\psi:\R\to\R\cup\{\infty\}$ be a proper closed convex disutility.  The population and empirical OCE risks are
\begin{align}
\Phi_{\OCE,\psi}(f_w;\cald)
&\coloneqq
\inf_{\eta\in\R}\Bigl\{\eta+\Ex\,\psi(L_w-\eta)\Bigr\},
\label{eq:app_oce_population_v3}
\\
\Phi_{\OCE,\psi}(f_w;S)
&\coloneqq
\inf_{\eta\in\R}\Bigl\{\eta+\frac1n\sum_{i=1}^n \psi\bigl(\ell(f_w,Z_i)-\eta\bigr)\Bigr\}.
\label{eq:app_oce_empirical_v3}
\end{align}
Thus the threshold $\eta$ is selected after seeing the batch of losses.

For CVaR at level $\alpha\in(0,1)$, write $\bar\alpha\coloneqq1-\alpha$; the disutility is $\psi_\alpha(t)=\bar\alpha^{-1}[t]_+$.  Hence
\begin{align}
\Phi_{\CVaR,\alpha}(f_w;\cald)
&\coloneqq
\inf_{\tau\in[0,1]}\Bigl\{\tau+\frac{1}{\bar\alpha}\Ex(L_w-\tau)_+\Bigr\},
\label{eq:app_cvar_population_v3}
\\
\Phi_{\CVaR,\alpha}(f_w;S)
&\coloneqq
\inf_{\tau\in[0,1]}\Bigl\{\tau+\frac{1}{\bar\alpha n}\sum_{i=1}^n (\ell(f_w,Z_i)-\tau)_+\Bigr\}.
\label{eq:app_cvar_empirical_v3}
\end{align}
This is the usual upper-tail risk.  If the loss distribution has no atom at an optimizing quantile, it equals the conditional mean over the worst $\bar\alpha$ tail; with atoms, the risk-envelope formula below is the unambiguous definition.  Empirical OCE objectives and their stochastic approximations have been studied in supervised and reinforcement-learning settings~\citep{lee2020risksensitive,wang2025ocerl}.

When $\psi$ is proper, closed, convex, nondecreasing, and normalized by $\psi(0)=0$, the duality condition (satisfied by the CVaR disutility and by the bounded OCEs used here) holds and it gives the dual representation
\begin{align}
\Phi_{\OCE,\psi}(f_w;\cald)
=
\sup\Bigl\{\Ex[q(Z)L_w]-\Ex[\psi^*(q(Z))]:
q(Z)\ge 0\ \text{a.s.},\ \Ex q(Z)=1\Bigr\}.
\label{eq:app_oce_dual_v3}
\end{align}
Here the nonnegativity of $q$ follows from the monotonicity of $\psi$, equivalently from the effective domain of $\psi^*$. For CVaR this reduces to the risk-envelope form
\begin{align}
\Phi_{\CVaR,\alpha}(f_w;\cald)
&=
\sup\Bigl\{\Ex[q(Z)L_w]:0\le q(Z)\le\bar\alpha^{-1}\ \text{a.s.},\ \Ex q(Z)=1\Bigr\},
\label{eq:app_cvar_dro_v3}
\\
\Phi_{\CVaR,\alpha}(f_w;S)
&=
\sup\Bigl\{\frac1n\sum_{i=1}^n q_i\ell(f_w,Z_i):0\le q_i\le\bar\alpha^{-1},\ \frac1n\sum_{i=1}^nq_i=1\Bigr\}.
\label{eq:app_cvar_dro_empirical_v3}
\end{align}
The risk-envelope form is used to expose convex \(L_2\) geometry and for common-feasible-set comparisons; the primal threshold form remains the average-form set-input representation used for CVaR.

\subsection{\texorpdfstring{$f$-divergence DRO and the Cressie--Read family}{f-divergence DRO and the Cressie--Read family}}
\label{app:reliability_fdro}

For a nominal law $P=\Law(Z)$ and an alternative law $Q\ll P$, write $\omega=dQ/dP$.  An $f$-divergence ball is specified by $D_f(Q\|P)=\Ex_P f(\omega(Z))$ and $\Ex_P \omega(Z)=1$.  The robust risk is
\begin{align}
\Phi_{f\text{-}\DRO,\epsilon}(f_w;\cald)
\coloneqq
\sup_{\omega\ge0}\Bigl\{
\Ex_{\cald}\!\big[\omega(Z)\ell(f_w(X),Y)\big]:
\Ex_{\cald}\omega(Z)=1,
\Ex_{\cald}f(\omega(Z))\le\epsilon
\Bigr\}.
\label{eq:app_fdro_population_short}
\end{align}
For the Cressie--Read generator \(f_c(t)=(t^c-ct+c-1)/\{c(c-1)\}\), \(c\in\R\setminus\{0,1\}\), the case $c=2$ is Pearson $\chi^2$, while the limits $c\to1$ and $c\to0$ recover KL and empirical-likelihood divergences.  For $c>1$, the constraints $\Ex f_c(\omega)\le\epsilon$ and $\Ex \omega=1$ are equivalent to $\Ex \omega(Z)^c\le1+c(c-1)\epsilon$.

For the empirical law $\widehat P_S=n^{-1}\sum_{i=1}^n\delta_{Z_i}$, the density ratio is a weight vector $q=(q_1,\ldots,q_n)$ with $n^{-1}\sum_iq_i=1$.  The empirical Cressie--Read robust value is
\begin{align}
\Phi_{\DRO,c,\epsilon}(f_w;S)
=
\sup_{q\ge0}\Bigl\{
\frac1n\sum_{i=1}^nq_i\ell(f_w(X_i),Y_i):
\frac1n\sum_{i=1}^nq_i=1,
\frac1n\sum_{i=1}^nq_i^c\le 1+c(c-1)\epsilon
\Bigr\}.
\label{eq:app_cr_dro_empirical_short}
\end{align}
The auxiliary optimizer is therefore an adversarial reweighting chosen after the empirical loss vector is observed.

\subsection{Cressie--Read DRO as an average-form objective}
\label{app:cr-dual-average}

The primal reweighting form above is useful for Euclidean Lipschitz bounds.  The following standard finite-dimensional Cressie--Read dual form places the same objective in the average-form class \eqref{eq:set_input_avg_def}; compare the Cressie--Read plug-in formulation of \citet{Duchi2021}.
\begin{lemma}[Cressie--Read dual average form]
\label{lem:cr_dual_average_form}
Let $c>1$, $c^\ast=c/(c-1)$, and
\[
\mathcal Q_{n,c,\epsilon}
\coloneqq
\Bigl\{q\in\mathbb R_+^n:
\frac1n\sum_{i=1}^n q_i=1,
\frac1n\sum_{i=1}^n q_i^c\le \kappa_{c,\epsilon}^c\Bigr\},
\qquad
\kappa_{c,\epsilon}=\{1+c(c-1)\epsilon\}^{1/c}.
\]
For bounded losses $r_i=\ell(f(X_i),Y_i)$,
\[
\sup_{q\in\mathcal Q_{n,c,\epsilon}}\frac1n\sum_{i=1}^n q_i r_i
=
\inf_{\eta\in\mathbb R}
\Biggl\{
\eta+
\kappa_{c,\epsilon}
\left(\frac1n\sum_{i=1}^n(r_i-\eta)_+^{c^\ast}\right)^{1/c^\ast}
\Biggr\}.
\]
\end{lemma}
See \citet{Duchi2021} for its derivation.

Consequently, \(\Phi_{\DRO,c,\epsilon,n}(f;S)\) has the average-form representation
\[
\operatorname{Opt}_{\eta\in\mathbb R}
\psi_\eta\left(\frac1n\sum_{i=1}^n \rho_\eta(f(X_i),Z_i)\right),
\qquad
\operatorname{Opt}=\inf,
\]
with $\rho_\eta(v,z)=\bigl(\ell(v,z)-\eta\bigr)_+^{c^\ast}$ and $\psi_\eta(m)=\eta+\kappa_{c,\epsilon}m^{1/c^\ast}$.

\section{Auxiliary inequalities for supremum and infimum objectives}
\label{app:opt_inequalities}

We collect two elementary inequalities used repeatedly in the paper. For arbitrary families $(a_{\pi})_{\pi\in\Pi}$ and $(b_{\pi})_{\pi\in\Pi}$,
\begin{align}
\bigl|\sup_{\pi\in\Pi} a_{\pi}-\sup_{\pi\in\Pi} b_{\pi}\bigr|
&\le \sup_{\pi\in\Pi}|a_{\pi}-b_{\pi}|,
\label{eq:sup_diff_aux}
\\
\bigl|\inf_{\pi\in\Pi} a_{\pi}-\inf_{\pi\in\Pi} b_{\pi}\bigr|
&\le \sup_{\pi\in\Pi}|a_{\pi}-b_{\pi}|.
\label{eq:inf_diff_aux}
\end{align}
These give the absolute-gap population--validation comparison for both supremum and infimum objectives. For one-sided bounds, fix a predictor $f$ and suppose the population and empirical inner objectives share a common feasible set $\Pi(f)$. If \(\psi_\pi(f,\cald)\le \Ex[\psi_{\pi,n}(f,S')\mid f]
\qquad\text{for all }\pi\in\Pi(f),\) then
\begin{align}
\sup_{\pi\in\Pi(f)}\psi_\pi(f,\cald)
&\le
\Ex\Bigl[\sup_{\pi\in\Pi(f)}\psi_{\pi,n}(f,S')\,\Big|\,f\Bigr].
\label{eq:sup_one_sided_aux}
\end{align}
Thus supremum-type objectives naturally yield upper bounds on $\Phi(f;\cald)-\Phi(f;S)$.

\section{Discussion about single-input optimization-valued objectives}
\label{app:single_input_regime}

This appendix records the simple reduction mentioned in Section~\ref{sec:set_input_prelim}. In the single-input regime, the downstream optimization is solved \emph{row by row}. Predict-then-optimize and other instance-wise learning-to-optimize problems fit this template: for each input row, the model predicts instance-specific quantities and a separate downstream optimization problem is solved using only that row \cite{elmachtoub2022spo,wilder2019melding,mandi2022dfl,iwata2025projection}. Once the predictor is fixed, the optimized value behaves exactly like an ordinary per-example loss.

Let \(\phi_1:\calx\times\caly\times\calv\times\Pi\to\R\) be a measurable objective, and let \(\Pi_1(x,y,v)\subseteq\Pi\) be the feasible set for one example. For a predictor \(f_w\) and one data point \((X,Y)\), define the optimized single-input loss by
\begin{align}
\phi_1^{f_w}(X,Y)
\coloneqq 
\operatorname{Opt}_{\pi\in\Pi_1(X,Y,f_w(X))}
\phi_1(X,Y,f_w(X);\pi),
\label{eq:single_input_value}
\end{align}
where \(\operatorname{Opt}\in\{\sup,\inf\}\). Throughout this appendix, the optimized value is understood to be a finite measurable version whenever it is used; in particular, the relevant feasible sets are assumed to be nonempty.

The corresponding empirical objective is the ordinary average
\[
\hat R^{\rm opt}(f_w;S)
\coloneqq
\frac1n\sum_{i=1}^n \phi_1^{f_w}(X_i,Y_i).
\]

The important point is that once \(f_W\) is fixed, the optimized quantity \(\phi_1^{f_W}(\widetilde{X}_{i,j},\widetilde{Y}_{i,j})\) is just an ordinary supersample loss. Hence, under boundedness or sub-Gaussian assumptions on \(\phi_1^{f_W}\), the standard eCMI bound \eqref{eq:standard_ecmi} applies exactly as in ordinary supervised learning. Although the optimized loss may depend on \(x\) as well as \(y\), the eCMI argument only uses the induced bounded supersample loss table.

\begin{proposition}[Single-input optimization-valued losses reduce to ordinary eCMI]
\label{prop:single_input_ecmi}
Assume that, for every \(w\) in the support of \(W\), the feasible set \(\Pi_1(x,y,f_w(x))\) is nonempty and the optimized per-example value \(\phi_1^{f_w}(x,y)\) from \eqref{eq:single_input_value} admits a finite measurable version lying in \([0,1]\). Set \(\phi_{i,j}
\coloneqq
\phi_1^{f_W}(\widetilde{X}_{i,j},\widetilde{Y}_{i,j}),
\qquad i\in[n],\ j\in\{0,1\},\) and write \(\widetilde V_W
\coloneqq
\bigl(f_W(\widetilde X_{i,j})\bigr)_{i\in[n],\,j\in\{0,1\}}\) for the supersample prediction array. Then the row-wise eCMI bound
\begin{align}
\left|\Ex\bigl[\hat R^{\mathrm{opt}}(f_W;S')-\hat R^{\mathrm{opt}}(f_W;S)\bigr]\right|
&\le
\frac1n\sum_{i=1}^n
\sqrt{2I((\phi_{i,0},\phi_{i,1});U_i\mid\widetilde{Z})}
\label{eq:single_input_row_ecmi}
\end{align}
holds. In addition,  the following bound uses the CMI between the loss table and the full vector \(U\), given \(\widetilde Z\), and hence also gives the corresponding fCMI bound:
\begin{equation}
\left|
\Ex\bigl[\hat R^{\mathrm{opt}}(f_W;S')-\hat R^{\mathrm{opt}}(f_W;S)\bigr]
\right|
\le
\sqrt{\frac{2\,I((\phi_{i,j})_{i,j};U\mid \widetilde{Z})}{n}}
\le
\sqrt{\frac{2\,I(\widetilde V_W;U\mid \widetilde{Z})}{n}}.
\label{eq:single_input_ecmi}
\end{equation}
If \(W\) is independent of \((\widetilde Z,U)\), in particular if \(f_w\) is fixed independently of the supersample and selector vector, then \(\Ex[\hat R^{\mathrm{opt}}(f_W;S')-\hat R^{\mathrm{opt}}(f_W;S)]=0.\) \end{proposition}

\begin{proof}
Because the optimization in \eqref{eq:single_input_value} is solved separately for each example, the array \(\Phi\coloneqq(\phi_{i,j})_{i\in[n],\,j\in\{0,1\}}\) is an ordinary bounded supersample loss table. Therefore
\[
\hat R^{\mathrm{opt}}(f_W;S)
=
\frac1n\sum_{i=1}^n \phi_{i,U_i},
\qquad
\hat R^{\mathrm{opt}}(f_W;S')
=
\frac1n\sum_{i=1}^n \phi_{i,1-U_i}.
\]
Applying the standard row-wise eCMI inequality \eqref{eq:standard_ecmi} to this bounded loss table, and applying the same argument to the negative gap, proves \eqref{eq:single_input_row_ecmi}.

 It remains to prove the second bound. Fix a deterministic loss table \(\phi=(\phi_{i,j})_{i\in[n],\,j\in\{0,1\}}\in[0,1]^{n\times 2},\) and define
\[
\Gamma_\phi(u)
\coloneqq
\frac1n\sum_{i=1}^n \phi_{i,1-u_i}
-
\frac1n\sum_{i=1}^n \phi_{i,u_i}.
\]
Writing \(d_i\coloneqq \phi_{i,1}-\phi_{i,0},\) we have \(\Gamma_\phi(u)
=
\frac1n\sum_{i=1}^n (1-2u_i)d_i.\) Under the uniform selector law \(Q_U=\mathrm{Unif}(\{0,1\}^n)\), the random variables \((1-2U_i)d_i/n,
\qquad i\in[n],\) are independent and mean-zero. Hence Hoeffding's lemma gives, for every \(\lambda\in\mathbb R\),
\[
\log\Ex_{Q_U}\exp\{\lambda\Gamma_\phi(U)\}
\le
\frac{\lambda^2}{2n^2}\sum_{i=1}^n d_i^2
\le
\frac{\lambda^2}{2n},
\]
where the last inequality uses \(\phi_{i,j}\in[0,1]\), so that \(|d_i|\le1\).

Now apply the conditional change-of-measure argument with reference law \(P_{\Phi\mid\widetilde Z}Q_U.\) For every \(\lambda>0\), \(\Ex\Gamma_\Phi(U)
\le
\frac{1}{\lambda}
I(\Phi;U\mid\widetilde Z)
+
\frac{\lambda}{2n}.\) Optimizing over \(\lambda>0\) yields \(\Ex\Gamma_\Phi(U)
\le
\sqrt{\frac{2I(\Phi;U\mid\widetilde Z)}{n}}.\) Applying the same argument to \(-\Gamma_\Phi(U)\) gives \(\left|\Ex\Gamma_\Phi(U)\right|
\le
\sqrt{\frac{2I(\Phi;U\mid\widetilde Z)}{n}}.\) Since \(\Gamma_\Phi(U)
=
\hat R^{\mathrm{opt}}(f_W;S')
-
\hat R^{\mathrm{opt}}(f_W;S),\) this proves the first inequality in \eqref{eq:single_input_ecmi}.

For the second inequality in \eqref{eq:single_input_ecmi}, observe that conditional on \(\widetilde Z\), the optimized loss table \(\Phi\) is a measurable function of the supersample prediction array \(\widetilde V_W\). Indeed,
\[
\phi_{i,j}
=
\operatorname{Opt}_{\pi\in\Pi_1(\widetilde X_{i,j},\widetilde Y_{i,j},f_W(\widetilde X_{i,j}))}
\phi_1(\widetilde X_{i,j},\widetilde Y_{i,j},f_W(\widetilde X_{i,j});\pi).
\]
Therefore conditional data processing gives \(I(\Phi;U\mid\widetilde Z)
\le
I(\widetilde V_W;U\mid\widetilde Z),\) which proves the fCMI bound.

Finally, suppose that \(W\) is independent of \((\widetilde Z,U)\). Then, conditional on \((W,\widetilde Z)\), the optimized loss table \(\Phi\) is fixed and \(U\) is uniform. Hence \(\Ex[\Gamma_\Phi(U)\mid W,\widetilde Z]=0.\) Taking expectation gives \(\Ex[\hat R^{\mathrm{opt}}(f_W;S')-\hat R^{\mathrm{opt}}(f_W;S)]
=
0.\) \end{proof}

\paragraph{Specialization to pointwise losses.}
Take \(\Phi_n=\widehat L\) and \(\phi_{i,j}=\ell(f_W,\widetilde Z_{i,j})\in[0,1]\) in the preceding proof.
Its fixed-table MGF bound gives \(L_{\Gamma,n}\le n^{-1/2}\), so Eq.~\eqref{eq:signed_val_train_gap} gives
\(\lvert\Ex\Delta_\Phi\rvert\le\sqrt{2I_{\mathrm{fCMI}}/n}\), with the constant in Eq.~\eqref{eq:standard_ecmi}.
Replacing one observation changes the empirical risk by at most \(\beta_n=1/n\), and the other rows cancel in \(D_t=(\phi_{t,0}-\phi_{t,1})/n\).
Conditional data processing therefore gives
\(I_{\mathrm{loc},t}\le I((\phi_{t,0},\phi_{t,1});U_t\mid\widetilde Z)\).
Substitution into Eq.~\eqref{eq:wang_min_bound} yields the row-wise eCMI bound in Eq.~\eqref{eq:standard_ecmi}.

\section{Proofs for Section~\ref{sec:main}}
\label{app:section3_proofs}

\subsection{  Proofs for Section~\ref{sec:general_set_input}: fCMI bound}
\label{app:proof_general_set_input}

\begin{lemma}[ Centering by complementary splits]
\label{lem:uniform_selector_pairing_app}
Fix a predictor \(f\) and a supersample realization \(\widetilde{z}\). Define
\begin{align}
\Gamma_{f,\widetilde{z}}(u)
\coloneqq 
\Phi_n(f;\widetilde{z}_{\overline{u}})-
\Phi_n(f;\widetilde{z}_u),
\qquad u\in\{0,1\}^n.
\label{eq:selector_gap_fixed_app}
\end{align}
Then
\begin{align}
\Gamma_{f,\widetilde{z}}(\overline{u})=-\Gamma_{f,\widetilde{z}}(u)
\qquad\text{for all }u\in\{0,1\}^n.
\label{eq:selector_gap_sign_flip_app}
\end{align}
Consequently, if \(U\sim Q_U=\mathrm{Unif}(\{0,1\}^n)\), then
\begin{align}
\Ex_{Q_U}\Gamma_{f,\widetilde{z}}(U)
&=
2^{-n}\sum_{u\in\{0,1\}^n}\Gamma_{f,\widetilde{z}}(u)
\notag\\
&=
2^{-n-1}\sum_{u\in\{0,1\}^n}
\bigl\{\Gamma_{f,\widetilde{z}}(u)+\Gamma_{f,\widetilde{z}}(\overline{u})\bigr\}
=0.
\label{eq:pairing_identity_zero_mean_app}
\end{align}
\end{lemma}

\begin{proof}
First, the sign flip follows by substituting \(\overline{u}\) into the definition of the gap. Since \(\overline{\overline{u}}=u\),
\begin{align*}
\Gamma_{f,\widetilde{z}}(\overline{u})
&=
\Phi_n(f;\widetilde{z}_{\overline{\overline{u}}})-
\Phi_n(f;\widetilde{z}_{\overline{u}})\\
&=
\Phi_n(f;\widetilde{z}_u)-
\Phi_n(f;\widetilde{z}_{\overline{u}})
=
-\Gamma_{f,\widetilde{z}}(u),
\end{align*}
This proves \eqref{eq:selector_gap_sign_flip_app}.

We now prove the conditional centering identity by  pairing each \(u\) with \(\bar u\). Since \(U\) is uniform on \(\{0,1\}^n\),
\begin{align*}
\Ex_{Q_U}\Gamma_{f,\widetilde{z}}(U)
&=2^{-n}\sum_{u\in\{0,1\}^n}\Gamma_{f,\widetilde{z}}(u).
\end{align*}
The complement map \(u\mapsto \overline{u}\) is a bijection on \(\{0,1\}^n\). Hence the same finite average can be written as the average over complementary pairs:
\begin{align*}
2^{-n}\sum_{u\in\{0,1\}^n}\Gamma_{f,\widetilde{z}}(u)
&=
2^{-n-1}\sum_{u\in\{0,1\}^n}
\bigl\{\Gamma_{f,\widetilde{z}}(u)+\Gamma_{f,\widetilde{z}}(\overline{u})\bigr\}.
\end{align*}
By the sign-flip identity \eqref{eq:selector_gap_sign_flip_app}, the term inside braces is zero for every \(u\). This proves \eqref{eq:pairing_identity_zero_mean_app}.
\end{proof}

\begin{remark}[Pairing when \(n=2\)]
When \(n=2\), the selector vector takes the four values \(00,01,10,11\). These form the two complementary pairs \(00\leftrightarrow 11\) and \(01\leftrightarrow 10\). For each fixed \((f,\widetilde{z})\), complementing the selector vector swaps the two selected batches in the difference, so \(\Gamma_{f,\widetilde{z}}(11)=-\Gamma_{f,\widetilde{z}}(00),
\qquad
\Gamma_{f,\widetilde{z}}(10)=-\Gamma_{f,\widetilde{z}}(01).\) Thus the average over the four selector vectors cancels pairwise. The proof of Lemma~\ref{lem:uniform_selector_pairing_app} is exactly the same argument on \(\{0,1\}^n\).
\end{remark}

\begin{proof}[Proof of Theorem~\ref{thm:validation_training_gap}]
Let us consider the sigma-algebra  \(\mathcal G\coloneqq \sigma\bigl(f_W(\widetilde X),\widetilde Z\bigr).\) Conditioning on $\mathcal G$ fixes the prediction table and the supersample.  Hence, for every selector vector $u$, the optimized empirical value $\Phi_n(f_W;\widetilde Z_u)$ is fixed and the map \(u\mapsto
\Gamma_{f_W,\widetilde Z}(u)
=
\Phi_n(f_W;\widetilde Z_{\overline u})-
\Phi_n(f_W;\widetilde Z_u)\) is deterministic.  Lemma~\ref{lem:uniform_selector_pairing_app} gives $\Ex_{Q_U}\Gamma_{f_W,\widetilde Z}(U)=0$ under the reference selector law $Q_U=\mathrm{Unif}(\{0,1\}^n)$.

For a realized value of $\mathcal G$, let $P_{U\mid\mathcal G}$ be a regular conditional law of $U$ given $\mathcal G$.  Since $Q_U$ has full support on the finite space of selector vectors, $P_{U\mid\mathcal G}\ll Q_U$.  Also $U$ is independent of $\widetilde Z$, so $Q_U=P_{U\mid\widetilde Z}$ almost surely.  Therefore
\begin{align}
\Ex D(P_{U\mid\mathcal G}\|Q_U)
&=
\Ex D(P_{U\mid f_W(\widetilde X),\widetilde Z}\|P_{U\mid\widetilde Z})
=I(U;f_W(\widetilde X)\mid\widetilde Z).
\label{eq:kl_equals_fcmi_app}
\end{align}
If this quantity is infinite, the theorem is trivial, so assume it is finite.

Next, we upper bound the generalization gap. By the Donsker--Varadhan variational inequality, for every $\lambda>0$ and every realized $\mathcal G$,
\begin{align}
\Ex_{P_{U\mid\mathcal G}}\Gamma_{f_W,\widetilde Z}(U)
&\le
\frac{1}{\lambda}D(P_{U\mid\mathcal G}\|Q_U)
+
\frac{1}{\lambda}
\log\Ex_{Q_U}e^{\lambda\Gamma_{f_W,\widetilde Z}(U)}
\notag\\
&\le
\frac{1}{\lambda}D(P_{U\mid\mathcal G}\|Q_U)
+
\frac{\lambda L_{\Gamma,n}^2}{2},
\label{eq:dv_signed_positive_app}
\end{align}
where the last step uses Assumption~\ref{ass:selector_gap_subgaussian}.  Optimizing over $\lambda>0$ gives \(\Ex_{P_{U\mid\mathcal G}}\Gamma_{f_W,\widetilde Z}(U)
\le
L_{\Gamma,n}\sqrt{2D(P_{U\mid\mathcal G}\|Q_U)}.\) Applying the same argument to $-\Gamma_{f_W,\widetilde Z}$ gives the corresponding lower bound, because Assumption~\ref{ass:selector_gap_subgaussian} holds for all $\lambda\in\mathbb R$.  Thus \(\left|\Ex_{P_{U\mid\mathcal G}}\Gamma_{f_W,\widetilde Z}(U)\right|
\le
L_{\Gamma,n}\sqrt{2D(P_{U\mid\mathcal G}\|Q_U)}.\) Taking expectations and applying Jensen's inequality together with \eqref{eq:kl_equals_fcmi_app},
\begin{align}
\left|\Ex\Gamma_{f_W,\widetilde Z}(U)\right|
&\le
L_{\Gamma,n}\sqrt{2I(f_W(\widetilde X);U\mid\widetilde Z)}.
\end{align}
Since $S=\widetilde Z_U$ and $S'=\widetilde Z_{\overline U}$, this is \eqref{eq:signed_val_train_gap}.

Next, we derive the absolute version bound. For $\lambda>0$, $e^{\lambda|x|}\le e^{\lambda x}+e^{-\lambda x}$.  Using Assumption~\ref{ass:selector_gap_subgaussian} for both signs gives, conditionally on $\mathcal G$,
\begin{align}
\log\Ex_{Q_U}e^{\lambda|\Gamma_{f_W,\widetilde Z}(U)|}
&\le
\log2+\frac{\lambda^2L_{\Gamma,n}^2}{2}.
\end{align}
Donsker--Varadhan applied to $|\Gamma|$ yields
\begin{align}
\Ex_{P_{U\mid\mathcal G}}|\Gamma_{f_W,\widetilde Z}(U)|
&\le
\frac{D(P_{U\mid\mathcal G}\|Q_U)+\log2}{\lambda}
+\frac{\lambda L_{\Gamma,n}^2}{2}.
\end{align}
Optimizing in $\lambda>0$, averaging over $\mathcal G$, and applying Jensen's inequality gives
\begin{align}
\Ex|\Gamma_{f_W,\widetilde Z}(U)|
&\le
L_{\Gamma,n}\sqrt{2\{I(f_W(\widetilde X);U\mid\widetilde Z)+\log2\}},
\end{align}
which is \eqref{eq:absolute_val_train_gap}.
\end{proof}

\paragraph{A sub-Gaussian parameter conditional on the prediction table and supersample.}
Write \(F=f_W(\widetilde X)\), \(G=(F,\widetilde Z)\), and let \(\Gamma_g\) denote the gap as a function of the selector vector for a fixed value \(g\) of \(G\).
Suppose that \(L(g)\ge0\) is measurable, \(\Ex L(G)^2<\infty\), and, for every such \(g\),
\[
\log\Ex_{u\sim Q_U}\exp\{\lambda\Gamma_g(u)\}
\le \lambda^2L(g)^2/2
\qquad(\lambda\in\mathbb R).
\]
Put \(K(g)=D(P_{U\mid G=g}\|Q_U)\).
Repeating the conditional Donsker--Varadhan step of the preceding proof for both signs gives \(K(g)/\lambda+\lambda L(g)^2/2\) as an upper bound on \(\lvert\Ex[\Delta_\Phi\mid G=g]\rvert\). Minimizing over \(\lambda>0\) gives
\(\lvert\Ex[\Delta_\Phi\mid G=g]\rvert\le L(g)\sqrt{2K(g)}\).
Taking expectations and using Cauchy--Schwarz and Eq.~\eqref{eq:kl_equals_fcmi_app} yields
\[
\lvert\Ex\Delta_\Phi\rvert
\le \Ex\!\left[L(G)\sqrt{2K(G)}\right]
\le \sqrt{2\,\Ex[L(G)^2]\,I_{\mathrm{fCMI}}}.
\]
The MGF condition concerns the entire reference distribution of the selector vector with \(g\) fixed. A value chosen from the observed selector vector does not by itself certify this condition.

\paragraph{Bounding fCMI for a given learner.}
For any data-independent prior distribution \(P_0\) on the weights,
\[
I_{\mathrm{fCMI}}
\le I(W;U\mid\widetilde Z)
\le I(W;S)
\le \Ex_S D(P_{W\mid S}\|P_0).
\]
The first inequality follows by conditional data processing since \(F\) is determined by \((W,\widetilde Z)\).
The second follows from \(I(W;S)=I(W;\widetilde Z)+I(W;U\mid\widetilde Z)\): \(S\) is determined by \((\widetilde Z,U)\), and \(W=A(S,R)\) uses independent randomness \(R\).
The last inequality follows from
\(\Ex_S D(P_{W\mid S}\|P_0)=I(W;S)+D(P_W\|P_0)\).
This KL comparison controls the information in the predictor; complement symmetry is still needed to center the difference between separately optimized objectives. The expected posterior--prior KL thus bounds the information term in Theorem~\ref{thm:validation_training_gap}; the result remains an expectation bound.
Even if \(I(W;S)\) is infinite, \(I_{\mathrm{fCMI}}\le H(U)=n\log 2\).

For size-\(k\) compression, \citet{steinke20a} give an \(O(k\log n)\) CMI bound, as well as bounds based on differential privacy and stability.
For SGD, \citet{neu2021sgd} use perturbation analysis involving gradient variance, trajectory-wise local smoothness, and final-output sensitivity. For SGLD, \citet{futami2023timeindependent} bound \(I(W_T;S)\) through KL divergences between output laws and obtain iteration-independent bounds under smoothness and dissipativity.
Bounds based on Gaussian perturbations or PAC-Bayes KL generally require randomized or noise-injected learners. These weight-based bounds complement, rather than replace, prediction-fCMI and VC/pseudo-dimension bounds.

\paragraph{Relation to PAC-Bayes.}
Avoiding an explicit complexity term for \(\Pi\) relies on complement symmetry \(u\mapsto\overline u\), not on change of measure alone.
Under the uniform selector law, the difference between the separately re-optimized training and validation values is exactly centered.
A PAC-Bayes argument with a data-independent prior on the weights does not by itself give this centering, while a posterior on \((W,\pi)\) would generally also pay for the auxiliary variable through its joint posterior--prior KL divergence.
A corresponding PAC-Bayes treatment would require a supersample- or split-conditioned prior preserving this symmetry.
\citet{grunwald2021pac} derive CMI-like PAC-Bayes bounds using priors conditioned on training and ghost samples, providing a starting point for such a treatment.

\subsection{Proofs for Section~\ref{sec:stability_certificate}: stability route}
\label{app:proof_stability_section}

\begin{proof}[Proof of Lemma~\ref{lem:stability_certificate_main}]
Fix a predictor \(f_w\) and a supersample realization \(\widetilde{z}\).  Write \(h(u)\coloneqq \Phi_n(f_w;\widetilde{z}_u),
\qquad
\Gamma(u)\coloneqq h(\overline{u})-h(u).\) Let \(u^{\oplus i}\) denote the selector vector obtained by flipping only the \(i\)-th coordinate.  The selected samples \(\widetilde{z}_u\) and \(\widetilde{z}_{u^{\oplus i}}\) differ in exactly one coordinate.  By Assumption~\ref{ass:one_point_stability},
\begin{align}
|h(u)-h(u^{\oplus i})|
\le
\beta_n.
\label{eq:stability_one_side_h_app}
\end{align}
The complementary selected samples \(\widetilde{z}_{\overline{u}}\) and \(\widetilde{z}_{\overline{u^{\oplus i}}}\) also differ in exactly one coordinate, so
\begin{align}
|h(\overline{u})-h(\overline{u^{\oplus i}})|
\le
\beta_n.
\label{eq:stability_other_side_h_app}
\end{align}
Combining the two displays gives
\begin{align}
|\Gamma(u)-\Gamma(u^{\oplus i})|
&=|h(\overline{u})-h(u)-h(\overline{u^{\oplus i}})+h(u^{\oplus i})|
\notag\\
&\le
|h(\overline{u})-h(\overline{u^{\oplus i}})|
+|h(u)-h(u^{\oplus i})|
\le
2\beta_n.
\label{eq:gap_bd_from_data_stability_app}
\end{align}
Thus \(\Gamma(U)\), as a function of independent selectors under \(Q_U\), has coordinate bounded differences at most \(2\beta_n\).  The bounded differences inequality gives, for all \(\lambda\in\R\),
\begin{align}
\log\Ex_{Q_U}
\exp\{\lambda(\Gamma(U)-\Ex_{Q_U}\Gamma(U))\}
&\le
\frac{\lambda^2}{8}
\sum_{i=1}^n\left(2\beta_n\right)^2
=
\frac{\lambda^2 n\beta_n^2}{2}.
\label{eq:uniform_stability_mgf_app}
\end{align}
Moreover, Lemma~\ref{lem:uniform_selector_pairing_app} gives the reference centering \(\Ex_{Q_U}\Gamma(U)=0\). Combining this centering with \eqref{eq:uniform_stability_mgf_app},  one may take \(L_{\Gamma,n}=\beta_n\sqrt n\) in Assumption~\ref{ass:selector_gap_subgaussian}.  Substituting this value into Theorem~\ref{thm:validation_training_gap} gives \eqref{eq:stability_vt_general_main}.
\end{proof}

\begin{proof}[Proof of Theorem~\ref{thm:wang_general}]
We first derive the upper bound of $|D_t|$.  For each \(t\), the samples \(S^{(t,0)}\) and \(S^{(t,1)}\) differ only in coordinate \(t\). Assumption~\ref{ass:one_point_stability} therefore gives
\begin{align}
|D_t|
=
\left|
\Phi_n(f_W;S^{(t,0)})-
\Phi_n(f_W;S^{(t,1)})
\right|
\le
\beta_n.
\label{eq:wang_dt_bound_app}
\end{align}
The definition of \(S^{(t,b)}\) uses only \(\widetilde{Z}\), \(U_{-t}\), and the predictions of \(f_W\) on the supersample. Hence \(D_t\) is measurable with respect to \((f_W(\widetilde{X}),\widetilde{Z},U_{-t})\).

Define the mixed path
\begin{align}
S^{(0)}\coloneqq S'=\widetilde{Z}_{\overline{U}},
\qquad
S^{(t)}\coloneqq (\widetilde{Z}_{1,U_1},\ldots,\widetilde{Z}_{t,U_t},\widetilde{Z}_{t+1,1-U_{t+1}},\ldots,\widetilde{Z}_{n,1-U_n}),
\quad t=1,\ldots,n.
\label{eq:wang_mixed_path_app}
\end{align}
Then \(S^{(n)}=S=\widetilde{Z}_U\), while \(S^{(t-1)}=S^{(t,1-U_t)}\) and \(S^{(t)}=S^{(t,U_t)}\). Therefore
\begin{align}
\Phi_n(f_W;S')-
\Phi_n(f_W;S)
&=
\sum_{t=1}^n
\left\{
\Phi_n(f_W;S^{(t-1)})-
\Phi_n(f_W;S^{(t)})
\right\}
\notag\\
&=
\sum_{t=1}^n
\left\{
\Phi_n(f_W;S^{(t,1-U_t)})-
\Phi_n(f_W;S^{(t,U_t)})
\right\}.
\label{eq:wang_telescoping_step_app}
\end{align}
If \(U_t=1\), the \(t\)-th summand is \(\Phi_n(f_W;S^{(t,0)})-\Phi_n(f_W;S^{(t,1)})=D_t\). If \(U_t=0\), the same summand is \(\Phi_n(f_W;S^{(t,1)})-\Phi_n(f_W;S^{(t,0)})=-D_t\). Hence the summand is always \(\varepsilon_tD_t\), proving \eqref{eq:wang_telescoping_identity}.

Let \(Q_t=\mathrm{Unif}(\{0,1\})\). Although \(D_t\) may be statistically dependent on \(U_t\) through the learned predictor,  the reference MGF holds \(D_t\) fixed and averages over an independent selector drawn from \(Q_t\). Conditional on \((D_t,\widetilde Z)\), the scalar map \(u_t\mapsto(2u_t-1)D_t\) satisfies
\begin{align}
\Ex_{Q_t}\exp\{\lambda(2U_t-1)D_t\}
&=
\frac12 e^{\lambda D_t}+\frac12 e^{-\lambda D_t}
\le
\exp\left\{\frac{\lambda^2D_t^2}{2}\right\}
\le
\exp\left\{\frac{\lambda^2\beta_n^2}{2}\right\}.
\label{eq:wang_local_mgf_dt_app}
\end{align}
Applying the conditional Donsker--Varadhan variational inequality to \(P_{U_t\mid D_t,\widetilde Z}\) relative to \(Q_t\), averaging the conditional KL divergence to get \(I(D_t;U_t\mid\widetilde Z)\), optimizing over \(\lambda>0\), and repeating the same argument with the negative sign yields
\begin{align}
\left|\Ex[\varepsilon_tD_t]\right|
\le
\beta_n\sqrt{2I(D_t;U_t\mid\widetilde{Z})}.
\label{eq:wang_dt_information_app}
\end{align}
Summing \eqref{eq:wang_dt_information_app} over \(t\) and using \eqref{eq:wang_telescoping_identity} proves the single-index term in \eqref{eq:wang_min_bound}.
\end{proof}

\begin{proof}[Proof of Corollary~\ref{cor:mc_stability}: MC via stability]
Fix a predictor \(f\), and let two samples \(s,s'\in\calz^n\) differ in at most one coordinate, say coordinate \(i\).  For \(\pi=(c,h,\eta)\in\Pi_{\MC}\), define
\begin{align}
g_\pi(f,z)\coloneqq \eta c(x)h(f(x))(y-f(x)).
\end{align}
Since \(c(x)\in[0,1]\), \(h(f(x))\in[-1,1]\), \(\eta\in\{\pm1\}\), and \(y-f(x)\in[-1,1]\), we have \(|g_\pi(f,z)|\le1\).  Hence, for every \(\pi\),
\begin{align}
\left|
\frac1n\sum_{k=1}^n g_\pi(f,z_k)-
\frac1n\sum_{k=1}^n g_\pi(f,z_k')
\right|
=
\frac1n|g_\pi(f,z_i)-g_\pi(f,z_i')|
\le
\frac2n.
\end{align}
Using \eqref{eq:sup_diff_aux},
\begin{align}
|\Phi_{\MC,n}(f;s)-\Phi_{\MC,n}(f;s')|
&\le
\sup_{\pi\in\Pi_{\MC}}
\left|
\frac1n\sum_{k=1}^n g_\pi(f,z_k)-
\frac1n\sum_{k=1}^n g_\pi(f,z_k')
\right|
\le
\frac2n.
\end{align}
Thus Assumption~\ref{ass:one_point_stability} holds with \(\beta_n=2/n\).   The global bound follows from Lemma~\ref{lem:stability_certificate_main}, and the single-index term follows from Theorem~\ref{thm:wang_general}; together these give \eqref{eq:mc_wang_bound_main}.
\end{proof}

\begin{proof}[Proof of Corollary~\ref{cor:cvar_stability}: CVaR via stability]
Let \(\bar\alpha\coloneqq 1-\alpha\).  Fix a predictor \(f\), and write \(L_i=\ell(f(x_i),y_i)\in[0,1]\).  For a threshold \(\tau\in[0,1]\), define
\begin{align}
G_\tau(L_1,\ldots,L_n)
\coloneqq 
\tau+
\frac1{\bar\alpha n}\sum_{i=1}^n(L_i-\tau)_+.
\end{align}
If two samples differ in at most one coordinate, then the corresponding loss vectors differ in at most one coordinate.  Since \((\cdot)_+\) is 1-Lipschitz and the loss is bounded in \([0,1]\),
\begin{align}
|G_\tau(L)-G_\tau(L')|
&\le
\frac1{\bar\alpha n}.
\end{align}
Taking infima over \(\tau\) and using \eqref{eq:inf_diff_aux},
\begin{align}
|\Phi_{\CVaR,\alpha,n}(f;s)-\Phi_{\CVaR,\alpha,n}(f;s')|
&\le
\sup_{\tau\in[0,1]}|G_\tau(L)-G_\tau(L')|
\le
\frac1{\bar\alpha n}.
\end{align}
Thus Assumption~\ref{ass:one_point_stability} holds with \(\beta_n=(\bar\alpha n)^{-1}\).  The global bound follows from Lemma~\ref{lem:stability_certificate_main}, while the local term follows from Theorem~\ref{thm:wang_general}. This proves \eqref{eq:cvar_stability_bounds_main}, \eqref{eq:cvar_lgamma_stability_main}, and the complete statement \eqref{eq:cvar_vt_stability_main} below.
\end{proof}

\begin{proof}[Proof of Corollary~\ref{cor:dro_stability_suboptimal}: suboptimal Cressie--Read DRO via stability]
Fix a predictor \(f\), and write the loss vector as \(r=(r_1,\ldots,r_n)\), where \(r_i=\ell(f(x_i),y_i)\in[0,1]\).  The empirical DRO value is
\begin{align}
\psi_n^{\DRO}(r)
=
\sup_{q\in\mathcal Q_{n,c,\epsilon}}\frac1n q^\top r.
\end{align}
If two samples differ in at most one coordinate, the corresponding loss vectors \(r,r'\) differ in at most one coordinate, say \(i\).  For any \(q\in\mathcal Q_{n,c,\epsilon}\),
\begin{align}
\left|\frac1nq^\top r-\frac1nq^\top r'\right|
&=
\frac1n q_i|r_i-r_i'|
\le
\frac1n\|q\|_\infty.
\end{align}
Using \eqref{eq:sup_diff_aux},
\begin{align}
|\psi_n^{\DRO}(r)-\psi_n^{\DRO}(r')|
&\le
\frac1n\sup_{q\in\mathcal Q_{n,c,\epsilon}}\|q\|_\infty.
\label{eq:dro_stability_first_step_app}
\end{align}
For any feasible \(q\), the Cressie--Read constraint (Eq.~\eqref{eq:app_cr_dro_empirical_short}) gives
\begin{align}
\frac1n\sum_{j=1}^nq_j^c
\le
\kappa_{c,\epsilon}^c.
\end{align}
Therefore \(q_j^c\le n\kappa_{c,\epsilon}^c\) for every \(j\), and hence \(\|q\|_\infty\le\kappa_{c,\epsilon}n^{1/c}\).  Substituting this bound into \eqref{eq:dro_stability_first_step_app} gives
\begin{align}
|\psi_n^{\DRO}(r)-\psi_n^{\DRO}(r')|
&\le
\kappa_{c,\epsilon}n^{1/c-1}
=
\frac{\kappa_{c,\epsilon}n^{1/c}}{n}.
\end{align}
Thus Assumption~\ref{ass:one_point_stability} holds with \(\beta_{\DRO,n}^{\mathrm{stab}}\le\kappa_{c,\epsilon}n^{1/c-1}\), proving \eqref{eq:dro_stability_beta_main}.  Lemma~\ref{lem:stability_certificate_main} gives
\begin{align}
L_{\Gamma,n}^{\DRO,\mathrm{stab}}
&\le
\kappa_{c,\epsilon}n^{1/c-1}\sqrt n
=
\kappa_{c,\epsilon}n^{1/c-1/2},
\end{align}
This proves \eqref{eq:dro_stability_lgamma_main}.  The global bound follows from Lemma~\ref{lem:stability_certificate_main} with \(\beta_n=\kappa_{c,\epsilon}n^{1/c-1}\), while the local term follows from Theorem~\ref{thm:wang_general}. This gives \eqref{eq:dro_stability_bounds_main} and the complete statement \eqref{eq:dro_stability_vt_main} below. The exponent \(1/c-1/2\) comes from the coordinate-wise \(\ell_\infty\) sensitivity of the adversarial weights; Section~\ref{sec:l2_certificate} replaces this by an \(\ell_2\)-geometric certificate.
\end{proof}

\subsection{Proofs for Section~\ref{sec:l2_certificate}: convex \(L_2\) route}
\label{app:proof_l2_section}
Evaluation of the following moment generating function is the key in our analysis. This result is adapted from \citet{wainwright2019high}. For the completeness, we show its proof here;
\begin{lemma}[Convex \texorpdfstring{\(L_2\)}{L2}-Lipschitz MGF concentration]
\label{lem:convex_l2_concentration_app}
Let \(X_1,\ldots,X_n\) be independent random variables taking values in \([0,1]\), and let \(\psi:[0,1]^n\to\mathbb R\) be convex and \(L\)-Lipschitz with respect to the Euclidean norm. Then, for every \(\lambda\in\mathbb R\),
\begin{align}
\log\mathbb E\exp\{\lambda(\psi(X)-\mathbb E\psi(X))\}
\le
\frac{\lambda^2L^2}{2}.
\label{eq:convex_lipschitz_mgf_app}
\end{align}
\end{lemma}
\begin{proof}
Let \(P\) be the law of \(X=(X_1,\ldots,X_n)\).  We first recall the
asymmetric coupling cost used in \citet{wainwright2019high}.  For probability measures \(P,Q\) on
a product space, let \(\Gamma(P,Q)\) denote the set of all couplings \(M\)
of random vectors \(X\sim P\) and \(Y\sim Q\).  Define
\[
C(Q,P)
:=
\inf_{M\in\Gamma(P,Q)}
\left\{
\int
\sum_{i=1}^n
M\{Y_i\ne x_i\mid X=x\}^2
\,dP(x)
\right\}^{1/2}.
\]
Equivalently, reversing the roles of \(P\) and \(Q\),
\[
C(P,Q)
:=
\inf_{M\in\Gamma(P,Q)}
\left\{
\int
\sum_{i=1}^n
M\{X_i\ne y_i\mid Y=y\}^2
\,dQ(y)
\right\}^{1/2}.
\]
For product measures \(P\), Samson's asymmetric inequality (\citet{wainwright2019high}) gives
\[
\max\{C(Q,P),C(P,Q)\}
\le
\sqrt{2D(Q\|P)}
\qquad\text{for all }Q\ll P.
\]

We next derive the variational inequality needed below.  Let
\(Q\ll P\), and let \(M\in\Gamma(P,Q)\).  First suppose that
\(h:[0,1]^n\to\mathbb R\) is convex and \(L\)-Lipschitz with respect to
the Euclidean norm.  At differentiability points, convexity gives
\[
h(y)-h(x)
\le
\langle \nabla h(y),y-x\rangle
\le
\sum_{j=1}^n
|\partial_j h(y)|\,\mathbf 1\{x_j\ne y_j\},
\]
where we used \(x_j,y_j\in[0,1]\).  Hence, with \(b_j(y):=M\{X_j\ne y_j\mid Y=y\},\) we have
\[
\begin{aligned}
\mathbb E_Q h-\mathbb E_P h
&\le
\int
\sum_{j=1}^n |\partial_j h(y)| b_j(y)\,dQ(y)\\
&\le
L
\int
\left(\sum_{j=1}^n b_j(y)^2\right)^{1/2}dQ(y)\\
&\le
L
\left\{
\int
\sum_{j=1}^n b_j(y)^2\,dQ(y)
\right\}^{1/2}.
\end{aligned}
\]
Taking the infimum over \(M\in\Gamma(P,Q)\) yields \(\mathbb E_Q h-\mathbb E_P h
\le
L\,C(P,Q)
\le
L\sqrt{2D(Q\|P)} .\)

If \(h\) is concave and \(L\)-Lipschitz, the same argument is applied with
the gradient evaluated at \(x\).  Indeed, concavity gives
\[
h(y)-h(x)
\le
\langle \nabla h(x),y-x\rangle
\le
\sum_{j=1}^n
|\partial_j h(x)|\,\mathbf 1\{x_j\ne y_j\}.
\]
Thus, with \(a_j(x):=M\{Y_j\ne x_j\mid X=x\},\) the same Cauchy--Schwarz and Jensen steps give \(\mathbb E_Q h-\mathbb E_P h
\le
L\,C(Q,P)
\le
L\sqrt{2D(Q\|P)} .\) The non-differentiable case follows by the standard a.e. differentiability of
Lipschitz convex functions, or by a routine approximation argument.  Therefore,
for every \(h\) that is either convex or concave and \(L\)-Lipschitz,
\begin{align}
\mathbb E_Q h-\mathbb E_P h
\le
L\sqrt{2D(Q\|P)}
\qquad\text{for all }Q\ll P.
\label{eq:convex_concave_variational_input_app}
\end{align}
This is the variational form of the convex/concave concentration argument in
Wainwright~\citep{wainwright2019high}.

We now prove the moment generating function bound.  Let \(h\) be either
convex or concave and \(L\)-Lipschitz, and fix \(\lambda>0\).  Define the tilted
measure
\[
\frac{dQ_\lambda}{dP}(x)
=
\frac{
\exp\{\lambda(h(x)-\mathbb E_P h)\}
}{
\mathbb E_P\exp\{\lambda(h(X)-\mathbb E_P h)\}
}.
\]
Set \(\Lambda_h(\lambda)
:=
\log\mathbb E_P e^{\lambda(h(X)-\mathbb E_P h)}.\) Then \(D(Q_\lambda\|P)
=
\lambda(\mathbb E_{Q_\lambda}h-\mathbb E_P h)
-
\Lambda_h(\lambda),\) and hence, by \eqref{eq:convex_concave_variational_input_app},
\[
\begin{aligned}
\Lambda_h(\lambda)
&=
\lambda(\mathbb E_{Q_\lambda}h-\mathbb E_P h)
-
D(Q_\lambda\|P)\\
&\le
\lambda L\sqrt{2D(Q_\lambda\|P)}
-
D(Q_\lambda\|P)\\
&\le
\sup_{r\ge0}\{\lambda L\sqrt{2r}-r\}
=
\frac{\lambda^2L^2}{2}.
\end{aligned}
\]
Thus, \(\log\mathbb E_P e^{\lambda(h(X)-\mathbb E_P h)}
\le
\frac{\lambda^2L^2}{2},
\qquad \lambda>0,\) for every convex or concave \(L\)-Lipschitz function \(h\).

Finally take \(h=\psi\).  Since \(\psi\) is convex and \(L\)-Lipschitz, the
preceding display gives \(\log\mathbb E_P e^{\lambda(\psi(X)-\mathbb E_P\psi)}
\le
\frac{\lambda^2L^2}{2},
\qquad \lambda>0.\) For \(\lambda<0\), put \(\mu=-\lambda>0\) and \(h=-\psi\).  Then \(h\) is
concave and \(L\)-Lipschitz, and \(\mu(h-\mathbb E_P h)
=
\lambda(\psi-\mathbb E_P\psi).\) Applying the positive-\(\mu\) bound to \(h=-\psi\) gives \(\log\mathbb E_P e^{\lambda(\psi(X)-\mathbb E_P\psi)}
\le
\frac{\lambda^2L^2}{2}.\) The case \(\lambda=0\) is trivial, so the bound holds for all
\(\lambda\in\mathbb R\).
\end{proof}

\begin{proof}[Proof of Lemma~\ref{lem:convex_l2_certificate_main}]
\emph{ } Fix a predictor \(f_w\) and a supersample realization \(\widetilde{z}\).  Write \(\psi_n\) for the convex aggregation map \(\tilde\psi\) in Assumption~\ref{ass:convex_l2_score_vector_main}.  For each selector vector \(u\in\{0,1\}^n\), define the selected score vector
\begin{align}
R_u\coloneqq R_{f_w,n}(\widetilde{z}_u)\in[0,1]^n.
\label{eq:selected_score_vector_app}
\end{align}
By Assumption~\ref{ass:convex_l2_score_vector_main},
\begin{align}
\Phi_n(f_w;\widetilde{z}_u)=\psi_n(R_u),
\qquad
\Phi_n(f_w;\widetilde{z}_{\overline{u}})=\psi_n(R_{\overline{u}}).
\label{eq:l2_representation_selected_app}
\end{align}
The selector gap is therefore
\begin{align}
\Gamma_{f_w,\widetilde{z}}(u)
=
\psi_n(R_{\overline{u}})-\psi_n(R_u).
\label{eq:l2_gap_as_score_difference_app}
\end{align}

The selector vector is not being relaxed.  The only point is that, after fixing \((f_w,\widetilde{z})\), the random vector \(R_U\) is a vector of independent bounded coordinates under the reference selector law.  Indeed, since \(R_{f_w,n}\) is coordinatewise, its \(i\)-th coordinate can be written as a measurable function of the \(i\)-th data point alone.  Thus, for fixed \((f_w,\widetilde{z})\), the coordinate \((R_U)_i\) depends only on the selector \(U_i\).  Because \(U_1,\ldots,U_n\) are independent under \(Q_U\), the coordinates of \(R_U\) are independent and take values in \([0,1]\).  The same reasoning applies to \(R_{\overline{U}}\), since its \(i\)-th coordinate depends only on \(U_i\) as well.  Consequently the convex \(L_2\)-Lipschitz concentration inequality in Lemma~\ref{lem:convex_l2_concentration_app} applies to both \(\psi_n(R_U)\) and \(\psi_n(R_{\overline{U}})\).

The gap is antisymmetric by definition: \(\Gamma_{f_w,\widetilde{z}}(\overline{u})=-\Gamma_{f_w,\widetilde{z}}(u)\). Lemma~\ref{lem:uniform_selector_pairing_app} therefore gives \(\Ex_{Q_U}\Gamma_{f_w,\widetilde{z}}(U)=0\). Let \(\mu\coloneqq \Ex_{Q_U}\psi_n(R_U)\).
The complement map is a bijection on \(\{0,1\}^n\), so
\begin{align}
\Ex_{Q_U}\psi_n(R_{\overline{U}})
&=
2^{-n}\sum_{u\in\{0,1\}^n}\psi_n(R_{\overline{u}})
=
2^{-n}\sum_{v\in\{0,1\}^n}\psi_n(R_v)
=
\mu,
\label{eq:l2_bar_mean_equals_mu_app}
\end{align}
where the middle equality uses the change of variables \(v=\overline{u}\).  Combining \eqref{eq:l2_gap_as_score_difference_app} and \eqref{eq:l2_bar_mean_equals_mu_app},
\begin{align}
\Gamma_{f_w,\widetilde{z}}(U)
=
\{\psi_n(R_{\overline{U}})-\mu\}-\{\psi_n(R_U)-\mu\}.
\label{eq:l2_gap_centered_decomp_app}
\end{align}

 By using Cauchy--Schwarz inequality,
\begin{align}
\Ex_{Q_U}e^{\lambda\Gamma_{f_w,\widetilde{z}}(U)}
&\le
\left(\Ex_{Q_U}e^{2\lambda(\psi_n(R_{\overline{U}})-\mu)}\right)^{1/2}
\left(\Ex_{Q_U}e^{-2\lambda(\psi_n(R_U)-\mu)}\right)^{1/2}.
\label{eq:l2_gap_cauchy_app}
\end{align}
Since \(R_U\) has independent \([0,1]\)-valued coordinates and \(\psi_n\) is convex and \(L_{\psi,n}\)-Lipschitz, Lemma~\ref{lem:convex_l2_concentration_app} gives
\begin{align}
\Ex_{Q_U}e^{-2\lambda(\psi_n(R_U)-\mu)}
\le
\exp\{2\lambda^2L_{\psi,n}^2\}.
\label{eq:l2_mgf_RU_app}
\end{align}
The same inequality applied to \(R_{\overline{U}}\) gives
\begin{align}
\Ex_{Q_U}e^{2\lambda(\psi_n(R_{\overline{U}})-\mu)}
\le
\exp\{2\lambda^2L_{\psi,n}^2\}.
\label{eq:l2_mgf_RbarU_app}
\end{align}
Substituting \eqref{eq:l2_mgf_RU_app} and \eqref{eq:l2_mgf_RbarU_app} into \eqref{eq:l2_gap_cauchy_app}, we obtain
\begin{align}
\Ex_{Q_U}e^{\lambda\Gamma_{f_w,\widetilde{z}}(U)}
&\le
\exp\{2\lambda^2L_{\psi,n}^2\}
=
\exp\left\{\frac{\lambda^2(2L_{\psi,n})^2}{2}\right\}.
\label{eq:l2_gap_mgf_final_app}
\end{align}
Thus Assumption~\ref{ass:selector_gap_subgaussian} holds with \(L_{\Gamma,n}\le2L_{\psi,n}\).
\end{proof}

\begin{lemma}[ A convex extension of the empirical MC objective]
\label{lem:mc_selector_interpolation_certificate}
For the MC objective~\eqref{eq:mc_objective_prelim}, the selector gap satisfies Assumption~\ref{ass:selector_gap_subgaussian} with \(L_{\Gamma,n}\le 4/\sqrt n\)  by the following convex extension.
\end{lemma}

\begin{proof}
Fix a predictor \(f\) and a supersample realization \(\widetilde{z}\).  For each witness \(\pi=(c,h,\eta)\in\Pi_{\MC}\), define
\begin{align}
a_{\pi,i,j}
\coloneqq 
\eta c(\widetilde{x}_{i,j})h(f(\widetilde{x}_{i,j}))
\bigl(\widetilde{y}_{i,j}-f(\widetilde{x}_{i,j})\bigr),
\qquad i\in[n],\ j\in\{0,1\}.
\end{align}
Since \(c\in[0,1]\), \(h\) takes values in \([-1,1]\), \(\eta\in\{\pm1\}\), and \(\widetilde{y}-f(\widetilde{x})\in[-1,1]\), we have \(|a_{\pi,i,j}|\le1\).   For \(\theta\in[0,1]^n\), we interpolate the two witness scores in each row and then take the supremum over witnesses:
\begin{align}
H(\theta)
\coloneqq 
\sup_{\pi\in\Pi_{\MC}}
\frac1n\sum_{i=1}^n
\bigl\{(1-\theta_i)a_{\pi,i,0}+\theta_i a_{\pi,i,1}\bigr\}.
\label{eq:mc_selector_interpolation_function_app}
\end{align}
This interpolation keeps the witness optimization intact.  For each fixed \(\pi\), the displayed average is affine in \(\theta\), so \(H\), as a supremum of affine functions, is convex.  Moreover, for any \(\theta,\theta'\in[0,1]^n\), the supremum-difference inequality gives
\begin{align}
|H(\theta)-H(\theta')|
&\le
\sup_{\pi\in\Pi_{\MC}}
\frac1n\left|
\sum_{i=1}^n(\theta_i-\theta_i')
(a_{\pi,i,1}-a_{\pi,i,0})
\right|
\notag\\
&\le
\frac1n\sup_{\pi\in\Pi_{\MC}}
\left(\sum_{i=1}^n(a_{\pi,i,1}-a_{\pi,i,0})^2\right)^{1/2}
\|\theta-\theta'\|_2
\le
\frac{2}{\sqrt n}\|\theta-\theta'\|_2.
\label{eq:mc_selector_interpolation_lipschitz_app}
\end{align}
For binary selector vectors, \(H(u)=\Phi_{\MC,n}(f;\widetilde{z}_u)\).  

Under the reference law \(Q_U\), the vector \(U\) has independent \([0,1]\)-valued coordinates, and \(H\) is convex and \(2/\sqrt n\)-Lipschitz.  Therefore Lemma~\ref{lem:convex_l2_concentration_app} applies to both \(H(U)\) and \(H(\overline U)\).  Since the selector gap is \(\Gamma_{f,\widetilde z}(U)=H(\overline U)-H(U)\) and \(H(U)\) and \(H(\overline U)\) have the same reference mean, the Cauchy--Schwarz MGF argument used in Lemma~\ref{lem:convex_l2_certificate_main} gives \(\log\Ex_{Q_U}e^{\lambda\Gamma_{f,\widetilde z}(U)}
\le
\frac{\lambda^2(4/\sqrt n)^2}{2}.\) Thus \(L_{\Gamma,n}\le4/\sqrt n\). 
\end{proof}
 The stability certificate gives the sharper MC value \(2/\sqrt n\), but the interpolation argument explains the convex-geometry entry in Table~\ref{tab:gap_summary}.

\begin{proof}[Proof of Corollary~\ref{cor:cvar_l2}: CVaR via convex \(L_2\) geometry]
For a fixed predictor \(f\), let \(R_{f,n}(s)\in[0,1]^n\) be the loss vector with coordinates \((R_{f,n}(s))_i=\ell(f(x_i),y_i)\). This map is coordinatewise, and the empirical CVaR value is \(\psi_n^{\CVaR}(R_{f,n}(s))\) for
\begin{align}
\psi_n^{\CVaR}(r)=\sup_{q\in Q_{\alpha,n}}\frac1nq^\top r,
\qquad r\in[0,1]^n,
\end{align}
where
\[
Q_{\alpha,n}=\left\{q\in\R_+^n:\frac1n\sum_{i=1}^nq_i=1,
\;0\le q_i\le\bar\alpha^{-1}\right\}.
\]
Because \(\psi_n^{\CVaR}\) is a supremum of affine functions of \(r\), it is convex.  The set \(Q_{\alpha,n}\) is compact and nonempty, so for every \(r\) a maximizer exists.  Choose \(q^\star(r)\in\arg\max_{q\in Q_{\alpha,n}}n^{-1}q^\top r\).  For any \(r,r'\in[0,1]^n\),
\begin{align}
\psi_n^{\CVaR}(r)-\psi_n^{\CVaR}(r')
&=
\frac1n q^\star(r)^\top r-
\sup_{q\in Q_{\alpha,n}}\frac1n q^\top r'
\notag\\
&\le
\frac1n q^\star(r)^\top r-
\frac1n q^\star(r)^\top r'
=
\frac1n q^\star(r)^\top(r-r')
\notag\\
&\le
\frac1n\|q^\star(r)\|_2\|r-r'\|_2.
\end{align}
For any \(q\in Q_{\alpha,n}\), since \(q_i\le\bar\alpha^{-1}\),
\begin{align}
q_i^2\le \bar\alpha^{-1}q_i,
\qquad
\sum_{i=1}^n q_i^2
\le
\bar\alpha^{-1}\sum_{i=1}^nq_i
=
\frac{n}{\bar\alpha}.
\end{align}
Thus \(\|q\|_2\le\sqrt{n/\bar\alpha}\), and
\begin{align}
\psi_n^{\CVaR}(r)-\psi_n^{\CVaR}(r')
&\le
\frac1{\sqrt{\bar\alpha n}}\|r-r'\|_2.
\end{align}
Interchanging \(r\) and \(r'\) gives the reverse difference, so \(\psi_n^{\CVaR}\) is \((\bar\alpha n)^{-1/2}\)-Lipschitz.  Lemma~\ref{lem:convex_l2_certificate_main} gives
\begin{align}
L_{\Gamma,n}^{\CVaR,L_2}
\le
2(\bar\alpha n)^{-1/2}.
\label{eq:cvar_l2_lgamma_main}
\end{align}
This proves \eqref{eq:cvar_l2_lgamma_main}.  Combining this value with \eqref{eq:cvar_lgamma_stability_main} gives \eqref{eq:cvar_lgamma_main}.
\end{proof}

\begin{proof}[Proof of Corollary~\ref{cor:dro_l2}]
For a fixed predictor \(f\), let \(R_{f,n}(s)\in[0,1]^n\) be the loss vector with coordinates \((R_{f,n}(s))_i=\ell(f(x_i),y_i)\). This map is coordinatewise, and the empirical Cressie--Read DRO value is \(\psi_n^{\DRO}(R_{f,n}(s))\), where
\begin{align}
\psi_n^{\DRO}(r)
=
\sup_{q\in\mathcal Q_{n,c,\epsilon}}\frac1nq^\top r,
\qquad r\in[0,1]^n.
\end{align}
It is a supremum of affine functions of \(r\), and hence is convex.  The feasible set \(\mathcal Q_{n,c,\epsilon}\) is closed, bounded, and nonempty, so a maximizer exists for every \(r\).  Choose \(q^\star(r)\in\arg\max_{q\in\mathcal Q_{n,c,\epsilon}}n^{-1}q^\top r\).  For any \(r,r'\in[0,1]^n\),
\begin{align}
\psi_n^{\DRO}(r)-\psi_n^{\DRO}(r')
&=
\frac1nq^\star(r)^\top r-
\sup_{q\in\mathcal Q_{n,c,\epsilon}}\frac1nq^\top r'
\notag\\
&\le
\frac1nq^\star(r)^\top r-
\frac1nq^\star(r)^\top r'
=
\frac1nq^\star(r)^\top(r-r')
\notag\\
&\le
\frac1n\|q^\star(r)\|_2\|r-r'\|_2.
\end{align}
Taking the supremum over feasible \(q\) and interchanging \(r\) and \(r'\),
\begin{align}
|\psi_n^{\DRO}(r)-\psi_n^{\DRO}(r')|
&\le
\frac1n\sup_{q\in\mathcal Q_{n,c,\epsilon}}\|q\|_2\,\|r-r'\|_2.
\label{eq:dro_l2_lip_detailed_app}
\end{align}
It remains to evaluate \(\sup_q\|q\|_2\).  The Cressie--Read constraint gives
\begin{align}
\frac1n\sum_{i=1}^nq_i^c
&\le
\kappa_{c,\epsilon}^c,
\qquad\text{so}\qquad
\|q\|_c
=
\left(\sum_{i=1}^nq_i^c\right)^{1/c}
\le
\kappa_{c,\epsilon}n^{1/c}.
\label{eq:dro_lc_norm_app}
\end{align}
If \(c\ge2\), the norm comparison \(\|q\|_2\le n^{1/2-1/c}\|q\|_c\) yields
\begin{align}
\|q\|_2
&\le
n^{1/2-1/c}\kappa_{c,\epsilon}n^{1/c}
=
\kappa_{c,\epsilon}n^{1/2}.
\end{align}
If \(1<c<2\), then \(\|q\|_2\le\|q\|_c\), and hence
\begin{align}
\|q\|_2
&\le
\kappa_{c,\epsilon}n^{1/c}.
\end{align}
Therefore
\begin{align}
\frac1n\sup_{q\in\mathcal Q_{n,c,\epsilon}}\|q\|_2
&\le
\kappa_{c,\epsilon}
\begin{cases}
 n^{-1/2}, & c\ge2,\\
 n^{1/c-1}, & 1<c<2,
\end{cases}
\notag\\
&=
\kappa_{c,\epsilon}
\begin{cases}
 n^{-1/2}, & c\ge2,\\
 n^{-1/c^\ast}, & 1<c<2,
\end{cases}
\notag\\
&=
\kappa_{c,\epsilon}n^{-1/(c^\ast\vee2)}.
\label{eq:dro_l2_lip_main}
\end{align}
Combining this with \eqref{eq:dro_l2_lip_detailed_app} proves \eqref{eq:dro_l2_lip_main}.  Lemma~\ref{lem:convex_l2_certificate_main} gives
\begin{align}
L_{\Gamma,n}^{\DRO,L_2}
&\le
2L_{\psi,n}^{\DRO}
\le
2\kappa_{c,\epsilon}n^{-1/(c^\ast\vee2)}.
\label{eq:dro_lgamma_main}
\end{align}
This proves \eqref{eq:dro_lgamma_main}.  Finally, Theorem~\ref{thm:validation_training_gap} gives
\begin{align}
|\Ex\Delta_{\DRO}|
&\le
2\kappa_{c,\epsilon}n^{-1/(c^\ast\vee2)}
\sqrt{2I(f_W(\widetilde{X});U\mid\widetilde{Z})},
\end{align}
which is \eqref{eq:dro_vt_bound_main}.
\end{proof}

\subsection{Proofs for Section~\ref{sec:population_validation_terms}: fixed-predictor terms}
\label{app:proof_population_validation_section}

\begin{corollary}[Fixed-predictor CVaR term]
\label{cor:cvar_population_validation}
Assume \(0\le \ell(f(X),Y)\le1\) for all predictors and data points. For CVaR, we have
\begin{align}
\sup_f
\Ex_{S'\sim\cald^n}
\left|
\Phi_{\CVaR,\alpha}(f;\cald)-
\Phi_{\CVaR,\alpha,n}(f;S')
\right|
\le
\frac{1}{\bar\alpha}\sqrt{\frac{\pi}{2n}}.
\label{eq:cvar_fixed_population_validation_main}
\end{align}
Therefore, with \(L_{\Gamma,n}^{\CVaR}\) chosen as in \eqref{eq:cvar_lgamma_main},
\begin{align}
\Ex\left|
\Phi_{\CVaR,\alpha}(f_W;\cald)-
\Phi_{\CVaR,\alpha,n}(f_W;S)
\right|
\le
\frac{1}{\bar\alpha}\sqrt{\frac{\pi}{2n}}
+
L_{\Gamma,n}^{\CVaR}
\sqrt{2\{I(f_W(\widetilde{X});U\mid\widetilde{Z})+\log 2\}}.
\label{eq:cvar_population_bound_main}
\end{align}
\end{corollary}

\begin{corollary}[Duchi--Namkoong fixed-predictor Cressie--Read DRO term]
\label{cor:dro_population_validation}
Let \(c>1\), \(c^\ast=c/(c-1)\), \(r_c\coloneqq 1/(c^\ast\vee2)\), and let \(\epsilon>0\). Define
\[
\kappa_{c,\epsilon}\coloneqq \{1+c(c-1)\epsilon\}^{1/c},
\qquad
\mathsf K_{c,\epsilon}
\coloneqq
\kappa_{c,\epsilon}^{2}
\left(\frac{\kappa_{c,\epsilon}}{\kappa_{c,\epsilon}-1}\vee 2\right).
\]
Assume \(0\le\ell(f(X),Y)\le1\) for all predictors and data points, \(n\ge c\vee3\), and the fixed-parameter conditions of \citet{Duchi2021} hold uniformly for the scalar loss laws induced by the predictors under consideration. Then
\begin{align}
\Ex_{f_W,S'}
\left|
\Phi_{\DRO,c,\epsilon}(f_W;\cald)-
\Phi_{\DRO,c,\epsilon,n}(f_W;S')
\right|
&\le
\widetilde{\mathcal{O}}\!\left(
\text{}\mathsf K_{c,\epsilon}n^{-r_c}
\right).
\label{eq:dro_fixed_population_validation_main}
\end{align}
Consequently, combining this fixed-predictor term with Corollary~\ref{cor:dro_l2} gives
\begin{align}
\Ex|
\Phi_{\DRO,c,\epsilon}(f_W;\cald)-
&\Phi_{\DRO,c,\epsilon,n}(f_W;S)
|
\le
\widetilde{\mathcal{O}}\!\left(
\text{}\mathsf K_{c,\epsilon}n^{-r_c}
\right)
\notag\\
&\quad+
2\text{}\kappa_{c,\epsilon}n^{-r_c}
\sqrt{2\{I(f_W(\widetilde{X});U\mid\widetilde{Z})+\log 2\}}.
\label{eq:dro_population_bound_main}
\end{align}
Here \(\widetilde{\mathcal O}\) hides only logarithmic factors in \(n\) and universal numerical constants, not the displayed \(\text{}\mathsf K_{c,\epsilon}\) dependence.  At \(\epsilon=0\), the ambiguity set collapses to the ordinary mean, and the first term can be replaced by the standard bounded-mean comparison.
\end{corollary}

\begin{proof}[Proof of Corollary~\ref{cor:cvar_population_validation}: fixed-predictor CVaR term]
Fix a predictor \(f\), and let \(L=\ell(f(X),Y)\in[0,1]\) and \(L_i=\ell(f(X_i'),Y_i')\).  For \(\tau\in[0,1]\), define
\begin{align}
G(\tau)&\coloneqq \tau+\frac1{\bar\alpha}\Ex(L-\tau)_+,\\
G_n(\tau)&\coloneqq \tau+\frac1{\bar\alpha n}\sum_{i=1}^n(L_i-\tau)_+.
\end{align}
Using \eqref{eq:inf_diff_aux},
\begin{align}
\left|
\Phi_{\CVaR,\alpha}(f;\cald)-\Phi_{\CVaR,\alpha,n}(f;S')
\right|
&\le
\sup_{\tau\in[0,1]}|G(\tau)-G_n(\tau)|
\notag\\
&=
\frac1{\bar\alpha}
\sup_{\tau\in[0,1]}
\left|
\Ex(L-\tau)_+-\frac1n\sum_{i=1}^n(L_i-\tau)_+
\right|.
\end{align}
The identity \((x-\tau)_+=\int_\tau^1\one\{x>s\}\,ds\) for \(x\in[0,1]\) gives
\begin{align}
&\sup_{\tau\in[0,1]}
\left|
\Ex(L-\tau)_+-\frac1n\sum_{i=1}^n(L_i-\tau)_+
\right|
\notag\\
&\qquad\le
\int_0^1
\sup_{s\in[0,1]}
\left|
\Pr(L>s)-\frac1n\sum_{i=1}^n\one\{L_i>s\}
\right|ds
\notag\\
&\qquad=
\sup_{s\in[0,1]}
\left|
\Pr(L>s)-\frac1n\sum_{i=1}^n\one\{L_i>s\}
\right|.
\end{align}
Let
\(
M_n\coloneqq
\sup_{s\in[0,1]}
\left|
\Pr(L>s)-n^{-1}\sum_{i=1}^n\one\{L_i>s\}
\right|.
\)
The Dvoretzky--Kiefer--Wolfowitz--Massart inequality gives
\(\Pr(M_n>t)\le 2\exp(-2nt^2)\) for all \(t>0\)
\citep{massart1990tight}.  Therefore
\begin{align}
\Ex M_n
&=
\int_0^\infty \Pr(M_n>t)\,dt
\le
\int_0^\infty 2e^{-2nt^2}\,dt
=
\sqrt{\frac{\pi}{2n}}.
\end{align}
This bound is distribution-free over all scalar laws supported on \([0,1]\).  Hence it holds uniformly in the fixed predictor \(f\), and it also applies after conditioning on the learned predictor \(f_W\).  Multiplying by \(\bar\alpha^{-1}\) proves \eqref{eq:cvar_fixed_population_validation_main}.

The total bound \eqref{eq:cvar_population_bound_main} follows from \eqref{eq:population_validation_split}, the triangle inequality, \eqref{eq:cvar_fixed_population_validation_main}, and the absolute validation--training bound \eqref{eq:absolute_val_train_gap} with \(L_{\Gamma,n}^{\CVaR}\) as in \eqref{eq:cvar_lgamma_main}.
\end{proof}

\begin{proof}[Proof of Corollary~\ref{cor:dro_population_validation}: fixed-predictor Cressie--Read DRO term]
Let \(f_W\) be the learned predictor and condition on a realization \(f_W=f\).  The scalar loss \(L_f(Z)=\ell(f(X),Y)\) is then fixed and takes values in \([0,1]\).  Write \(X_f
\coloneqq
\left|
\Phi_{\DRO,c,\epsilon}(f;\cald)-
\Phi_{\DRO,c,\epsilon,n}(f;S')
\right|.\) For \(f_c(t)=(t^c-ct+c-1)/(c(c-1))\), \citet{Duchi2021} gives, under the stated fixed-parameter conditions, for every \(t>0\),
\begin{align}
X_f
&\le
10\text{}\mathsf K_{c,\epsilon}
 n^{-r_c}
\left(\frac1c+\sqrt t+2\log n\right)
\label{eq:dn_hp_bound_app}
\end{align}
with probability at least \(1-2e^{-t}\).  The constants are uniform over the scalar loss laws considered, so the same display holds after conditioning on any realization of \(f_W\).

Integrating the conditional tail bound yields the expectation bound.  Indeed, with \(B_{c,\epsilon,\text{}n}=10\text{}\mathsf K_{c,\epsilon}n^{-r_c}\) and \(a_{c,n}=c^{-1}+2\log n\), \eqref{eq:dn_hp_bound_app} implies
\[
\Pr\{X_f>B_{c,\epsilon,\text{}n}(a_{c,n}+s)\mid f_W=f\}
\le
2e^{-s^2},
\qquad s\ge0.
\]
Therefore
\begin{align}
\Ex[X_f\mid f_W=f]
&\le
B_{c,\epsilon,\text{}n}a_{c,n}
+
B_{c,\epsilon,\text{}n}
\int_0^\infty \min\{1,2e^{-s^2}\}\,ds
\notag\\
&\le
C \text{}\mathsf K_{c,\epsilon}n^{-r_c}(1+\log n),
\end{align}
for a universal numerical constant \(C\).  Averaging over \(f_W\) proves \eqref{eq:dro_fixed_population_validation_main}.  The boundary case \(\epsilon=0\) is handled separately by the ordinary bounded-mean comparison, since the ambiguity set is the singleton containing the nominal law.

This argument uses only the fixed-predictor theorem after conditioning on \(f_W\).  It does not use the separate uniform-in-class covering-number bound of \citet{Duchi2021}; the remaining randomness is the independent validation sample \(S'\), while the algorithm-dependent part is controlled by the validation--training fCMI term.

Using \eqref{eq:population_validation_split} and the triangle inequality,
\begin{align}
\Ex\left|
\Phi_{\DRO,c,\epsilon}(f_W;\cald)-
\Phi_{\DRO,c,\epsilon,n}(f_W;S)
\right|
&\le
\Ex\left|
\Phi_{\DRO,c,\epsilon}(f_W;\cald)-
\Phi_{\DRO,c,\epsilon,n}(f_W;S')
\right|
\notag\\
&\quad+
\Ex\left|
\Phi_{\DRO,c,\epsilon,n}(f_W;S')-
\Phi_{\DRO,c,\epsilon,n}(f_W;S)
\right|.
\end{align}
The first term is bounded by \eqref{eq:dro_fixed_population_validation_main}.   Applying Theorem~\ref{thm:validation_training_gap}, Eq.~\eqref{eq:absolute_val_train_gap}, with \(L_{\Gamma,n}^{\DRO,L_2}=2\text{}\kappa_{c,\epsilon}n^{-r_c}\) bounds the second term and gives \eqref{eq:dro_population_bound_main}.
\end{proof}

\subsection{Proof of uniform convergence of Multicalibration}\label{app:mc_proof} 

\begin{lemma}[Fixed-predictor MC empirical-process bound]
\label{lem:mc_fixed_emp_process}
For every fixed predictor \(f:\mathcal X\to[0,1]\), define \(\UC_n^{\MC}(f)
:=
\Ex_{S'\sim\cald^n}
\left|
\Phi_{\MC}(f;\cald)-\Phi_{\MC,n}(f;S')
\right|.\) Let \(\cC_{\pm}:=\{\eta c:c\in\cC,\ \eta\in\{\pm1\}\}\) and define the fixed-predictor MC product class
\[
\mathcal G_f
:=
\left\{
(x,y)\mapsto a(x)h(f(x))(y-f(x)):
a\in\cC_{\pm},\ h\in\cH
\right\}.
\]
Here \(\mathfrak R_{\cald,n}(\mathcal G_f)\) is the usual Rademacher
average on \(Z\)-valued samples, while
\(\mathfrak R_{\cald,n}(\cC_{\pm})\) is evaluated under the \(X\)-marginal
of \(\cald\). Then
\begin{align}
\UC_n^{\MC}(f)
\le
2\mathfrak R_{\cald,n}(\mathcal G_f).
\label{eq:mc_fixed_emp_process_general_app}
\end{align}
Moreover, the product structure gives the sufficient separable bound
\begin{align}
\mathfrak R_{\cald,n}(\mathcal G_f)
\lesssim
\mathfrak R_{\cald,n}(\cC_{\pm})
+
\inf_{\delta\in[0,1]}
\left\{
\delta+
\int_\delta^1
\sqrt{
\frac{\log N(\varepsilon,\cH,\|\cdot\|_\infty)}{n}
}\,d\varepsilon
\right\}.
\label{eq:mc_fixed_emp_process_separable_app}
\end{align}
Consequently,
\begin{align}
\UC_n^{\MC}(f)
\lesssim
\mathfrak R_{\cald,n}(\cC_{\pm})
+
\inf_{\delta\in[0,1]}
\left\{
\delta+
\int_\delta^1
\sqrt{
\frac{\log N(\varepsilon,\cH,\|\cdot\|_\infty)}{n}
}\,d\varepsilon
\right\}.
\label{eq:mc_fixed_emp_process_bound_app}
\end{align}
\end{lemma}

\begin{proof}
We first reduce the fixed-predictor population--validation discrepancy to an
empirical-process term. For \(a=\eta c\in\cC_{\pm}\) and \(h\in\cH\), set \(g_{a,h}^f(x,y)
=
a(x)h(f(x))(y-f(x)).\) Since \(|a|\le1\), \(|h|\le1\), and \(y-f(x)\in[-1,1]\), every
\(g_{a,h}^f\) is bounded by one. The supremum-difference inequality gives
\[
\left|
\Phi_{\MC}(f;\cald)-\Phi_{\MC,n}(f;S')
\right|
\le
\sup_{g\in\mathcal G_f}|Pg-P_ng|,
\]
where \(P=\cald\) and \(P_n\) is the empirical law of \(S'\). Standard
symmetrization, as in \citet{wainwright2019high}, yields \(\UC_n^{\MC}(f)
\le
2\mathfrak R_{\cald,n}(\mathcal G_f),\) which proves \eqref{eq:mc_fixed_emp_process_general_app}.

It remains to prove the separable product-class bound. Fix a sample \(S=((X_i,Y_i))_{i=1}^n,
\qquad
r_i:=Y_i-f(X_i),
\qquad |r_i|\le1.\) Conditioning on this sample, write
\[
\widehat{\mathfrak R}_{S}(\mathcal G_f)
=
\Ex_\sigma
\sup_{a\in\cC_{\pm},\,h\in\cH}
\frac1n\sum_{i=1}^n
\sigma_i a(X_i)h(f(X_i))r_i,
\]
and let
\[
\widehat{\mathfrak R}_{X}(\cC_{\pm})
=
\Ex_\sigma
\sup_{a\in\cC_{\pm}}
\frac1n\sum_{i=1}^n
\sigma_i a(X_i)
\]
be the empirical Rademacher average of the subgroup class on the observed
\(X\)-sample.

We approximate the score witness \(h\) by a dyadic chain in
\(\|\cdot\|_\infty\). If one of the covering numbers below is infinite, the
right-hand side is infinite and there is nothing to prove. Thus assume finite
covers at the relevant scales. Fix \(\delta\in(0,1]\), set \(\varepsilon_k:=2^{-k},
\qquad
\varepsilon_0:=1,\) and choose \(m\) such that \(\varepsilon_m\le\delta<2\varepsilon_m.\) For each \(k=1,\ldots,m\), let \(\widetilde\cH_k\) be a minimal
\(\varepsilon_k\)-net of \(\cH\) in \(\|\cdot\|_\infty\), and write \(N_k
:=
|\widetilde\cH_k|
=
N(\varepsilon_k,\cH,\|\cdot\|_\infty).\) For every \(h\in\cH\), choose one net point
\(h_k(h)\in\widetilde\cH_k\) satisfying \(\|h-h_k(h)\|_\infty\le\varepsilon_k,\) and set \(h_0(h)\equiv0\). Then
\[
h
=
\sum_{k=1}^m\{h_k(h)-h_{k-1}(h)\}
+
\{h-h_m(h)\}.
\]

For each scale \(k\), define the finite set of actual increment functions \(\mathcal D_k
:=
\left\{
 h_k(h)-h_{k-1}(h):h\in\cH
\right\}.\) Each \(d\in\mathcal D_k\) is a function on the score space \([0,1]\), because
both \(h_k(h)\) and \(h_{k-1}(h)\) are score-witness functions. Hence
\(d(f(X_i))\) is well-defined. The word ``actual'' means that we include only
pairs of net points that arise from the same \(h\in\cH\) under the chosen
nearest-net assignments.

The cardinality of \(\mathcal D_k\) is controlled by the covering numbers. For
\(k=1\), an increment is determined only by the fine net point \(h_1(h)\), since
\(h_0(h)\equiv0\), so \(|\mathcal D_1|\le N_1.\) For \(k\ge2\), an increment is determined by a pair of net points, \(h_k(h)\in\widetilde\cH_k,
\qquad
h_{k-1}(h)\in\widetilde\cH_{k-1},\) and hence \(|\mathcal D_k|\le N_kN_{k-1}.\) Since \(\varepsilon_k\le\varepsilon_{k-1}\), covering numbers are monotone and
\(N_{k-1}\le N_k\). Therefore, with \(M_k:=|\mathcal D_k|\), \(\log M_k
\lesssim
\log N_k
=
\log N(\varepsilon_k,\cH,\|\cdot\|_\infty)\) for every \(k\), up to universal numerical constants.

We now decompose the empirical Rademacher average into the residual term and
the dyadic increment terms. For each \(h\in\cH\), write \(d_k^h:=h_k(h)-h_{k-1}(h),
\qquad
e_m^h:=h-h_m(h).\) Then, for each sample point,
\[
h(f(X_i))r_i
=
\sum_{k=1}^m d_k^h(f(X_i))r_i
+
e_m^h(f(X_i))r_i.
\]
Substituting this decomposition into \(\widehat{\mathfrak R}_{S}(\mathcal G_f)\)
and using subadditivity of the supremum gives
\[
\begin{aligned}
\widehat{\mathfrak R}_{S}(\mathcal G_f)
&\le
\Ex_\sigma
\sup_{a\in\cC_{\pm},\,h\in\cH}
\left|
\frac1n\sum_{i=1}^n
\sigma_i a(X_i)e_m^h(f(X_i))r_i
\right| \\
&\quad+
\sum_{k=1}^m
\Ex_\sigma
\sup_{a\in\cC_{\pm},\,h\in\cH}
\frac1n\sum_{i=1}^n
\sigma_i a(X_i)d_k^h(f(X_i))r_i .
\end{aligned}
\]
Indeed, this is the elementary inequality
\[
\sup_{a,h}\left\{R_{a,h}+\sum_{k=1}^m I_{k,a,h}\right\}
\le
\sup_{a,h}|R_{a,h}|+
\sum_{k=1}^m\sup_{a,h}I_{k,a,h}.
\]
The residual satisfies \(\|e_m^h\|_\infty\le\varepsilon_m\le\delta\), and
therefore
\[
\Ex_\sigma
\sup_{a,h}
\left|
\frac1n\sum_{i=1}^n
\sigma_i a(X_i)e_m^h(f(X_i))r_i
\right|
\le
\delta,
\]
because \(|a(X_i)|\le1\) and \(|r_i|\le1\).

For the \(k\)-th increment, since \(d_k^h\in\mathcal D_k\),
\[
\sup_{a,h}
\frac1n\sum_{i=1}^n
\sigma_i a(X_i)d_k^h(f(X_i))r_i
\le
\max_{d\in\mathcal D_k}
\sup_{a\in\cC_{\pm}}
\frac1n\sum_{i=1}^n
\sigma_i a(X_i)d(f(X_i))r_i .
\]
Define the sample-dependent multiplier vector \(b_i^d:=d(f(X_i))r_i,
\qquad d\in\mathcal D_k,\ i\in[n],\) and set
\[
T_k
:=
\Ex_\sigma
\max_{d\in\mathcal D_k}
\sup_{a\in\cC_{\pm}}
\frac1n\sum_{i=1}^n
\sigma_i a(X_i)b_i^d .
\]
Combining the preceding displays gives \(\widehat{\mathfrak R}_{S}(\mathcal G_f)
\le
\delta+
\sum_{k=1}^mT_k.\) Thus it remains to bound each \(T_k\).

The size of the multiplier vector is controlled by the scale of the
corresponding witness increment. If \(d=d_k^h=h_k(h)-h_{k-1}(h)\), then for
\(k\ge2\),
\[
\|d\|_\infty
\le
\|h_k(h)-h\|_\infty+
\|h-h_{k-1}(h)\|_\infty
\le
\varepsilon_k+
\varepsilon_{k-1}
\le
2\varepsilon_{k-1}.
\]
For \(k=1\), since \(h_0(h)\equiv0\) and \(|h_1(h)|\le1=\varepsilon_0\), the
same bound holds up to a universal constant. Hence, with \(B_k
:=
\sup_{d\in\mathcal D_k}\max_{i\in[n]} |b_i^d|,\) we have \(B_k\lesssim \varepsilon_{k-1}.\)

Fix \(k\) and first consider a single increment \(d\in\mathcal D_k\). Define \(Z_d(\sigma)
:=
\sup_{a\in\cC_{\pm}}
\frac1n\sum_{i=1}^n
\sigma_i a(X_i)b_i^d .\) Because \(\cC_{\pm}\) is symmetric, this supremum is the same as the
corresponding absolute supremum. For fixed \(d\), the coordinatewise maps \(\phi_i(t):=b_i^d t\) satisfy \(\phi_i(0)=0\) and \(|\phi_i(t)-\phi_i(t')|
\le
|b_i^d|\,|t-t'|
\le
B_k|t-t'|.\) The Rademacher contraction inequality gives
\[
\begin{aligned}
\Ex_\sigma Z_d(\sigma)
&\le
\Ex_\sigma
\sup_{a\in\cC_{\pm}}
\left|
\frac1n\sum_{i=1}^n
\sigma_i b_i^d a(X_i)
\right| \\
&\le
C_{\rm ctr}B_k
\Ex_\sigma
\sup_{a\in\cC_{\pm}}
\left|
\frac1n\sum_{i=1}^n
\sigma_i a(X_i)
\right| \\
&=
C_{\rm ctr}B_k\,\widehat{\mathfrak R}_{X}(\cC_{\pm}),
\end{aligned}
\]
where \(C_{\rm ctr}\) is a universal contraction constant; the last equality
again uses the symmetry of \(\cC_{\pm}\).

We next take the maximum over the finite set \(\mathcal D_k\). For fixed
\(d\), changing one Rademacher sign \(\sigma_j\) changes \(Z_d(\sigma)\) by at
most \(\frac{2|b_j^d|}{n}
\le
\frac{2B_k}{n}.\) Hence the bounded-differences inequality gives, for every
\(\lambda\in\mathbb R\),
\[
\Ex_\sigma
\exp\left\{
\lambda\bigl(Z_d(\sigma)-\Ex_\sigma Z_d(\sigma)\bigr)
\right\}
\le
\exp\left\{
\frac{\lambda^2B_k^2}{2n}
\right\}.
\]
Therefore, for every \(\lambda>0\),
\[
\begin{aligned}
\Ex_\sigma
\max_{d\in\mathcal D_k}
\{Z_d(\sigma)-\Ex_\sigma Z_d(\sigma)\}
&\le
\frac1\lambda
\log
\sum_{d\in\mathcal D_k}
\Ex_\sigma
\exp\left\{
\lambda\bigl(Z_d(\sigma)-\Ex_\sigma Z_d(\sigma)\bigr)
\right\} \\
&\le
\frac{\log M_k}{\lambda}
+
\frac{\lambda B_k^2}{2n}.
\end{aligned}
\]
Optimizing over \(\lambda>0\) yields
\[
\Ex_\sigma
\max_{d\in\mathcal D_k}
\{Z_d(\sigma)-\Ex_\sigma Z_d(\sigma)\}
\le
B_k\sqrt{\frac{2\log M_k}{n}},
\]
with the convention that this term is zero when \(M_k=1\). Consequently,
\[
\begin{aligned}
T_k
&=
\Ex_\sigma\max_{d\in\mathcal D_k}Z_d(\sigma) \\
&\le
\max_{d\in\mathcal D_k}\Ex_\sigma Z_d(\sigma)
+
\Ex_\sigma
\max_{d\in\mathcal D_k}
\{Z_d(\sigma)-\Ex_\sigma Z_d(\sigma)\} \\
&\le
C_{\rm ctr}B_k\,\widehat{\mathfrak R}_{X}(\cC_{\pm})
+
B_k\sqrt{\frac{2\log M_k}{n}}.
\end{aligned}
\]
Using \(B_k\lesssim\varepsilon_{k-1}\) and the cardinality bound on
\(M_k\),
\[
T_k
\lesssim
\varepsilon_{k-1}\widehat{\mathfrak R}_{X}(\cC_{\pm})
+
\varepsilon_{k-1}
\sqrt{
\frac{
\log N(\varepsilon_k,\cH,\|\cdot\|_\infty)
}{n}
}.
\]

Summing over \(k=1,\ldots,m\) gives
\[
\widehat{\mathfrak R}_{S}(\mathcal G_f)
\lesssim
\delta
+
\left(\sum_{k=1}^m\varepsilon_{k-1}\right)
\widehat{\mathfrak R}_{X}(\cC_{\pm})
+
\sum_{k=1}^m
\varepsilon_{k-1}
\sqrt{
\frac{
\log N(\varepsilon_k,\cH,\|\cdot\|_\infty)
}{n}
}.
\]
The first sum is bounded by a universal constant, because \(\sum_{k=1}^m\varepsilon_{k-1}
=
\sum_{k=1}^m2^{-(k-1)}
\le2.\)

It remains to compare the dyadic entropy sum with the Dudley integral. Set \(F(\varepsilon)
:=
\sqrt{
\frac{\log N(\varepsilon,\cH,\|\cdot\|_\infty)}{n}
}.\) Because covering numbers are nonincreasing in the radius, \(F\) is
nonincreasing as a function of \(\varepsilon\). For each \(k\), \(\varepsilon_{k-1}F(\varepsilon_k)
=
4(\varepsilon_k-\varepsilon_{k+1})F(\varepsilon_k).\) Moreover, for \(\varepsilon\in[\varepsilon_{k+1},\varepsilon_k]\), we have
\(F(\varepsilon)\ge F(\varepsilon_k)\). Hence \(\varepsilon_{k-1}F(\varepsilon_k)
\le
4\int_{\varepsilon_{k+1}}^{\varepsilon_k}F(\varepsilon)\,d\varepsilon .\) Summing over \(k=1,\ldots,m\) gives \(\sum_{k=1}^m\varepsilon_{k-1}F(\varepsilon_k)
\lesssim
\int_{\varepsilon_{m+1}}^1F(\varepsilon)\,d\varepsilon .\) By the choice of \(m\), \(\varepsilon_m\le\delta<2\varepsilon_m\), and hence
\(\varepsilon_{m+1}=\varepsilon_m/2\ge\delta/4\). Therefore \(\int_{\varepsilon_{m+1}}^1F(\varepsilon)\,d\varepsilon
\le
\int_{\delta/4}^1F(\varepsilon)\,d\varepsilon .\) Combining the preceding displays,
\[
\widehat{\mathfrak R}_{S}(\mathcal G_f)
\lesssim
\widehat{\mathfrak R}_{X}(\cC_{\pm})
+
\delta
+
\int_{\delta/4}^{1}
\sqrt{
\frac{\log N(\varepsilon,\cH,\|\cdot\|_\infty)}{n}
}\,d\varepsilon .
\]
Renaming the lower limit and absorbing universal constants into
\(\lesssim\), we obtain
\[
\widehat{\mathfrak R}_{S}(\mathcal G_f)
\lesssim
\widehat{\mathfrak R}_{X}(\cC_{\pm})
+
\inf_{\delta\in[0,1]}
\left\{
\delta+
\int_\delta^1
\sqrt{
\frac{\log N(\varepsilon,\cH,\|\cdot\|_\infty)}{n}
}\,d\varepsilon
\right\}.
\]
For large truncation levels, the same display follows from the trivial bound
\(\widehat{\mathfrak R}_{S}(\mathcal G_f)\le1\) after adjusting the universal
constant, so the infimum may be taken over the full interval \([0,1]\).

Finally, taking expectation over \(S\sim\cald^n\) proves
\eqref{eq:mc_fixed_emp_process_separable_app}. Combining this with
\eqref{eq:mc_fixed_emp_process_general_app} proves
\eqref{eq:mc_fixed_emp_process_bound_app}.
\end{proof}

\subsection{Convergence of smooth CE}\label{app:mc_smooth_ce_refinement} 

\begin{lemma}[Lipschitz-witness refinement for finite subgroup classes]
\label{lem:mc_smooth_ce_refinement}
Assume $\cH=\operatorname{Lip}_1([0,1],[-1,1])$ and $\cC$ is finite. For every fixed predictor $f:\calx\to[0,1]$,
\begin{align}
\UC_n^{\MC}(f)
&\lesssim
\frac{1}{\sqrt n}
+
\sqrt{\frac{\log(2|\cC|)}{n}}.
\label{eq:mc_smooth_ce_refinement_app}
\end{align}
The hidden constant is universal.
\end{lemma}

\begin{proof}
We give a direct entropy proof, avoiding any dependence on a finite discretization of the Lipschitz witness class.  Lemma~\ref{lem:mc_fixed_emp_process} gives
\[
\UC_n^{\MC}(f)
\lesssim
\mathfrak R_{\cald,n}(\cC_{\pm})
+
\inf_{\delta\ge0}
\left\{
\delta+
\int_\delta^1
\sqrt{\frac{\log N(\varepsilon,\operatorname{Lip}_1([0,1],[-1,1]),\|\cdot\|_\infty)}{n}}\,d\varepsilon
\right\}.
\]
Because $|\cC_{\pm}|\le2|\cC|$ and all functions in $\cC_{\pm}$ are bounded by one, Massart's finite-class lemma gives \(\mathfrak R_{\cald,n}(\cC_{\pm})
\lesssim
\sqrt{\frac{\log(2|\cC|)}{n}}.\) It remains to evaluate the entropy integral.  A standard grid construction gives
\[
\log N(\varepsilon,\operatorname{Lip}_1([0,1],[-1,1]),\|\cdot\|_\infty)
\lesssim
\frac{1}{\varepsilon}\log\frac{1}{\varepsilon},
\qquad 0<\varepsilon\le1.
\]
Indeed, take a mesh of spacing proportional to $\varepsilon$, quantize the function values on the mesh at resolution proportional to $\varepsilon$, and extend between mesh points using the Lipschitz constraint.  Therefore
\[
\int_0^1
\sqrt{\frac{\log N(\varepsilon,\operatorname{Lip}_1([0,1],[-1,1]),\|\cdot\|_\infty)}{n}}\,d\varepsilon
\lesssim
\frac{1}{\sqrt n}
\int_0^1 \sqrt{\frac{\log(1/\varepsilon)}{\varepsilon}}\,d\varepsilon
\lesssim \frac{1}{\sqrt n}.
\]
Substituting these two estimates into Lemma~\ref{lem:mc_fixed_emp_process} proves \eqref{eq:mc_smooth_ce_refinement_app}.
\end{proof}
This order matches the results in \citet{futami2026smooth}.

\begin{proof}[Proof of Corollary~\ref{cor:mc_population_validation}: MC population--training bound]
Conditioning on the learned predictor and applying Lemma~\ref{lem:mc_fixed_emp_process} gives
\begin{align}
\Ex\left|
\Phi_{\MC}(f_W;\cald)-\Phi_{\MC,n}(f_W;S')
\right|
\le
\Ex\UC_n^{\MC}(f_W).
\label{eq:mc_uc_intermediate_app}
\end{align}
Now decompose
\begin{align}
\Phi_{\MC}(f_W;\cald)-\Phi_{\MC,n}(f_W;S)
&=
\{\Phi_{\MC}(f_W;\cald)-\Phi_{\MC,n}(f_W;S')\}
\notag\\
&\quad+
\{\Phi_{\MC,n}(f_W;S')-\Phi_{\MC,n}(f_W;S)\}.
\end{align}
Taking absolute values and expectations, the first term is bounded by \(\Ex\UC_n^{\MC}(f_W)\).  The second term is bounded by  the bound on \(\Ex|\Delta_\Phi|\) in \eqref{eq:absolute_val_train_gap} with \(L_{\Gamma,n}^{\MC}\le2/\sqrt n\).  This gives
\begin{align}
\Ex\bigl|
\Phi_{\MC}(f_W;\cald)-\Phi_{\MC,n}(f_W;S)
\bigr|
&\le
\Ex\UC_n^{\MC}(f_W)
+
\frac{2}{\sqrt n}
\sqrt{2\{I(f_W(\widetilde{X});U\mid\widetilde{Z})+\log 2\}},
\end{align}
which is \eqref{eq:mc_abs_cor}.

Finally, we prove the one-sided comparison.  Since MC is a supremum over witnesses with unbiased empirical witness averages, Jensen's inequality gives
\begin{align}
\Phi_{\MC}(f;\cald)
\le
\Ex[\Phi_{\MC,n}(f;S')\mid f]
\end{align}
for every fixed \(f\).  Applying this with \(f=f_W\) and subtracting \(\Ex\Phi_{\MC,n}(f_W;S)\), we obtain
\begin{align}
\Ex\Phi_{\MC}(f_W;\cald)-\Ex\Phi_{\MC,n}(f_W;S)
&\le
\Ex[\Phi_{\MC,n}(f_W;S')-\Phi_{\MC,n}(f_W;S)].
\label{eq:mc_one_sided_to_vt_app}
\end{align}
Let \(D_t^{\MC}\) denote the MC increment and set \(I_{\mathrm{loc},t}^{\MC}\coloneqq I(D_t^{\MC};U_t\mid\widetilde Z)\). Since Corollary~\ref{cor:mc_stability} gives \(|D_t^{\MC}|\le 2/n\), Theorem~\ref{thm:wang_general} yields
\begin{align}
\Ex[\Phi_{\MC,n}(f_W;S')-\Phi_{\MC,n}(f_W;S)]
&\le
\frac{2}{n}
\min\Bigl\{
\sqrt{2nI_{\mathrm{fCMI}}},
\sum_{t=1}^n\sqrt{2I_{\mathrm{loc},t}^{\MC}}
\Bigr\}.
\label{eq:mc_one_sided_min_vt_app}
\end{align}
Combining \eqref{eq:mc_one_sided_to_vt_app} and \eqref{eq:mc_one_sided_min_vt_app} proves \eqref{eq_onesided_MC}.
\end{proof}

\subsubsection{One-sided comparisons for common-feasible-set supremum objectives}
\label{app:one_sided_common_feasible_appendix}

The one-sided argument used for multicalibration is a general Jensen argument. It is not applied to the primal infimum form of CVaR; for CVaR and CR DRO we use dual common-feasible-set forms or the absolute plug-in comparisons above. Let
\begin{align}
\Phi(f;\cald)=\sup_{\pi\in\Pi}\Ex_{Z\sim\cald} g_{f,\pi}(Z),
\qquad
\Phi_n(f;S')=\sup_{\pi\in\Pi}\frac1n\sum_{i=1}^n g_{f,\pi}(Z'_i),
\label{eq:common_feasible_supremum_app}
\end{align}
where \(S'\sim\cald^n\).  Write \(P'_n g_{f,\pi}\coloneqq n^{-1}\sum_{i=1}^n g_{f,\pi}(Z'_i)\). Then
\begin{align}
\Phi(f;\cald)
&=
\sup_{\pi\in\Pi}\Ex_{S'\sim\cald^n} P'_n g_{f,\pi}
\le
\Ex_{S'\sim\cald^n}\sup_{\pi\in\Pi} P'_n g_{f,\pi}
=
\Ex_{S'\sim\cald^n}\Phi_n(f;S').
\label{eq:common_feasible_one_sided_app}
\end{align}
Consequently, for any learned predictor \(f_W\),
\begin{align}
\Ex \Phi(f_W;\cald)-\Ex\Phi_n(f_W;S)
&\le
\Ex\{\Phi_n(f_W;S')-\Phi_n(f_W;S)\}.
\label{eq:common_feasible_one_sided_vt_app}
\end{align}
Combining \eqref{eq:common_feasible_one_sided_vt_app} with either  the global bound in Theorem~\ref{thm:validation_training_gap} or the single-index refinement yields the corresponding signed population--training upper bounds. This applies directly to multicalibration and to any dual CVaR/DRO formulation written with a common feasible supremum class. This Jensen comparison is not applied to the primal infimum form of CVaR; for the main CVaR statement we use the absolute plug-in comparison in Corollary~\ref{cor:cvar_population_validation}. When the finite-sample DRO constraint itself is empirical, the absolute plug-in comparisons in Corollaries~\ref{cor:cvar_population_validation} and~\ref{cor:dro_population_validation} are the statements used in the main text.

\subsection{Growth-function control of global fCMI}
\label{app:vc_fcmi_growth}

The VC-rate statements in the main text use the following standard growth-function observation.

\begin{lemma}[VC growth-function bound for global fCMI]
\label{lem:vc_fcmi_growth}
Assume \(\mathcal V=\{0,1\}\) and that the learning algorithm outputs predictors in a binary class \(\mathcal F\subseteq\{0,1\}^{\mathcal X}\) with VC dimension \(d<\infty\). Let \(d_n\coloneqq d\wedge 2n\). Then
\begin{align}
I(f_W(\widetilde{X});U\mid\widetilde{Z})
\le
H(f_W(\widetilde{X})\mid\widetilde{Z})
\le
\log \Pi_{\mathcal F}(2n)
\le
 d_n\log\left(\frac{2en}{d_n}\right),
\label{eq:vc_fcmi_growth}
\end{align}
where \(\Pi_{\mathcal F}(m)\) is the growth function of \(\mathcal F\). Consequently, if \(d\le 2n\), then \(I(f_W(\widetilde{X});U\mid\widetilde{Z})\lesssim d\log(en/d)\).
\end{lemma}

\begin{proof}
Condition on a supersample realization \(\widetilde{Z}=\widetilde{z}\).  The prediction table \(f_W(\widetilde{X})\) is one of the binary labelings of the \(2n\) supersample inputs induced by \(\mathcal F\). Hence its conditional support size is at most \(\Pi_{\mathcal F}(2n)\), regardless of the learning algorithm and its internal randomness. Therefore \(I(f_W(\widetilde{X});U\mid\widetilde{Z}=\widetilde{z})
\le
H(f_W(\widetilde{X})\mid\widetilde{Z}=\widetilde{z})
\le
\log \Pi_{\mathcal F}(2n).\) Averaging over \(\widetilde{Z}\) gives the first two inequalities.  The trivial bound gives \(\Pi_{\mathcal F}(2n)\le2^{2n}\), while Sauer--Shelah gives \(\Pi_{\mathcal F}(2n)\le (2en/d)^d\) when \(1\le d\le 2n\). These two cases are summarized by the displayed bound with \(d_n=d\wedge2n\).
\end{proof}

\citet{harutyunyan2021} show that
\begin{align}
I(f_W(\widetilde{X});U\mid\widetilde{Z}=\widetilde{z})
\leq
\max\{(d + 1)\log2,d \log (2en/d) \}.
\label{eq:batch-ecmi-vc}
\end{align}

\subsection{Complete statements}

\begin{corollary}[CVaR via stability]
\label{cor:cvar_stability_complete}
Assume \(0\le \ell(f_w(X),Y)\le1\) for all predictors and data points, and write \(\bar\alpha\coloneqq 1-\alpha\). For the CVaR objective~\eqref{eq:cvar_empirical}, Assumption~\ref{ass:one_point_stability} holds with \(\beta_n=\beta_{\CVaR,n}\coloneqq(\bar\alpha n)^{-1}\). Hence
\begin{align}
L_{\Gamma,n}^{\CVaR,\mathrm{stab}}
&\le\frac{1}{\bar\alpha\sqrt n},
\label{eq:cvar_lgamma_stability_main}
\end{align}
and, writing \(D_t^{\CVaR}\) for the local increment in Theorem~\ref{thm:wang_general} specialized to CVaR, set \(I_{\mathrm{loc},t}^{\CVaR}\coloneqq I(D_t^{\CVaR};U_t\mid\widetilde Z)\). Then
\begin{align}
|\Ex\Delta_{\CVaR}|
&\le
\frac{1}{\bar\alpha n}
\min\Bigl\{
\sqrt{2nI_{\mathrm{fCMI}}},
\sum_{t=1}^n\sqrt{2I_{\mathrm{loc},t}^{\CVaR}}
\Bigr\}.
\label{eq:cvar_vt_stability_main}
\end{align}
\end{corollary}

\begin{corollary}[Suboptimal Cressie--Read DRO bound via stability]
\label{cor:dro_stability_suboptimal_complete}
Assume \(0\le \ell(f_w(X),Y)\le1\) for all predictors and data points. Let \(c>1\), \(c^\ast=c/(c-1)\), and \(\kappa_{c,\epsilon}\coloneqq \{1+c(c-1)\epsilon\}^{1/c}\). For the empirical Cressie--Read DRO objective~\eqref{eq:cr_dro_empirical_prelim}, Assumption~\ref{ass:one_point_stability} holds with
\begin{align}
\beta_{\DRO,n}^{\mathrm{stab}}
&\le
\kappa_{c,\epsilon}n^{1/c-1}.
\label{eq:dro_stability_beta_main}
\end{align}
Consequently,
\begin{align}
L_{\Gamma,n}^{\DRO,\mathrm{stab}}
&\le
\kappa_{c,\epsilon}n^{1/c-1/2},
\label{eq:dro_stability_lgamma_main}
\end{align}
and, writing \(D_t^{\DRO}\) for the local increment in Theorem~\ref{thm:wang_general} specialized to DRO, set \(I_{\mathrm{loc},t}^{\DRO}\coloneqq I(D_t^{\DRO};U_t\mid\widetilde Z)\). Then
\begin{align}
|\Ex\Delta_{\DRO}|
&\le
\kappa_{c,\epsilon}n^{1/c-1}
\min\Bigl\{
\sqrt{2nI_{\mathrm{fCMI}}},
\sum_{t=1}^n\sqrt{2I_{\mathrm{loc},t}^{\DRO}}
\Bigr\}.
\label{eq:dro_stability_vt_main}
\end{align}
\end{corollary}

\Needspace{5\baselineskip}

\subsection{Comparison with existing bounds}
\label{app:comparison_summary}
Table~\ref{tab:full_objective_comparison} compares the full population--training bounds, including both fixed-predictor and learner-dependent terms, with specialized existing results.
Here \(I_{\rm ECE}\) is the ECE-specific loss-CMI, and \(\delta\) is the failure probability in the CVaR guarantee. In the CVaR row, \(\mathfrak R_{\cald,n}\) is evaluated on the loss class \(\{\ell(f,\cdot)\}_{f\in\calf}\).
The CR DRO constant \(\mathsf K_{c,\epsilon}\) is specified in Corollary~\ref{cor:dro_population_validation}.

\suppressfloats[t]
\begin{table}[t]
\centering\small

\caption{Full-bound comparison with specialized results.}
\label{tab:full_objective_comparison}
\setlength{\tabcolsep}{1.5pt}
\thinmuskip=1mu\relax
\medmuskip=2mu\relax
\thickmuskip=2mu\relax
\renewcommand{\arraystretch}{1.25}
\begin{tabular*}{\linewidth}{@{\extracolsep{\fill}}lll@{}}
\toprule
\textbf{Objective} & \textbf{Existing} & \textbf{Ours} \\
\midrule
\addlinespace[5pt]
Binning ECE
& \(\displaystyle\sqrt{8(I_{\rm ECE}+B\log2)}n^{-1/2}\)\;\citep{futami2024information}
& \(\displaystyle\sqrt{2B\log2}n^{-1/2}+\sqrt{8(I_{\mathrm{fCMI}}+\log2)}n^{-1/2}\) \\
\addlinespace[10pt]
CVaR
& \(\displaystyle \bar\alpha^{-1}O(\mathfrak R_{\cald,n}+\sqrt{n^{-1}\log\delta^{-1}})\)\;\citep{lee2020risksensitive}
& \(\displaystyle\bar\alpha^{-1}n^{-1/2}\bigl[\sqrt{\pi/2}+\min\{1,2\sqrt{\bar\alpha}\}\sqrt{2(I_{\mathrm{fCMI}}+\log2)}\bigr]\) \\
\addlinespace[10pt]
CR DRO
& \(\displaystyle\widetilde O(\text{}\mathsf K_{c,\epsilon}n^{-\frac{1}{c^\ast\vee2}})\)\;\citep{Duchi2021}
& \(\displaystyle\widetilde O(\text{}\mathsf K_{c,\epsilon}n^{-\frac{1}{c^\ast\vee2}})+2\text{}\kappa_{c,\epsilon}n^{-\frac{1}{c^\ast\vee2}}\sqrt{2(I_{\mathrm{fCMI}}+\log2)}\) \\
\addlinespace[5pt]
\bottomrule
\end{tabular*}

\end{table}

\paragraph{Multicalibration and binning ECE.} For fixed bins \(J_1,\ldots,J_B\), choosing \(\mathcal C=\{1\}\) and \(\mathcal H_{\rm ECE}=\{h_a(v)=\sum_{b=1}^B a_b\mathbf1\{v\in J_b\}:a\in\{-1,+1\}^B\}\) makes our MC objective equal binning ECE.  The bound of \citet{futami2024information} combines this information quantity with bin complexity; our bound in Table~\ref{tab:full_objective_comparison} separates fixed-predictor and learner-dependent terms and removes explicit \(B\) from the latter. Since the information quantities differ, neither bound uniformly dominates; for fixed \(B\) and \(O(1)\) information, both scale as \(n^{-1/2}\). The comparison with smooth calibration by \citet{futami2026smooth} concerns only the fixed-predictor term in Appendix~\ref{app:mc_smooth_ce_refinement}.

\paragraph{CVaR and tail mass.} \citet{lee2020risksensitive} give a high-probability excess-risk bound  for the empirical OCE minimizer, whereas our full bound in Table~\ref{tab:full_objective_comparison} controls the expected population--training gap for an arbitrary learner. Our improvement to \(\bar\alpha^{-1/2}\) applies only to the validation--training term; the fixed-predictor term retains \(\bar\alpha^{-1}\).
The \(\bar\alpha^{-1/2}\) dependence is worst-case tight for the convex-\(L_2\) certificate: ignoring integer rounding, a feasible vector assigning weight \(1/\bar\alpha\) to \(n\bar\alpha\) tail observations has \(\|q\|_2^2=n/\bar\alpha\). This concerns the certificate, not a minimax lower bound for all CVaR estimators. The validation--training term vanishes when \(I_{\mathrm{fCMI}}+\log2=o(n\bar\alpha)\); the full bound~\eqref{eq:cvar_population_bound_main} additionally requires \(\bar\alpha\sqrt n\to\infty\).
The PAC-Bayes CVaR bound of \citet{mhammedi2020cvar} has leading error \(\widetilde O(\sqrt{\widehat{\operatorname{CVaR}}\,K/(n\bar\alpha)}+K/(n\bar\alpha))\), where \(\widehat{\operatorname{CVaR}}\) is the empirical CVaR and \(K\) contains the posterior--prior KL and confidence terms. Its worst-case order has the same effective-tail scaling as our convex-\(L_2\) validation--training term. We do not claim that our full bound improves the existing high-probability guarantees.

\paragraph{CR DRO.}
The first term in our bound is the fixed-predictor population--validation plug-in error from \citet{Duchi2021}.
The second is the validation--training error due to the learned predictor's adaptivity.
The convex \(L_2\) geometry of the adversarial weights yields the same \(n^{-1/(c^\ast\vee2)}\) exponent, with an additional fCMI factor; if \(I_{\mathrm{fCMI}}=O(1)\), both terms retain that exponent.
In Corollaries~\ref{cor:dro_l2} and~\ref{cor:dro_population_validation}, the exponent is \(r_c=(c-1)/c\) for \(1<c<2\) and \(r_c=1/2\) for \(c\ge2\). The limit \(r_c\to0\) as \(c\downarrow1\) does not give a rate for exact KL-DRO. For fixed \(c>1\), increasing the positive radius \(\epsilon\) increases the coefficient \(\kappa_{c,\epsilon}\) but leaves \(r_c\) unchanged; the fixed-predictor constant \(\mathsf K_{c,\epsilon}\) need not be monotone. Equal numerical radii for different \(c\) do not specify equal ambiguity sets.
Thus, unification means this decomposition and recovery of known rates, not uniform improvement.

\FloatBarrier

\section{Discussion about measurability of optimized set-input values}
\label{app:measurability_set_input}
This paper defines set-input objectives in a general form, but our concrete targets are the MC, CVaR, and DRO metrics. For these objectives, measurability has already been studied in the existing literature. On the other hand, when considering more general set-input objectives, one needs to account for measurability issues as follows.

The main text works with finite real-valued optimized objectives.  We make the measurability convention explicit here, because  the proof of Theorem~\ref{thm:validation_training_gap} conditions on the entire supersample prediction table and then evaluates the optimized objective for all selector vectors.

If the auxiliary set $\Pi$ is finite or countable and each row map $z\mapsto \rho_\pi(f(x),z)$ is measurable, then all empirical and population optimized values are measurable.  For uncountable auxiliary classes, we use the following separable-version convention.  For each fixed predictor $f$, there is a countable subset $\Pi_0(f)\subseteq\Pi$, independent of the empirical realization, such that for every probability law $Q$ used in the argument, the population law, empirical laws, and mixed empirical laws arising in the supersample interpolation,
\begin{align}
\operatorname{Opt}_{\pi\in\Pi}
\psi_\pi\!\left(\int \rho_\pi(f(x),z)\,dQ(z)\right)
=
\operatorname{Opt}_{\pi\in\Pi_0(f)}
\psi_\pi\!\left(\int \rho_\pi(f(x),z)\,dQ(z)\right),
\label{eq:separable_version_convention_app}
\end{align}
where \(\operatorname{Opt}\in\{\sup,\inf\}\). The optimized value is then a countable supremum or infimum of measurable maps.  When the predictor is random, the convention is applied after conditioning on the realized predictor or, equivalently, to a measurable separable version of the optimized value.  This is sufficient for the conditional maps $u\mapsto \Phi_n(f_W;\widetilde Z_u)$ and $u\mapsto\Gamma_{f_W,\widetilde Z}(u)$ used in the proofs.

The convention covers finite auditor classes used for binned multicalibration.  For compact parametric or smooth witness classes, it follows from a countable dense subclass whenever the displayed integral and the post-aggregation map are continuous in the auxiliary parameter on the relevant feasible set.  If such a separable version is not available, the statements should be read with outer expectations or essential suprema/infima; we do not use that extra notation in the paper.

\section{Additional Discussion}\label{app:additional_refinements}
This appendix collects refinements and alternative comparisons that are not used in the main theorem.  They are included to clarify  the scope of Theorem~\ref{thm:validation_training_gap}, the relation to single-index fast-rate bounds for ordinary losses, and variants in which the validation set is replaced by a smaller subset.  Throughout this appendix we use the notation of Section~\ref{sec:main}: the local increments are the random variables \(D_t\) from Theorem~\ref{thm:wang_general}, and under uniform stability they satisfy \(|D_t|\le \beta_n\).

\subsection{The intuition behind the population--validation term}
\label{app:population_validation_discussion}
Another distinction concerns the role of the population--validation discrepancy. For ordinary
losses, the difference between population and validation evaluation is expressed through a mean, and hence there is no optimization mismatch. In our setting, however,
the object of interest is an optimized criterion, so the population objective and the finite validation
set objective can differ significantly. We therefore control the population--validation set difference
as a ``bias'' term by uniform convergence, since learning is not involved in that comparison.
By contrast, the validation--training set difference involves the learned predictor and is controlled
by IT generalization analysis. This decomposition prevents the objective-specific population--validation cost from being misattributed to the IT-controlled validation--training term.

\subsection{Conditioning and information relations for local increments}
\label{app:conditioning_information_relations}

The main single-index bound uses the unconditioned single-coordinate quantities
\(I(D_t;U_t\mid\widetilde Z)\).  This subsection records how these
quantities relate to the more strongly conditioned leave-one-out
quantities that sometimes arise in diagnostics.  Write \(F\coloneqq f_W(\widetilde X).\) Throughout this subsection, the selectors
\(U_1,\ldots,U_n\) are assumed to be conditionally independent given
\(\widetilde Z\).

\begin{lemma}[Unconditioned and leave-one-out local information]
\label{lem:conditioning_local_information}
Fix \(t\in[n]\).  Let \(D_t\) be a local increment satisfying \(D_t\in\sigma(F,\widetilde Z,U_{-t}).\) Here \(\sigma(F,\widetilde Z,U_{-t})\) denotes the sigma-algebra generated by \(F\), \(\widetilde Z\), and \(U_{-t}\), so \(D_t\in\sigma(F,\widetilde Z,U_{-t})\) means that \(D_t\) is measurable with respect to these variables.

Then
\begin{align}
I(D_t;U_t\mid\widetilde Z)
&\le
I(D_t;U_t\mid\widetilde Z,U_{-t}),
\label{eq:uncond_le_cond_local_info}\\
I(D_t;U_t\mid\widetilde Z,U_{-t})
&\le
I(F;U_t\mid\widetilde Z,U_{-t}).
\label{eq:cond_local_dpi_full_table}
\end{align}
\end{lemma}

\begin{proof}
For \eqref{eq:uncond_le_cond_local_info}, use the entropy form of
mutual information.  Since \(U_t\perp\!\!\!\perp U_{-t}\mid\widetilde Z\),
we have \(H(U_t\mid\widetilde Z,U_{-t})
=
H(U_t\mid\widetilde Z).\) Therefore
\begin{align*}
&I(D_t;U_t\mid\widetilde Z,U_{-t})
-
I(D_t;U_t\mid\widetilde Z)
\\
&\quad =
H(U_t\mid\widetilde Z,U_{-t})
-
H(U_t\mid D_t,\widetilde Z,U_{-t})
-
H(U_t\mid\widetilde Z)
+
H(U_t\mid D_t,\widetilde Z)
\\
&\quad =
H(U_t\mid D_t,\widetilde Z)
-
H(U_t\mid D_t,\widetilde Z,U_{-t})
\\
&\quad =
I(U_t;U_{-t}\mid D_t,\widetilde Z)
\ge 0.
\end{align*}
This proves \eqref{eq:uncond_le_cond_local_info}.

For \eqref{eq:cond_local_dpi_full_table}, condition on
\((\widetilde Z,U_{-t})\).  Under this conditioning, the assumption
\(D_t\in\sigma(F,\widetilde Z,U_{-t})\) says that \(D_t\) is a measurable
function of \(F\).  Hence, conditionally on \((\widetilde Z,U_{-t})\),
the variables form the Markov chain \(U_t \longrightarrow F \longrightarrow D_t .\) The conditional data-processing inequality gives \(I(D_t;U_t\mid\widetilde Z,U_{-t})
\le
I(F;U_t\mid\widetilde Z,U_{-t}),\) as claimed.
\end{proof}

\begin{lemma}[ Information about \(U\) and \(U_t\) in the predictions]
\label{lem:predictor_one_bit_information_relations}
Let \(F\) be any random element jointly distributed with
\((\widetilde Z,U)\).  Assume that the selectors
\(U_1,\ldots,U_n\) are conditionally independent given \(\widetilde Z\).
Then
\begin{align}
\sum_{t=1}^n I(F;U_t\mid\widetilde Z)
\le
I(F;U\mid\widetilde Z)
\le
\sum_{t=1}^n I(F;U_t\mid\widetilde Z,U_{-t}).
\label{eq:predictor_information_sandwich}
\end{align}
\end{lemma}

\begin{proof}
For the first inequality, use the chain rule with
\(U_{<t}\coloneqq (U_1,\ldots,U_{t-1})\): \(I(F;U\mid\widetilde Z)
=
\sum_{t=1}^n I(F;U_t\mid\widetilde Z,U_{<t}).\) Since \(U_t\perp\!\!\!\perp U_{<t}\mid\widetilde Z\), \(H(U_t\mid\widetilde Z,U_{<t})
=
H(U_t\mid\widetilde Z).\) Thus
\begin{align*}
I(F;U_t\mid\widetilde Z,U_{<t})
&=
H(U_t\mid\widetilde Z)
-
H(U_t\mid F,\widetilde Z,U_{<t})
\\
&\ge
H(U_t\mid\widetilde Z)
-
H(U_t\mid F,\widetilde Z)
\\
&=
I(F;U_t\mid\widetilde Z),
\end{align*}
where the inequality is the monotonicity of conditional entropy under
additional conditioning.  Summing over \(t\) proves the first inequality.

For the second inequality, expand both sides in entropy form.  Conditional
independence gives
\begin{align*}
\sum_{t=1}^n I(F;U_t\mid\widetilde Z,U_{-t})
&=
\sum_{t=1}^n H(U_t\mid\widetilde Z)
-
\sum_{t=1}^n H(U_t\mid F,\widetilde Z,U_{-t}),
\end{align*}
whereas
\[
I(F;U\mid\widetilde Z)
=
\sum_{t=1}^n H(U_t\mid\widetilde Z)
-
H(U\mid F,\widetilde Z).
\]
Therefore
\begin{align*}
&\sum_{t=1}^n I(F;U_t\mid\widetilde Z,U_{-t})
-
I(F;U\mid\widetilde Z)
\\
&\quad =
H(U\mid F,\widetilde Z)
-
\sum_{t=1}^n H(U_t\mid F,\widetilde Z,U_{-t}).
\end{align*}
With \(U_{>t}\coloneqq (U_{t+1},\ldots,U_n)\), the last display equals \(\sum_{t=1}^n
I(U_t;U_{<t}\mid F,\widetilde Z,U_{>t}),\) which is the conditional dual total correlation of \(U\) given
\((F,\widetilde Z)\).  It is nonnegative, proving the second inequality.
\end{proof}

The first lemma explains why the unconditioned local information used in
the main theorem is no larger than the leave-one-out conditional diagnostic: \(I(D_t;U_t\mid\widetilde Z)
\le
I(D_t;U_t\mid\widetilde Z,U_{-t}).\) Moreover, under the mixed-path measurability condition
\(D_t\in\sigma(F,\widetilde Z,U_{-t})\), this leave-one-out diagnostic is
controlled by the corresponding leave-one-out single-coordinate information of the
full prediction table: \(I(D_t;U_t\mid\widetilde Z,U_{-t})
\le
I(F;U_t\mid\widetilde Z,U_{-t}).\) However, these inequalities do not identify the unconditioned local term
with the unconditioned single-coordinate table information
\(I(F;U_t\mid\widetilde Z)\), nor do they imply that the sum of local
terms  is bounded by the CMI for the full vector \(U\), namely
\(I(F;U\mid\widetilde Z)\).

\begin{remark}[ Conditioning on the supersample alone]
There is no general ordering between
\(I(D_t;U_t\mid\widetilde Z)\) and
\(I(F;U_t\mid\widetilde Z)\) under the mixed-path definition.  The
obstruction is that \(D_t\) is a function of
\((F,\widetilde Z,U_{-t})\), not necessarily of \((F,\widetilde Z)\)
alone.  Thus, after conditioning only on \(\widetilde Z\), the Markov
chain \(U_t \longrightarrow F \longrightarrow D_t\) need not hold. Consequently, the preceding inequalities do not imply a general bound of
the form \(\sum_{t=1}^n I(D_t;U_t\mid\widetilde Z)
\le
I(F;U\mid\widetilde Z).\)  The global fCMI \(I(F;U\mid\widetilde Z)\) remains the natural quantity for VC and
growth-function arguments, whereas the one-dimensional local terms are
mainly useful as estimable diagnostics unless additional locality
structure is imposed on the increments.
\end{remark}

\subsection{Bernstein-type local refinements}
\label{app:fast_rate_discussion_added}

Theorem~\ref{thm:wang_general} uses a Hoeffding-type bound for the
one-dimensional variables \(\varepsilon_tD_t\).  Since
\(|D_t|\le \beta_n\), a Bernstein-type comparison is also available.
The comparison below is intentionally kept as a diagnostic statement:
it does not by itself imply a fast rate, and in the present conditional
Rademacher setting it is not the sharpest possible local bound.  Its role
is to identify the additional variance control that would be needed for a
Bernstein-style route to be useful.

For each \(t\in[n]\), define \(I_t\coloneqq I(D_t;U_t\mid\widetilde{Z}),
\qquad
\bar\nu_t\coloneqq \Ex[D_t^2].\)

\begin{proposition}[Bernstein-type local comparison]
\label{prop:bernstein_local_refinement_added}
Under Assumption~\ref{ass:one_point_stability}, for every \(t\in[n]\)
and every \(0<\lambda<3/\beta_n\),
\begin{align}
\bigl|\Ex[\varepsilon_tD_t]\bigr|
&\le
\frac{I_t}{\lambda}
+
\frac{\lambda\bar\nu_t}{2(1-\lambda\beta_n/3)}.
\label{eq:bernstein_local_unoptimized_added}
\end{align}
Consequently,
\begin{align}
\bigl|\Ex[\varepsilon_tD_t]\bigr|
&\le
\sqrt{2\bar\nu_t I_t}
+
\frac{\beta_n}{3}I_t.
\label{eq:bernstein_local_optimized_added}
\end{align}
Therefore,
\begin{align}
\left|\Ex\bigl[\Phi_n(f_W;S')-\Phi_n(f_W;S)\bigr]\right|
&\le
\sum_{t=1}^n
\left(
\sqrt{2\bar\nu_t I_t}
+
\frac{\beta_n}{3}I_t
\right).
\label{eq:bernstein_global_refinement_added}
\end{align}
\end{proposition}

\begin{proof}
Fix \(t\).  Conditionally on \(\widetilde Z\), use as reference measure \(P_{D_t\mid \widetilde Z}\otimes Q_t,
\qquad
Q_t=\mathrm{Unif}(\{0,1\}).\) Equivalently, under the reference measure, \(D_t\) keeps its actual
conditional marginal distribution, while \(U_t\) is resampled independently
from \(Q_t\).  Thus, conditionally on \((D_t,\widetilde Z)\), the variable
\(\varepsilon_tD_t\) is centered, has conditional second moment \(D_t^2\),
and is bounded in absolute value by \(\beta_n\).  The Bernstein MGF bound
for a centered bounded variable gives
\[
\log \Ex_{Q_t}
\left[\exp\{\lambda\varepsilon_tD_t\}\mid D_t,\widetilde Z\right]
\le
\frac{\lambda^2D_t^2}{2(1-\lambda\beta_n/3)}
\qquad (0<\lambda<3/\beta_n).
\]
Applying the conditional Donsker--Varadhan variational inequality to
\(P_{U_t\mid D_t,\widetilde Z}\) relative to \(Q_t\), and then averaging
over \((D_t,\widetilde Z)\), yields \(\lambda\Ex[\varepsilon_tD_t]
\le
I_t+\frac{\lambda^2\bar\nu_t}{2(1-\lambda\beta_n/3)}.\) The same argument applied to \(-\varepsilon_tD_t\) gives the reverse
inequality, and hence the absolute-value bound in
\eqref{eq:bernstein_local_unoptimized_added}.  Dividing by \(\lambda\)
proves the unoptimized bound.

The standard sub-gamma optimization inequality
\[
\inf_{0<\lambda<3/\beta_n}
\left\{
\frac{a}{\lambda}
+
\frac{\lambda v}{2(1-\lambda\beta_n/3)}
\right\}
\le
\sqrt{2av}
+
\frac{\beta_n}{3}a
\]
with \(a=I_t\) and \(v=\bar\nu_t\) proves
\eqref{eq:bernstein_local_optimized_added}.  Finally, summing over \(t\)
and using the telescoping identity \eqref{eq:wang_telescoping_identity}
proves \eqref{eq:bernstein_global_refinement_added}.
\end{proof}
Here, the Bernstein-type bound is not intended to improve the main bound.
Rather, it shows that any Bernstein-style fast-rate conclusion would need
additional control of the local variance proxy \(\bar\nu_t\).

When the objective is nonnegative, the variance proxy can be related to
mixed-sample objective values.

\begin{proposition}[Positive-objective variance proxy]
\label{prop:positive_objective_variance_proxy_added}
Assume \(\Phi_n(f_W;T)\ge0\) almost surely for every batch \(T\).  Then,
for every \(t\in[n]\),
\begin{align}
D_t^2
&\le
\beta_n |D_t|
\le
\beta_n\Bigl(\Phi_n(f_W;S^{(t,0)})+
\Phi_n(f_W;S^{(t,1)})\Bigr).
\label{eq:positive_objective_variance_proxy_added}
\end{align}
Consequently, with \(M_t\coloneqq \frac12\Ex\!\bigl[
\Phi_n(f_W;S^{(t,0)})+
\Phi_n(f_W;S^{(t,1)})
\bigr],\) one has
\begin{align}
\bar\nu_t
\le
\min\{\beta_n^2,2\beta_n M_t\}.
\label{eq:positive_objective_variance_proxy_expectation_added}
\end{align}
\end{proposition}

\begin{proof}
The stability assumption gives \(|D_t|\le\beta_n\), so
\(D_t^2\le\beta_n|D_t|\).  Since the two mixed-sample values are
nonnegative,
\[
|D_t|
=
\bigl|\Phi_n(f_W;S^{(t,0)})-\Phi_n(f_W;S^{(t,1)})\bigr|
\le
\Phi_n(f_W;S^{(t,0)})
+
\Phi_n(f_W;S^{(t,1)}).
\]
This proves \eqref{eq:positive_objective_variance_proxy_added}.  Taking
expectations gives \(\bar\nu_t\le2\beta_n M_t\), while
\(|D_t|\le\beta_n\) also gives \(\bar\nu_t\le\beta_n^2\).
\end{proof}

\begin{corollary}[Explicit conservative Bernstein constants for CVaR and CR DRO]
\label{cor:bernstein_constants_examples_added}
Assume the base loss satisfies \(0\le \ell(f_w(X),Y)\le1\).
\begin{enumerate}
    \item For CVaR, \(\beta_n=(\bar\alpha n)^{-1}\), and therefore
\begin{align}
    \left|\Ex[\Phi_{\CVaR,\alpha,n}(f_W;S')-\Phi_{\CVaR,\alpha,n}(f_W;S)]\right|
    \le
    \frac{1}{\bar\alpha n}
    \sum_{t=1}^n
    \left(\sqrt{2 I_t}+\frac{1}{3}I_t\right),
    \label{eq:bernstein_cvar_explicit_added}
\end{align}
    using the crude bound \(\bar\nu_t\le\beta_n^2\).

    \item For CR DRO under the stability route,
    \(\beta_n\le\kappa_{c,\epsilon}n^{1/c-1}\), and hence
\begin{align}
    |\Ex\Delta_{\DRO}|
    \le
    \kappa_{c,\epsilon}n^{1/c-1}
    \sum_{t=1}^n
    \left(\sqrt{2 I_t}+\frac{1}{3}I_t\right).
    \label{eq:bernstein_dro_stab_explicit_added}
\end{align}
\end{enumerate}
Here \(I_t=I(D_t;U_t\mid\widetilde{Z})\), with \(D_t\) specialized to the
corresponding objective.
\end{corollary}

\begin{proof}
For both objectives, the stated values of \(\beta_n\) are those proved in
the stability route.  The crude variance bound
\(\bar\nu_t\le\beta_n^2\) follows from \(|D_t|\le\beta_n\).  Substituting
this bound into Proposition~\ref{prop:bernstein_local_refinement_added}
gives the two displays.
\end{proof}

The preceding corollary shows that a Bernstein-type comparison is
available, but it also makes clear why this comparison does not
automatically produce a fast rate.  A useful Bernstein-style refinement
would require an additional self-bounding relation for the local variance
proxy.

\begin{proposition}[Bernstein closure under a mixed-sample self-bounding assumption]
\label{prop:fast_rate_under_self_bounding_added}
Assume \(\Phi_n\ge0\), and suppose there exist constants \(a,b\ge0\) such
that for every \(t\in[n]\),
\begin{align}
M_t
\le
a\,\Ex[\Phi_n(f_W;S)] + b.
\label{eq:mixed_sample_self_bounding_added}
\end{align}
Then
\begin{align}
\left|\Ex\bigl[\Phi_n(f_W;S')-\Phi_n(f_W;S)\bigr]\right|
&\le
\sum_{t=1}^n
\left(
\sqrt{
2\min\{
\beta_n^2,
2\beta_n(a\Ex[\Phi_n(f_W;S)]+b)
\}I_t
}
+
\frac{\beta_n}{3}I_t
\right).
\label{eq:fast_rate_under_self_bounding_general_added}
\end{align}
In particular, if \(b=0\) and
\(a\Ex[\Phi_n(f_W;S)]\ll \beta_n\), then the leading variance term behaves
like \(\sum_{t=1}^n
\sqrt{
4\beta_n a\Ex[\Phi_n(f_W;S)]I_t
},\) up to the residual linear term \(\frac{\beta_n}{3}\sum_{t=1}^n I_t.\) This is only a local variance-sensitive, fast-rate-shaped dependence on
\(\Ex[\Phi_n(f_W;S)]\); it should not be interpreted as an ordinary-loss
fast rate without further aggregation or row-local variance structure.
\end{proposition}

\begin{proof}
Combine \eqref{eq:mixed_sample_self_bounding_added} with
Proposition~\ref{prop:positive_objective_variance_proxy_added}, and then
substitute the resulting variance proxy into
Proposition~\ref{prop:bernstein_local_refinement_added}.
\end{proof}

\subsection{A shifted-sign attempt and why it does not close in general}
\label{app:shifted_sign_attempt_added}

For the pointwise special case \(\Phi_n(f;S)=\widehat L(f;S)\), the increments in Theorem~\ref{thm:wang_general} reduce to \(D_t=n^{-1}\{\ell(f_W,\widetilde Z_{t,0})-\ell(f_W,\widetilde Z_{t,1})\}\), and the single-index bound recovers the standard row-wise eCMI bound, as shown in Appendix~\ref{app:single_input_regime}. In this case, the training and validation risks are averages of loss values from the same loss table. This row-wise structure also permits the shifted-Rademacher fast-rate analysis of Wang and Mao~\citep{wang23}, yielding bounds linear in the information terms when the training risk is zero. This refinement uses the empirical-risk structure in addition to the single-index decomposition; it does not follow from the square-root bound in Theorem~\ref{thm:wang_general} alone. For general set-input optimization-valued objectives, the additional structure need not hold, as explained below.

For ordinary row-separable losses, fast-rate refinements based on
shifted Rademacher variables exploit an additional row-wise representation
of the empirical loss table~\citep{wang23}.  This representation is not available from
the mixed-path increments \(D_t\) alone for a general optimized set-input
objective.  To see the obstruction, fix \(C>0\) and define \(\Delta_C\coloneqq \Phi_n(f_W;S')-(1+C)\Phi_n(f_W;S).\) Using the telescoping identity \eqref{eq:wang_telescoping_identity},
\begin{align}
\Delta_C
&=
\sum_{t=1}^n\varepsilon_tD_t-C\Phi_n(f_W;S).
\label{eq:shifted_raw_current_added}
\end{align}
For any shift parameter \(\eta\in\mathbb R\), this can be rewritten exactly as
\begin{align}
\Delta_C
&=
\sum_{t=1}^n(\varepsilon_t-\eta)D_t
+
\left(\eta\sum_{t=1}^nD_t-C\Phi_n(f_W;S)\right).
\label{eq:shifted_decomp_current_added}
\end{align}
The first term has the desired shifted-sign form.  The difficulty is the
remainder \(R_{\eta,C}
\coloneqq
\eta\sum_{t=1}^nD_t-C\Phi_n(f_W;S).\) For ordinary pointwise empirical risks, shifted-Rademacher proofs use the
fact that, after the algorithmic output is fixed, the empirical training
risk is an average of row-wise losses from the same loss-pair table.  That
extra row-wise structure allows the shifted correlation to be tied directly
to the empirical risk.

For a general optimized set-input objective, however, the batch value
\(\Phi_n(f_W;S)\) is not determined by the sum of the local replacement
increments \(D_t\).  The optimizer defining \(\Phi_n\) may change after a
single replacement, and \(D_t\) records only a mixed-path difference between
two optimized batch values.  Hence there is no fixed choice of \(\eta\)
for which the remainder \(R_{\eta,C}\) vanishes, or is controlled by
\(\Phi_n(f_W;S)\), under the assumptions used in the main theorem.  This
is the obstruction to importing the ordinary shifted-Rademacher fast-rate
argument directly.  The point is not that no shifted argument can ever be
designed for a special objective; rather, the telescoping identity alone
does not provide the row-wise empirical-risk representation required by
the ordinary-loss proof.

\subsection{An alternative route: uniform convergence over \(\Pi\) followed by ordinary-loss fast-rate bounds}
\label{app:uc_then_wang_added}

This subsection is included only as a comparison route and is not used in
the main theorem.  The main analysis controls the optimized
validation--training gap directly and therefore avoids an explicit
complexity term for the optimizer class.  A different, more classical route
is possible for supremum-type objectives whose inner values are averages
of row-wise nonnegative losses.  This route first pays a
uniform-convergence cost over \(\Pi\), then freezes the optimizer selected
on the training sample and applies an ordinary-loss bound to the resulting
augmented algorithmic output.

Assume that
\begin{align}
\Phi_n(f_W;S)
=
\sup_{\pi\in\Pi}\psi_n^{\pi}(f_W;S),
\qquad
\psi_n^{\pi}(f_W;S)
\coloneqq 
\frac{1}{n}\sum_{i=1}^n \ell_{\pi}(f_W,Z_i),
\label{eq:alt_route_rowsup_added}
\end{align}
where each row-wise loss takes values in \([0,1]\).  Assume that a
measurable training-side maximizer exists, and write
\begin{align}
\widehat\pi=\widehat\pi(W,S)
\in
\operatorname*{arg\,max}_{\pi\in\Pi}\psi_n^{\pi}(f_W;S).
\label{eq:alt_route_train_optimizer_added}
\end{align}
For readability, define the validation--training uniform-convergence
modulus \(A_\Pi(f_W;S',S)
\coloneqq
\sup_{\pi\in\Pi}
\bigl|
\psi_n^{\pi}(f_W;S')-\psi_n^{\pi}(f_W;S)
\bigr|.\) Also define
\begin{align}
\mathrm{UC}^{\Pi,\mathrm{vt}}_n(f_W)
\coloneqq 
\Ex\bigl[A_\Pi(f_W;S',S)\bigr].
\label{eq:alt_route_uc_pi_added}
\end{align}

First, the direct uniform-convergence comparison gives the pointwise bound
\begin{align}
\bigl|
\Phi_n(f_W;S')-\Phi_n(f_W;S)
\bigr|
&\le
A_\Pi(f_W;S',S).
\label{eq:alt_route_direct_uc_abs_added}
\end{align}
Consequently,
\begin{align}
\bigl|
\Ex[
\Phi_n(f_W;S')-\Phi_n(f_W;S)
]
\bigr|
&\le
\Ex\bigl|
\Phi_n(f_W;S')-\Phi_n(f_W;S)
\bigr|
\le
\mathrm{UC}^{\Pi,\mathrm{vt}}_n(f_W).
\label{eq:alt_route_direct_uc_expectation_added}
\end{align}
Indeed, for every \(\pi\), \(\psi_n^\pi(f_W;S')
\le
\psi_n^\pi(f_W;S)+A_\Pi(f_W;S',S),\) and taking the supremum over \(\pi\) gives
\(\Phi_n(f_W;S')\le \Phi_n(f_W;S)+A_\Pi(f_W;S',S)\).
The reverse inequality follows by swapping \(S\) and \(S'\).

The following looser decomposition is nevertheless useful for comparison
with ordinary-loss fast-rate bounds, because it isolates the validation--
training gap of the optimizer selected on the training sample.

\begin{proposition}[Freezing the training-side optimizer after paying uniform convergence]
\label{prop:alt_route_uc_split_added}
Under \eqref{eq:alt_route_rowsup_added},
\begin{align}
\Phi_n(f_W;S')-\Phi_n(f_W;S)
&\le
2A_\Pi(f_W;S',S)
\notag\\
&\quad+
\Bigl(
\psi_n^{\widehat\pi}(f_W;S')-
\psi_n^{\widehat\pi}(f_W;S)
\Bigr).
\label{eq:alt_route_gap_split_added}
\end{align}
Consequently,
\begin{align}
\Ex\bigl[\Phi_n(f_W;S')-\Phi_n(f_W;S)\bigr]
\le
2\,\mathrm{UC}^{\Pi,\mathrm{vt}}_n(f_W)
+
\Ex\Bigl[
\psi_n^{\widehat\pi}(f_W;S')-
\psi_n^{\widehat\pi}(f_W;S)
\Bigr].
\label{eq:alt_route_gap_split_expectation_added}
\end{align}
\end{proposition}

\begin{proof}
Since \(\Phi_n(f_W;S)=\psi_n^{\widehat\pi}(f_W;S)\),
\begin{align*}
\Phi_n(f_W;S')-\Phi_n(f_W;S)
&=
\Bigl(
\sup_{\pi\in\Pi}\psi_n^{\pi}(f_W;S')-
\psi_n^{\widehat\pi}(f_W;S')
\Bigr)
\\
&\quad+
\Bigl(
\psi_n^{\widehat\pi}(f_W;S')-
\psi_n^{\widehat\pi}(f_W;S)
\Bigr).
\end{align*}
For the first bracket, for every \(\pi\in\Pi\),
\begin{align*}
\psi_n^\pi(f_W;S')-\psi_n^{\widehat\pi}(f_W;S')
&=
\bigl(\psi_n^\pi(f_W;S')-\psi_n^\pi(f_W;S)\bigr)
\\
&\quad+
\bigl(\psi_n^\pi(f_W;S)-\psi_n^{\widehat\pi}(f_W;S)\bigr)
\\
&\quad+
\bigl(\psi_n^{\widehat\pi}(f_W;S)-\psi_n^{\widehat\pi}(f_W;S')\bigr)
\\
&\le
A_\Pi(f_W;S',S)+0+A_\Pi(f_W;S',S),
\end{align*}
where the middle term is nonpositive by the optimality of \(\widehat\pi\)
on \(S\).  Taking the supremum over \(\pi\) proves
\eqref{eq:alt_route_gap_split_added}, and taking expectations gives
\eqref{eq:alt_route_gap_split_expectation_added}.
\end{proof}

To interpret the remaining term as an ordinary validation--training gap,
augment the algorithmic output by the selected optimizer: \(A^\star\coloneqq (W,\widehat\pi).\) Define the ordinary frozen loss
\begin{align}
\ell^\star(A^\star,z)
\coloneqq
\ell_{\widehat\pi}(f_W,z).
\label{eq:alt_route_frozen_loss_added}
\end{align}
Then
\[
\psi_n^{\widehat\pi}(f_W;S)
=
\frac1n\sum_{i=1}^n
\ell^\star(A^\star,\widetilde Z_{i,U_i}),
\qquad
\psi_n^{\widehat\pi}(f_W;S')
=
\frac1n\sum_{i=1}^n
\ell^\star(A^\star,\widetilde Z_{i,\overline U_i}).
\]
Thus the second term in \eqref{eq:alt_route_gap_split_expectation_added}
is the validation--training gap of the ordinary loss
\(\ell^\star\) for the augmented output \(A^\star\).

\begin{corollary}[Uniform-convergence-plus-ordinary-fast-rate route]
\label{cor:alt_route_wang_added}
Assume \eqref{eq:alt_route_rowsup_added}.  Let
\begin{align}
L_n^{\star}
&\coloneqq
\Ex\Bigl[
\frac1n\sum_{i=1}^n
\ell^\star(A^\star,\widetilde Z_{i,U_i})
\Bigr]
=
\Ex[\Phi_n(f_W;S)],
\label{eq:alt_route_Lnstar_added}
\\
L_{\mu}^{\star}
&\coloneqq
\Ex\Bigl[
\frac1n\sum_{i=1}^n
\ell^\star(A^\star,\widetilde Z_{i,\overline U_i})
\Bigr]
=
\Ex\bigl[\psi_n^{\widehat\pi}(f_W;S')\bigr].
\label{eq:alt_route_Lmustar_added}
\end{align}
Let \(L_t^{\star,+}\) denote the single-index loss-pair projection
associated with the ordinary loss \(\ell^\star\) in the sense of
Wang and Mao~\citep{wang23}, and set \(\mathcal I_t^\star
\coloneqq
I(L_t^{\star,+};U_t\mid\widetilde Z)\) under the conditional supersample convention used in this paper.
Suppose that the ordinary-loss fast-rate theorem applies to the augmented
ordinary-loss problem \((A^\star,\ell^\star)\), giving
\begin{align}
L_\mu^\star-L_n^\star
&\le
\frac{4}{n}\sum_{t=1}^n \mathcal I_t^\star
+
4\sqrt{
L_n^\star\cdot
\frac1n\sum_{t=1}^n \mathcal I_t^\star
}.
\label{eq:alt_route_invoked_wang_fast_rate_added}
\end{align}
Then
\begin{align}
\Ex\bigl[\Phi_n(f_W;S')-\Phi_n(f_W;S)\bigr]
&\le
2\,\mathrm{UC}^{\Pi,\mathrm{vt}}_n(f_W)
+
\frac{4}{n}\sum_{t=1}^n \mathcal I_t^\star
\notag\\
&\quad+
4\sqrt{
\Ex[\Phi_n(f_W;S)]
\cdot
\frac1n\sum_{t=1}^n \mathcal I_t^\star
}.
\label{eq:alt_route_final_bound_added}
\end{align}
\end{corollary}

\begin{proof}
By Proposition~\ref{prop:alt_route_uc_split_added}, it remains to bound
the ordinary validation--training gap of the frozen loss \(\ell^\star\),
namely \(L_\mu^\star-L_n^\star\).  Applying
\eqref{eq:alt_route_invoked_wang_fast_rate_added} and using
\(L_n^\star=\Ex[\Phi_n(f_W;S)]\) gives
\eqref{eq:alt_route_final_bound_added}.
\end{proof}

The bound \eqref{eq:alt_route_final_bound_added} is included only as a
comparison with the ordinary-loss fast-rate route after the training-side
optimizer has been frozen.  It should not be interpreted as an improvement
over the direct uniform-convergence inequality
\eqref{eq:alt_route_direct_uc_expectation_added}.  The main theorem follows
a different route: it controls the optimized validation--training gap
 directly by Theorem~\ref{thm:validation_training_gap}, avoiding an explicit
uniform-convergence complexity term for the optimizer class \(\Pi\).

\subsection{A smaller validation subset}
\label{app:small_validation_subset_added}

In some applications it is useful to compare training performance to a smaller validation subset rather than to the full holdout sample. The following expectation-level result applies to finite supremum-type objectives.

Assume that, for every sample \(T\) of size \(m\),
\begin{align}
\Phi_m(f_W;T)
\coloneqq 
\sup_{\pi\in\Pi}\frac1m\sum_{z\in T}g_\pi(f_W,z),
\qquad
0\le g_\pi(f_W,z)\le B,
\label{eq:small_val_phi_m_added}
\end{align}
where \(1\le |\Pi|<\infty\). Let \(S'=(Z_1',\ldots,Z_n')\), and let \(J\subset[n]\) be sampled uniformly without replacement among all subsets of size \(n'\), where \(1\le n'<n\), independently of all other randomness. Write \(S'_J=(Z_j')_{j\in J}\).

\begin{theorem}[Small validation subset bound]
\label{thm:small_validation_subset_added}
Under \eqref{eq:small_val_phi_m_added},
\begin{align}
\Ex\bigl[\Phi_{n'}(f_W;S'_J)-\Phi_n(f_W;S)\bigr]
&\le
\Ex\bigl[\Phi_n(f_W;S')-\Phi_n(f_W;S)\bigr]
\notag\\
&\quad+2B\sqrt{\frac{(n-n'+1)(\log|\Pi|+1)}{2nn'}}.
\label{eq:small_val_bound_basic_added}
\end{align}
If \(\Ex[\Phi_n(f_W;S')-\Phi_n(f_W;S)]
\le \mathrm{IT}_n(f_W),\) then
\begin{align}
\Ex\bigl[\Phi_{n'}(f_W;S'_J)-\Phi_n(f_W;S)\bigr]
&\le
\mathrm{IT}_n(f_W)
+2B\sqrt{\frac{(n-n'+1)(\log|\Pi|+1)}{2nn'}}.
\label{eq:small_val_bound_it_added}
\end{align}
\end{theorem}

\begin{proof}
For each \(\pi\in\Pi\), define
\[
P_ng_\pi\coloneqq \frac1n\sum_{i=1}^n g_\pi(f_W,Z_i'),
\qquad
P_Jg_\pi\coloneqq \frac1{n'}\sum_{j\in J}g_\pi(f_W,Z_j').
\]
Then
\[
\Phi_n(f_W;S')=\sup_{\pi\in\Pi}P_ng_\pi,
\qquad
\Phi_{n'}(f_W;S'_J)=\sup_{\pi\in\Pi}P_Jg_\pi.
\]
Therefore,
\begin{align}
\Phi_{n'}(f_W;S'_J)-\Phi_n(f_W;S')
&=
\sup_{\pi\in\Pi}P_Jg_\pi-\sup_{\pi\in\Pi}P_ng_\pi
\notag\\
&\le
\sup_{\pi\in\Pi}(P_J-P_n)g_\pi.
\label{eq:small_val_gap_decomp1_added}
\end{align}
Adding and subtracting \(\Phi_n(f_W;S')\), taking expectations, and using \eqref{eq:small_val_gap_decomp1_added} gives
\begin{align}
\Ex\bigl[\Phi_{n'}(f_W;S'_J)-\Phi_n(f_W;S)\bigr]
&\le
\Ex\bigl[\Phi_n(f_W;S')-\Phi_n(f_W;S)\bigr]
+
\Ex\Bigl[\sup_{\pi\in\Pi}(P_J-P_n)g_\pi\Bigr].
\label{eq:small_val_gap_expectation_decomp_added}
\end{align}

It remains to bound the last term. Conditional on \(S'\) and \(f_W\), for each fixed \(\pi\), the values \(g_\pi(f_W,Z_1'),\ldots,g_\pi(f_W,Z_n')\) form a finite population contained in \([0,B]\), and \(P_Jg_\pi\) is the average of a uniform sample of size \(n'\) drawn without replacement from this population. Hence the Hoeffding inequality gives, for every \(\epsilon>0\),
\[
\Pr_J\!\left((P_J-P_n)g_\pi\ge\epsilon\mid S',f_W\right)
\le
\exp\!\left(-\frac{2nn'\epsilon^2}{(n-n'+1)B^2}\right).
\]
By a union bound over \(\Pi\),
\[
\Pr_J\!\left(
\sup_{\pi\in\Pi}(P_J-P_n)g_\pi\ge\epsilon
\mid S',f_W
\right)
\le
|\Pi|
\exp\!\left(-\frac{2nn'\epsilon^2}{(n-n'+1)B^2}\right).
\]
Let \(Y\coloneqq \sup_{\pi\in\Pi}(P_J-P_n)g_\pi,
\qquad
c\coloneqq \frac{2nn'}{(n-n'+1)B^2}.\) Since \(Y\le Y_+\), integrating the tail yields
\[
\Ex[Y\mid S',f_W]
\le
\int_0^\infty
\min\{1,|\Pi|e^{-c\epsilon^2}\}\,d\epsilon
\le
2\sqrt{\frac{\log|\Pi|+1}{c}}.
\]
Thus
\[
\Ex\Bigl[\sup_{\pi\in\Pi}(P_J-P_n)g_\pi\mid S',f_W\Bigr]
\le
2B\sqrt{\frac{(n-n'+1)(\log|\Pi|+1)}{2nn'}}.
\]
Taking expectation over \(S'\) and \(f_W\), and substituting this bound into \eqref{eq:small_val_gap_expectation_decomp_added}, proves \eqref{eq:small_val_bound_basic_added}. The IT bound \eqref{eq:small_val_bound_it_added} follows by replacing the full validation--training term by \(\mathrm{IT}_n(f_W)\).
\end{proof}

The additional term satisfies \(\frac{n-n'+1}{nn'}
=
\frac1{n'}-\frac1n+\frac1{nn'}.\) Since \(n'<n\), this is within a universal constant factor of \(\frac1{n'}-\frac1n.\) Thus the correction term is of order \(B\sqrt{\left(\frac1{n'}-\frac1n\right)(\log|\Pi|+1)}\) up to constants. For moderate \(|\Pi|\), this can be substantially smaller than a crude deterministic subset-size comparison when \(n'\) is not too small.

The balanced supersample removes explicit \(\Pi\)-complexity from the validation--training bound, whereas the current asymmetric-ratio extension reintroduces such complexity. Finiteness of \(\Pi\) in this extension is not fundamental: replacing the union bound by uniform convergence can replace \(\log|\Pi|\), under suitable boundedness and measurability, by covering entropy, Rademacher complexity, or another empirical-process complexity.

\section{Pseudo-dimension control for binned multicalibration}
\label{app:binned-mc-pdim}

We give a scale-sensitive control for binned multicalibration witnesses.
There are two different maps from prediction values to discretized
objects in this appendix, and they play different roles.

First, \(q_B\) is the binning map that is part of the binned MC objective
itself.  It tells us which score bin a prediction belongs to.  Second,
\(Q_\delta\) is an auxiliary quantizer introduced only for the proof.  It
will be used to replace the real-valued prediction inside the residual
\(Y-f(x)\) by a finite-valued approximation.

Throughout the deterministic statements below, \(f\) denotes a fixed
predictor.  After these deterministic bounds are established, we apply
them to the random learned predictor \(f_W\).

Let \(0=\tau_0<\tau_1<\cdots<\tau_B=1\) be fixed bin endpoints, and set \(I_1=[\tau_0,\tau_1],
\qquad
I_b=(\tau_{b-1},\tau_b]\quad (b=2,\ldots,B).\) Define the corresponding bin index map \(q_B:[0,1]\to[B]\) by \(q_B(t)=b
\quad\Longleftrightarrow\quad
t\in I_b.\) Equivalently, \(q_B(t)
=
1+\sum_{k=1}^{B-1}\mathbf 1\{t>\tau_k\}.\)

In the notation of the general MC objective, the binned score witnesses
are \(h_b(v)=\mathbf 1\{q_B(v)=b\},
\qquad b\in[B].\) Hence, for a dataset \(S=\{(X_j,Y_j)\}_{j=1}^n\), the binned
multicalibration objective is
\[
\Phi_{\mathrm{MC},B}(f;S)
=
\sup_{c\in\mathcal C,\ b\in[B],\ \eta\in\{\pm1\}}
\eta\frac1n
\sum_{j=1}^n
c(X_j)
\mathbf 1\{q_B(f(X_j))=b\}
\bigl(Y_j-f(X_j)\bigr).
\]
Assume throughout this appendix that \(Y\in[0,1]\), \(f(x)\in[0,1]\),
and \(0\le c(x)\le1\) for all \(c\in\mathcal C\).

The same prediction value \(f(X_j)\) appears in two different places in
this objective: \(\mathbf 1\{q_B(f(X_j))=b\}
\qquad\text{and}\qquad
Y_j-f(X_j).\) These two occurrences cannot be discretized in the same way.  The first
one is the score-witness part \(h_b(f(X_j))\), namely the bin indicator.
It is discontinuous as a function of \(f(X_j)\), so replacing the
prediction by a nearby rounded value can change the bin whenever the
prediction is close to a bin boundary.  Without a margin assumption, this
can happen for many sample points and does not give a deterministic small
approximation error.

By contrast, the second occurrence is inside the residual \(Y_j-f(X_j)\).
This part is Lipschitz in \(f(X_j)\): if \(f(X_j)\) is replaced by a
\(\delta\)-accurate quantized value, the summand changes by at most
\(\delta\).  Therefore we keep the bin index exactly, using \(q_B\), and
quantize only the residual value, using the auxiliary map \(Q_\delta\).

Let \(\widetilde Z\in\mathcal Z^{n\times2}\) be the supersample, with
entries \(\widetilde Z_{i,j}
=
(\widetilde X_{i,j},\widetilde Y_{i,j}),
\qquad
i\in[n],\ j\in\{0,1\}.\) For a selector vector \(v\in\{0,1\}^n\), write \(\widetilde z_v
=
\{\widetilde Z_{i,v_i}\}_{i=1}^n,
\qquad
\widetilde z_{\overline v}
=
\{\widetilde Z_{i,1-v_i}\}_{i=1}^n.\) In particular, \(S_U=\widetilde Z_U,
\qquad
S_{\overline U}=\widetilde Z_{\overline U}.\)

We now introduce the auxiliary discretization.  Its purpose is not to
approximate the binning map.  Rather, it turns the real-valued residual
prediction \(f(\widetilde X_{i,j})\) into a finite-valued table, so that
the learned supersample table has a pseudo-dimension entropy bound.  The
interval-threshold form below is used only in the pseudo-dimension
counting step: it ensures that \(Q_\delta(f(x))\) is determined by
finitely many threshold comparisons of \(f(x)\).

We use a fixed monotone interval quantizer
\(Q_\delta:[0,1]\to[0,1]\).  More precisely, there are thresholds \(0=\gamma_0<\gamma_1<\cdots<\gamma_{M_\delta}<\gamma_{M_\delta+1}=1\) and grid values \(g_0,\ldots,g_{M_\delta}\in[0,1]\) such that \(M_\delta+1\le \left\lceil\frac1\delta\right\rceil+1\) and \(Q_\delta(t)=g_r
\quad\text{for }t\in J_r,\) where \(J_0=[\gamma_0,\gamma_1],
\qquad
J_r=(\gamma_r,\gamma_{r+1}]
\quad (r=1,\ldots,M_\delta).\) We assume \(|Q_\delta(t)-t|\le \delta
\qquad
\text{for all }t\in[0,1].\)

With \(Q_\delta\) defined, the reason for not using
\(q_B(Q_\delta(f(x)))\) is now explicit: the map \(q_B\) is
discontinuous, while the approximation guarantee for \(Q_\delta\) is only
metric.  A prediction \(f(x)\) and its quantized value \(Q_\delta(f(x))\)
can lie on different sides of a bin boundary.  Thus \(Q_\delta\) is used
only in the residual, not inside the bin indicator.

\begin{remark}[A concrete quantizer]
One convenient choice is the nearest uniform-grid quantizer.  Let \(M=\left\lceil\frac1\delta\right\rceil,
\qquad
g_r=\frac rM,
\qquad
r=0,\ldots,M.\) Define the midpoints \(\gamma_r=\frac{g_{r-1}+g_r}{2},
\qquad
r=1,\ldots,M.\) Then set
\[
Q_\delta(t)=
\begin{cases}
g_0, & t\in[0,\gamma_1],\\
g_r, & t\in(\gamma_r,\gamma_{r+1}],
      \quad r=1,\ldots,M-1,\\
g_M, & t\in(\gamma_M,1].
\end{cases}
\]
This quantizer has \(M+1\le \lceil1/\delta\rceil+1\) output values and
\(M\le\lceil1/\delta\rceil\) internal thresholds.  Moreover, \(|Q_\delta(t)-t|
\le
\frac{1}{2M}
\le
\delta\) for every \(t\in[0,1]\).
\end{remark}

For fixed supersample inputs \(\widetilde X\), define the discretized
supersample prediction table
\[
A_{\delta,B}(f,\widetilde X)
=
\Bigl(
(a^{\mathrm{bin}}_{i,j}(f),a^{\mathrm{val}}_{i,j}(f))
\Bigr)_{i\in[n],\,j\in\{0,1\}},
\]
where
\[
a^{\mathrm{bin}}_{i,j}(f)
=
q_B(f(\widetilde X_{i,j})),
\qquad
a^{\mathrm{val}}_{i,j}(f)
=
Q_\delta(f(\widetilde X_{i,j})).
\]
Thus each supersample input stores two pieces of information: exact bin information and quantized residual-value information. The first component is used in the bin indicator, and the second
component is used inside the residual.  When \(f=f_W\), we write simply \(A_{\delta,B}
=
A_{\delta,B}(f_W,\widetilde X).\)

For a generic table \(a\), write its entries as \(a^{\mathrm{bin}}_{i,j}\in[B],
\qquad
a^{\mathrm{val}}_{i,j}\in[0,1].\) The table \(a\) need not correspond to an actual predictor; the following
display defines a finite-table surrogate objective.  For a selector vector
\(v\in\{0,1\}^n\), the selected batch \(\widetilde z_v\) uses the entries
\((i,v_i)\).  Define
\[
\Phi_{\mathrm{MC},B}^{\delta}(a;\widetilde z_v)
=
\sup_{c\in\mathcal C,\ b\in[B],\ \eta\in\{\pm1\}}
\eta\frac1n
\sum_{i=1}^n
c(\widetilde X_{i,v_i})
\mathbf 1\{a^{\mathrm{bin}}_{i,v_i}=b\}
\bigl(\widetilde Y_{i,v_i}-a^{\mathrm{val}}_{i,v_i}\bigr).
\]
In words, this is the original binned MC objective with the two
occurrences of the prediction treated differently: the score-witness
part uses the exact stored bin \(a^{\mathrm{bin}}\), while the residual
uses the quantized stored value \(a^{\mathrm{val}}\).

\begin{lemma}[Binned MC discretization]
\label{lem:binned_mc_discretization}
For every predictor \(f\), every supersample realization
\(\widetilde z\), and every selector vector \(v\in\{0,1\}^n\),
\[
\left|
\Phi_{\mathrm{MC},B}(f;\widetilde z_v)
-
\Phi_{\mathrm{MC},B}^{\delta}
\bigl(A_{\delta,B}(f,\widetilde X);\widetilde z_v\bigr)
\right|
\le
\delta .
\]
Consequently, for the learned predictor \(f_W\),
\[
\left|
\mathbb E[
\Phi_{\mathrm{MC},B}(f_W;S_{\overline U})
-
\Phi_{\mathrm{MC},B}(f_W;S_U)
]
\right|
\le
2\delta
+
2\sqrt{
\frac{
2I(A_{\delta,B};U\mid\widetilde Z)
}{n}
}.
\]
\end{lemma}

\begin{proof}
Fix \(f\), \(\widetilde z\), and \(v\), and set \(a=A_{\delta,B}(f,\widetilde X).\) For a fixed witness \((c,b,\eta)\), the original witness average and the
surrogate witness average have the same bin indicator, because \(a^{\mathrm{bin}}_{i,v_i}
=
q_B(f(\widetilde X_{i,v_i})).\) They differ only through the replacement of \(f(\widetilde X_{i,v_i})
\quad\text{by}\quad
a^{\mathrm{val}}_{i,v_i}
=
Q_\delta(f(\widetilde X_{i,v_i}))\) inside the residual.  Hence
\[
\begin{aligned}
&\left|
\eta\frac1n\sum_{i=1}^n
c(\widetilde X_{i,v_i})
\mathbf 1\{a^{\mathrm{bin}}_{i,v_i}=b\}
\left\{
f(\widetilde X_{i,v_i})
-
a^{\mathrm{val}}_{i,v_i}
\right\}
\right|
\\
&\qquad\le
\frac1n\sum_{i=1}^n
c(\widetilde X_{i,v_i})
\mathbf 1\{a^{\mathrm{bin}}_{i,v_i}=b\}
\left|
f(\widetilde X_{i,v_i})
-
Q_\delta(f(\widetilde X_{i,v_i}))
\right|
\\
&\qquad\le
\delta .
\end{aligned}
\]
The last inequality uses \(0\le c\le1\), the fact that the bin indicator
is bounded by one, and the quantization guarantee.  Taking the supremum
over the same witness class on both sides and using the elementary
supremum-difference inequality gives
\[
\left|
\Phi_{\mathrm{MC},B}(f;\widetilde z_v)
-
\Phi_{\mathrm{MC},B}^{\delta}
\bigl(A_{\delta,B}(f,\widetilde X);\widetilde z_v\bigr)
\right|
\le
\delta .
\]

Applying this approximation once to \(S_U\) and once to
\(S_{\overline U}\), with \(f=f_W\), gives \(\left|
G_{\mathrm{MC},B}(f_W)
-
G_{\mathrm{MC},B}^{\delta}(A_{\delta,B})
\right|
\le
2\delta,\) where \(G_{\mathrm{MC},B}(f_W)
=
\Phi_{\mathrm{MC},B}(f_W;S_{\overline U})
-
\Phi_{\mathrm{MC},B}(f_W;S_U)\) and \(G_{\mathrm{MC},B}^{\delta}(A_{\delta,B})
=
\Phi_{\mathrm{MC},B}^{\delta}(A_{\delta,B};S_{\overline U})
-
\Phi_{\mathrm{MC},B}^{\delta}(A_{\delta,B};S_U).\) Therefore, \(\left|
\mathbb E G_{\mathrm{MC},B}(f_W)
\right|
\le
2\delta
+
\left|
\mathbb E G_{\mathrm{MC},B}^{\delta}(A_{\delta,B})
\right|.\)

It remains to control the expectation of the discretized gap.  Fix
\(\widetilde Z=\widetilde z\) and a table \(A_{\delta,B}=a\).  Define
\[
G_a(u)
=
\Phi_{\mathrm{MC},B}^{\delta}(a;\widetilde z_{\overline u})
-
\Phi_{\mathrm{MC},B}^{\delta}(a;\widetilde z_u),
\qquad
u\in\{0,1\}^n.
\]
Under the reference law \(Q_U=\mathrm{Unif}(\{0,1\}^n)\), the complement
map \(u\mapsto\overline u\) is measure preserving and \(G_a(\overline u)=-G_a(u).\) Thus \(\mathbb E_{Q_U}G_a(U)=0.\)

We next show that \(G_a(U)\) is sub-Gaussian under \(Q_U\).  If two
selector vectors \(u\) and \(u^{\oplus i}\) differ only in coordinate \(i\),
then the selected batch changes in one coordinate.  For the surrogate
objective, every witness summand is bounded in absolute value by one: \(\left|
c(\widetilde X_{i,j})
\mathbf 1\{a^{\mathrm{bin}}_{i,j}=b\}
(\widetilde Y_{i,j}-a^{\mathrm{val}}_{i,j})
\right|
\le 1.\) Therefore one coordinate replacement changes
\(\Phi_{\mathrm{MC},B}^{\delta}(a;\cdot)\) by at most \(2/n\).  The
complementary batch also changes in one coordinate, so \(|G_a(u)-G_a(u^{\oplus i})|
\le
\frac4n .\) By the bounded-differences Hoeffding lemma,
\[
\log\mathbb E_{Q_U}
\exp\{\lambda G_a(U)\}
\le
\frac{\lambda^2}{8}
\sum_{i=1}^n
\left(\frac4n\right)^2
=
\frac{2\lambda^2}{n}
=
\frac{\lambda^2(2/\sqrt n)^2}{2}
\]
for all \(\lambda\in\mathbb R\).

Applying the conditional change-of-measure inequality to the actual
conditional law \(P_{U\mid A_{\delta,B},\widetilde Z}\) relative to
\(Q_U\), and then averaging over \((A_{\delta,B},\widetilde Z)\), gives
\[
\left|
\mathbb E
G_{\mathrm{MC},B}^{\delta}(A_{\delta,B})
\right|
\le
\frac{2}{\sqrt n}
\sqrt{
2\,
\mathbb E
D\!\left(
P_{U\mid A_{\delta,B},\widetilde Z}
\,\middle\|\, Q_U
\right)
}.
\]
Since \(U\) is independent of \(\widetilde Z\), the reference law
\(Q_U\) is \(P_{U\mid\widetilde Z}\).  Therefore \(\mathbb E
D\!\left(
P_{U\mid A_{\delta,B},\widetilde Z}
\,\middle\|\, Q_U
\right)
=
I(A_{\delta,B};U\mid\widetilde Z).\) Consequently,
\[
\left|
\mathbb E
G_{\mathrm{MC},B}^{\delta}(A_{\delta,B})
\right|
\le
2\sqrt{
\frac{
2I(A_{\delta,B};U\mid\widetilde Z)
}{n}
}.
\]
Combining this with the \(2\delta\) discretization error proves the
claim.
\end{proof}

\begin{lemma}[Multi-threshold growth bound for pseudo-dimension]
\label{lem:pdim_multithreshold_table}
Let \(\mathcal F\subseteq[0,1]^{\mathcal X}\) have pseudo-dimension
\(d_0\).  Fix input points \(x_1,\ldots,x_m\) and thresholds
\(a_1,\ldots,a_T\).  The number of joint threshold tables \(\left(
\mathbf 1\{f(x_i)>a_k\}
\right)_{i\in[m],\,k\in[T]},
\qquad
f\in\mathcal F,\) is at most \(\left(\frac{emT}{d_0}\right)^{d_0}\) when \(1\le d_0\le mT\), with the trivial bound \(2^{mT}\) covering the
remaining cases.

Consequently, suppose \(f_W\in\mathcal F\) almost surely, and let \(T_{\delta,B}
=
\left|
\{\tau_1,\ldots,\tau_{B-1}\}
\cup
\{\gamma_1,\ldots,\gamma_{M_\delta}\}
\right|\) be the number of distinct thresholds needed to determine both
\(q_B(f(x))\) and \(Q_\delta(f(x))\).  Then, with \(d_\star=d_0\wedge (2nT_{\delta,B}),\) we have \(H\!\left(A_{\delta,B}(f_W,\widetilde X)\mid\widetilde Z\right)
\le
d_\star
\log\left(
\frac{2enT_{\delta,B}}{d_\star}
\right)\) whenever \(d_\star\ge1\).  If \(d_\star=0\), the entropy is zero.
In particular, in the nontrivial regime \(1\le d_0\le2nT_{\delta,B}\), \(H\!\left(A_{\delta,B}(f_W,\widetilde X)\mid\widetilde Z\right)
\le
d_0
\log\left(
\frac{2enT_{\delta,B}}{d_0}
\right).\) \end{lemma}

\begin{proof}
The pseudo-dimension of \(\mathcal F\) is the VC dimension of the
subgraph threshold class \(\mathcal S_{\mathcal F}
=
\left\{
(x,a)\mapsto\mathbf 1\{f(x)>a\}: f\in\mathcal F
\right\}.\) The joint threshold table on \((x_i,a_k)\) is exactly a binary labeling
of the \(mT\) augmented points \((x_i,a_k)\in\mathcal X\times[0,1],
\qquad
i\in[m],\ k\in[T],\) by the class \(\mathcal S_{\mathcal F}\).  Sauer--Shelah therefore gives
at most \(\left(\frac{emT}{d_0}\right)^{d_0}\) possible tables when \(1\le d_0\le mT\).  If \(d_0>mT\), the trivial
bound \(2^{mT}\) applies; this is captured by replacing \(d_0\) by
\(d_0\wedge mT\).

Now fix a supersample realization \(\widetilde Z=\widetilde z\).  The
table \(A_{\delta,B}(f,\widetilde X)
=
\Bigl(
q_B(f(\widetilde X_{i,j})),
Q_\delta(f(\widetilde X_{i,j}))
\Bigr)_{i,j}\) is determined by threshold comparisons of \(f\) at the \(2n\)
supersample inputs \(\widetilde X_{i,j},
\qquad
i\in[n],\ j\in\{0,1\},\) against the union of two finite threshold sets: \(\{\tau_1,\ldots,\tau_{B-1}\}
\quad\text{and}\quad
\{\gamma_1,\ldots,\gamma_{M_\delta}\}.\) The first set determines the exact bin index \(q_B(f(x))\).  The second
set determines the interval cell of \(f(x)\) under \(Q_\delta\), and
therefore determines the quantized value \(Q_\delta(f(x))\).  Hence the
number of distinct thresholds needed is \(T_{\delta,B}
=
\left|
\{\tau_1,\ldots,\tau_{B-1}\}
\cup
\{\gamma_1,\ldots,\gamma_{M_\delta}\}
\right|
\le
(B-1)+M_\delta.\)

Since \(A_{\delta,B}(f,\widetilde X)\) is a function of these threshold
comparison tables, the number of possible values of
\(A_{\delta,B}(f,\widetilde X)\), as \(f\) ranges over \(\mathcal F\), is
at most the Sauer--Shelah growth bound with \(m=2n,
\qquad
T=T_{\delta,B}.\) Therefore, conditionally on \(\widetilde Z\), \(H\!\left(A_{\delta,B}(f_W,\widetilde X)\mid
\widetilde Z=\widetilde z\right)
\le
d_\star
\log\left(
\frac{2enT_{\delta,B}}{d_\star}
\right),\) with the zero-entropy convention when \(d_\star=0\).  Averaging over
\(\widetilde Z\) proves the entropy bound.  The displayed nontrivial
regime follows by taking \(d_\star=d_0\).
\end{proof}

\begin{proof}[Proof of Corollary~\ref{cor:binned_mc_pdim_main}]
Choose \(\delta=\frac1n.\) Use any interval quantizer satisfying the conditions above; for example,
the uniform-grid quantizer in the preceding remark.  By construction,
the quantizer has \(M_\delta+1\) output values and \(M_\delta+1
\le
\left\lceil\frac1\delta\right\rceil+1.\) Therefore \(M_\delta
\le
\left\lceil\frac1\delta\right\rceil.\) With \(\delta=1/n\), this becomes \(M_\delta
\le
\left\lceil n\right\rceil
=
n.\)

The table \(A_{\delta,B}\) is determined by two kinds of thresholds.
The exact bin \(q_B(f(x))\) is determined by the \(B-1\) bin boundaries \(\tau_1,\ldots,\tau_{B-1},\) while the quantized residual value \(Q_\delta(f(x))\) is determined by
the \(M_\delta\) quantizer thresholds \(\gamma_1,\ldots,\gamma_{M_\delta}.\) Hence the total number of distinct thresholds is
\[
T_{\delta,B} = \left| \{\tau_1,\ldots,\tau_{B-1}\} \cup \{\gamma_1,\ldots,\gamma_{M_\delta}\} \right| \le (B-1)+M_\delta \le (B-1)+n \le B+n.
\]
If some bin thresholds and quantizer thresholds coincide, the union is
even smaller.  The important point is that the threshold count is
additive, not multiplicative: one needs the bin-boundary comparisons and
the quantizer-cell comparisons, not all pairs of bins and quantizer
values.

Since mutual information is bounded by conditional entropy,
Lemma~\ref{lem:pdim_multithreshold_table} gives \(I(A_{\delta,B};U\mid\widetilde Z)
\le
H(A_{\delta,B}\mid\widetilde Z).\) Using the threshold count \(T_{\delta,B}\le B+n\), we obtain, in the
usual nontrivial regime, \(I(A_{\delta,B};U\mid\widetilde Z)
\le
H(A_{\delta,B}\mid\widetilde Z)
\lesssim
d_0
\log\left(
\frac{en(B+n)}{d_0}
\right).\) More precisely, one may replace \(d_0\) by \(d_\star=d_0\wedge O(n(B+n))\) inside the entropy bound.  The displayed expression is the form used in
the main text.

Substituting this entropy bound and \(\delta=1/n\) into
Lemma~\ref{lem:binned_mc_discretization} yields
\[
\left|
\mathbb E\{
\Phi_{\mathrm{MC},B}(f_W;S_{\overline U})
-
\Phi_{\mathrm{MC},B}(f_W;S_U)
\}
\right|
\lesssim
\frac1n
+
\sqrt{
\frac{
d_0\log\{en(B+n)/d_0\}
}{n}
}.
\]
Since \(S_{\overline U}=S'\), \(S_U=S\), and
\(\Phi_{\mathrm{MC},B,n}\equiv\Phi_{\mathrm{MC},B}\) on empirical
batches, this is exactly Eq.~\eqref{eq:binned_mc_pdim_main}.
\end{proof}

\paragraph{Why the threshold count is additive}
The bound \(T_{\delta,B}
=
\left|
\{\tau_1,\ldots,\tau_{B-1}\}
\cup
\{\gamma_1,\ldots,\gamma_{M_\delta}\}
\right|
\le
(B-1)+M_\delta\) is an additive threshold count, not a multiplicative one.  The reason is
that both the bin index and the quantized residual value are determined
from the same scalar prediction value \(t=f(x)\in[0,1].\) The bin index \(q_B(t)\) is determined by comparing \(t\) with the
\(B-1\) bin boundaries \(\tau_1,\ldots,\tau_{B-1},\) while the quantized value \(Q_\delta(t)\) is determined by comparing
\(t\) with the \(M_\delta\) quantizer thresholds \(\gamma_1,\ldots,\gamma_{M_\delta}.\) Thus the pair \((q_B(t),Q_\delta(t))\) is determined by the union of these two one-dimensional threshold sets.

Equivalently, the bin partition and the quantizer partition are two
partitions of the same interval \([0,1]\).  Their common refinement is
obtained by placing all boundary points on the same line.  If the total
number of distinct boundary points is \(T_{\delta,B}\), then this common
refinement has at most \(T_{\delta,B}+1\) cells.  Hence the number of
relevant one-dimensional comparisons is controlled by \(T_{\delta,B}
\le
(B-1)+M_\delta,\) rather than by the product \(B(M_\delta+1)\).

\section{ Experimental settings and additional results}
\label{app:experimental-settings}

 This appendix gives the experimental settings and additional results for logistic regression on Flights200K.

\subsection{Flights200K: settings and estimation}
\label{app:flights-settings}

\paragraph{Settings.}
We use $n=2^9,2^{10},\ldots,2^{15}$ and $\sigma\in\{0.05,0.10\}$.
The linear logistic regression model is fitted with the \texttt{newton-cholesky} solver and regularization parameter $C=1$ in \texttt{scikit-learn}.
For each sample configuration, we newly split the data into a 20\% held-out sample, a 10\% reference sample $S_{\mathrm{ref}}$, and a 70\% candidate pool.
The nine groups are defined using the medians of the first three features in the reference sample; the five score bins have equal width.
Within each configuration, the same group definitions are used for training, validation, and both selector branches.

We estimate the generalization gap by Monte Carlo as follows. Specifically, we average the validation--training MC differences over sample configurations and Gaussian perturbations, then take the absolute value.
For one estimate, we generate 32 sample configurations, indexed by $j=1,\ldots,J$ with $J=32$.
Each configuration consists of a reference sample, a supersample of $n$ pairs drawn without replacement from its candidate pool, and a selector vector. These determine the training sample $S_j$ and the complementary validation sample $S'_j$.
For each $j$, we fit the model on $S_j$ and generate 16 independent Gaussian perturbations, indexed by $r=1,\ldots,R$ with $R=16$.
The perturbed parameters are $W_{jr}=\widehat\theta(S_j)+\sigma\xi_{jr}$, where $\xi_{jr}\sim N(0,I_8)$ are independent, as in Section~\ref{sec:flights-experiment}.
We evaluate each perturbed predictor on the same $S_j$ and $S'_j$ and average the resulting $JR$ signed differences, giving
\begin{equation}
\widehat L_n
=
\left|
\frac{1}{JR}\sum_{j=1}^{J}\sum_{r=1}^{R}
\bigl\{\Phi_{\mathrm{MC}}(f_{W_{jr}};S'_j)
-\Phi_{\mathrm{MC}}(f_{W_{jr}};S_j)\bigr\}
\right|.
\tag{H.1}\label{eq:flights-gap-estimate}
\end{equation}
The average inside the absolute value approximates the expectation over the sample configuration and Gaussian perturbation in Eq.~\eqref{eq:mc_wang_bound_main}.

\paragraph{Conditional mutual information.}
Let $D_t^{\mathrm{MC}}$ denote the increment $D_t$ in Theorem~\ref{thm:wang_general} specialized to the MC objective.
To upper-bound the local information in Eq.~\eqref{eq:mc_wang_bound_main}, we evaluate the conditional mutual information between the learned parameter $W$ and the selector $U_t$.
Write $C_t=(\widetilde Z,U_{-t},S_{\mathrm{ref}})$ and fix $C_t=c$.
Let $\mu_0(c)$ and $\mu_1(c)$ be the fitted parameters obtained by retraining with $U_t=0$ and $U_t=1$, respectively.
The Gaussian perturbation model gives \(W\mid(U_t=u,C_t=c)\sim N(\mu_u(c),\sigma^2 I_8).\) Let $p_0,p_1$ be these Gaussian densities and put
$d_t(c)=\|\mu_1(c)-\mu_0(c)\|_2/\sigma$.
The selector vector is generated independently of $(\widetilde Z,S_{\mathrm{ref}})$, and its coordinates are independent and uniform on $\{0,1\}$.
Thus $U_t$ remains uniform given $C_t=c$, and the conditional density of $W$ is $(p_0+p_1)/2$.
The definition of mutual information gives
\[
I(W;U_t\mid C_t=c)
=\log 2-\frac12\sum_{u=0}^{1}
\mathbb E_{W\sim p_u}\log\left(1+\frac{p_{1-u}(W)}{p_u(W)}\right).
\]
Conditioning on $U_t=u$ and $C_t=c$, both means are fixed vectors and
$W=\mu_u(c)+\sigma\xi$ with $\xi\sim N(0,I_8)$.
Substituting the Gaussian densities yields
\[
\log\frac{p_u(W)}{p_{1-u}(W)}
=\frac{d_t(c)^2}{2}
+\frac{\{\mu_u(c)-\mu_{1-u}(c)\}^{\mathsf T}\xi}{\sigma}.
\]
The second term is Gaussian with mean zero and variance $d_t(c)^2$.
For a standard normal random variable $Z$, the expectation in the mutual information is therefore
\[
\mathbb E_{W\sim p_u}\log\left(1+\frac{p_{1-u}(W)}{p_u(W)}\right)
=\mathbb E_{Z\sim N(0,1)}\log\left(1+e^{-d_t(c)^2/2-d_t(c)Z}\right).
\]
The right-hand side does not depend on $u$, so averaging over the two branches gives
\begin{equation}
\begin{aligned}
I(W;U_t\mid C_t=c)&=G(d_t(c)),\\
G(d)&=\log2-\mathbb E_{Z\sim N(0,1)}\log\left(1+e^{-d^2/2-dZ}\right).
\end{aligned}
\tag{H.2}\label{eq:flights-gaussian-mi}
\end{equation}
When the means coincide, the log-density ratio is zero and the same expression gives $G(0)=0$.
We use natural logarithms and evaluate this one-dimensional integral by 80-point Gauss--Hermite quadrature.

Independence of the selectors allows $U_{-t}$ and $S_{\mathrm{ref}}$ to be added to the conditioning, as in Eq.~\eqref{eq:uncond_le_cond_local_info}.
For fixed $C_t$, the two mixed-path samples in Theorem~\ref{thm:wang_general} and the group definitions are fixed, so $D_t^{\mathrm{MC}}$ is a function of the predictions $F=f_W(\widetilde X)$.
Eq.~\eqref{eq:cond_local_dpi_full_table} and data processing give
\begin{equation}
\begin{aligned}
I_{\mathrm{loc},t}^{\mathrm{MC}}
&\le I(D_t^{\mathrm{MC}};U_t\mid C_t)
\le I(F;U_t\mid C_t)\\
&\le I(W;U_t\mid C_t)
=\mathbb E_{C_t}[G(d_t(C_t))].
\end{aligned}
\tag{H.3}\label{eq:flights-information-bound}
\end{equation}
For $\Delta_{\mathrm{MC}}=\Phi_{\mathrm{MC}}(f_W;S')-\Phi_{\mathrm{MC}}(f_W;S)$, the single-index term in Eq.~\eqref{eq:mc_wang_bound_main} therefore gives
\begin{equation}
|\mathbb E\Delta_{\mathrm{MC}}|
\le B_{G,n}
=\frac{2}{n}\sum_{t=1}^{n}\sqrt{2\,\mathbb E_{C_t}[G(d_t(C_t))]}.
\tag{H.4}\label{eq:flights-gaussian-bound}
\end{equation}

\paragraph{Estimating the bound.}
Here $t$ indexes one of the $n$ pairs in the supersample.
For each estimation run, we uniformly sample 16 indices without replacement from $\{1,\ldots,n\}$, independently of the data, and denote this set by $T_m$, with $m=16$.
For these indices, we evaluate the mutual information using the same 32 configurations as above.
Let $\widetilde Z_j$, $U^{(j)}$, and $S_{\mathrm{ref},j}$ denote the supersample, selector vector, and reference sample in configuration $j$.
For each $t\in T_m$, fix \(C_{t,j}=(\widetilde Z_j,U_{-t}^{(j)},S_{\mathrm{ref},j}),\) change only the $t$-th selector to zero or one, and fit the two models.
The distance between $\mu_0(C_{t,j})$ and $\mu_1(C_{t,j})$ determines $G(d_t(C_{t,j}))$ through Eq.~\eqref{eq:flights-gaussian-mi}.
Although the index $t$ is held fixed across configurations, the data pair occupying that position is drawn anew in each configuration.

For each $t$, we average $G(d_t(C_{t,j}))$ over the 32 configurations to estimate the expectation over $C_t$ in Eq.~\eqref{eq:flights-information-bound}.
We substitute this average into Eq.~\eqref{eq:flights-gaussian-bound} before taking the square root, and approximate the average over all $n$ indices by the average over the 16 sampled indices.
One estimation run thus gives
\begin{equation}
\widehat B_{G,n}
=\frac{2}{m}\sum_{t\in T_m}
\sqrt{\frac{2}{J}\sum_{j=1}^{J}G\bigl(d_t(C_{t,j})\bigr)},
\qquad J=32.
\tag{H.5}\label{eq:flights-bound-estimate}
\end{equation}

\paragraph{Means and standard deviations of the estimates.}
For each sample size and noise level, we independently repeat the entire procedure 20 times, obtaining 20 gap estimates from Eq.~\eqref{eq:flights-gap-estimate} and 20 bound estimates from Eq.~\eqref{eq:flights-bound-estimate}.
In each run, we generate new sample configurations and selector vectors, 16 perturbations for each fitted model, and 16 pair indices.
Each point in Figure~\ref{fig:flights-bound} is the mean of the corresponding 20 estimates.
Error bars show this mean plus or minus one sample standard deviation, computed with 19 degrees of freedom.
They describe the variation of the finite Monte Carlo estimate when the 32-configuration estimation procedure is repeated.
For the bound, Eq.~\eqref{eq:flights-gaussian-mi} integrates over the Gaussian noise.
Its estimate still varies because regenerating the samples, selector vectors, reference samples, and pair indices changes the retrained parameter centers.

\subsection{CVaR}
\label{app:flights-cvar}
Using the training, sample-generation, Gaussian-perturbation, and repetition settings above, we next examine the generalization gap and upper bound for CVaR, which evaluates the tail of the loss distribution.
The training objective is unchanged, and evaluation uses the Brier loss $\ell(f_W(x),y)=(f_W(x)-y)^2\in[0,1]$.
We use $\alpha\in\{0.2,0.5,0.8,0.95\}$, corresponding to the average over the largest 80\%, 50\%, 20\%, and 5\% of the loss distribution, respectively.
Within each repetition, all levels use the same training and validation samples and perturbed models.
The gap is estimated by replacing MC with CVaR in Eq.~\eqref{eq:flights-gap-estimate}.

\paragraph{Computing CVaR.}
With the perturbed predictor $f_W$ fixed, we compute the $n$ Brier loss values separately on the training and validation samples.
For an evaluation sample $s=((x_i,y_i))_{i=1}^n$, let $r_i=\ell(f_W(x_i),y_i)$ and sort these values in descending order as $r_{(1)}\ge\cdots\ge r_{(n)}$.
With $\bar\alpha=1-\alpha$ as in the main text, put $a=n\bar\alpha$, $k=\lfloor a\rfloor$, and $\delta=a-k$.
Eq.~\eqref{eq:cvar_empirical} can be rewritten as~\citep{rockafellar2002general}
\[
\Phi_{\mathrm{CVaR},\alpha}(f_W;s)
=\frac{\displaystyle\sum_{i=1}^{k}r_{(i)}+\delta r_{(k+1)}}{a}.
\]
The second term is zero when $\delta=0$, and the sum is zero when $k=0$.

For the upper bound, we apply Corollary~\ref{cor:cvar_l2} for loss values in $[0,1]$ and take the smaller of the stability and convex $L_2$-Lipschitz bounds for the loss vector.
To upper-estimate $I_{\mathrm{fCMI}}$, we use the conditional mutual-information estimates obtained above.
These information estimates are shared across all values of $\alpha$; the level dependence of the bound comes from the coefficient in Corollary~\ref{cor:cvar_l2}.

\begin{figure}[H]
    \centering
    \includegraphics[width=\linewidth]{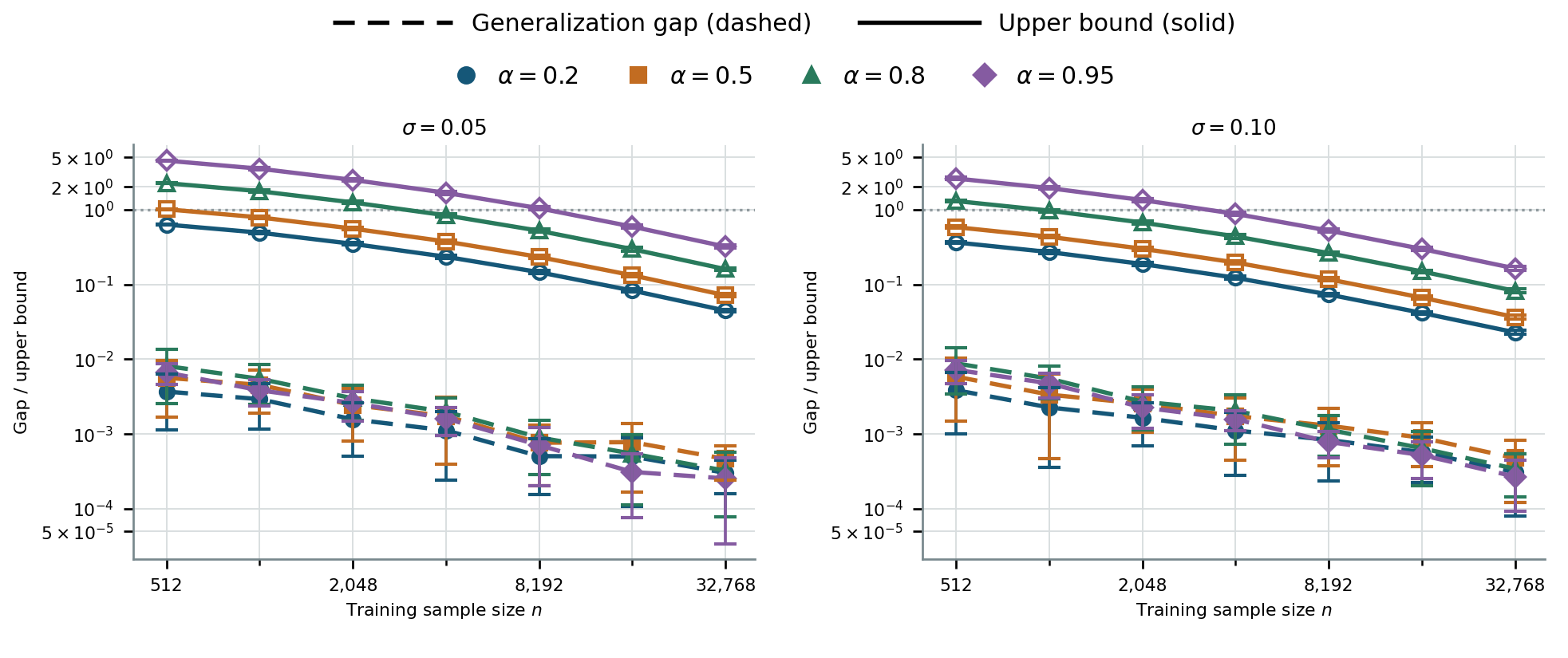}
    \caption{CVaR generalization-gap and upper-bound estimates from Corollary~\ref{cor:cvar_l2}, with $\sigma=0.05$ (left) and $\sigma=0.10$ (right). Dashed lines with filled markers show gaps; solid lines with open markers show bounds. Colors identify $\alpha$. Points and error bars show the mean and one sample standard deviation over 20 independent repetitions. The horizontal dotted line is one.}
    \label{fig:flights-cvar-alpha}
\end{figure}

As for MC, Figure~\ref{fig:flights-cvar-alpha} shows decreasing gap and bound estimates as the sample size increases, at both noise levels.
Increasing $\alpha$ reduces the tail fraction $\bar\alpha$ and concentrates evaluation on fewer, larger loss values.
The upper bound increases with $\alpha$, and more samples are needed before it falls below one.
However, the estimated generalization gap does not increase consistently with $\alpha$.
As in the MC experiment, the bound estimates exceed the gap estimates and remain conservative.

For CVaR, as for MC, the bounds at $\sigma=0.10$ are smaller than those at $\sigma=0.05$.
This is consistent with the Gaussian information above being determined by the distance between the two fitted parameter vectors relative to the noise level.
In contrast, the CVaR gap is larger at $\sigma=0.10$ in many settings.
Noise changes both the information and the prediction loss, so its effects on the bound and the generalization gap can be distinguished.

\subsection{CR DRO}
\label{app:flights-crdro}
Using the same experimental settings, we evaluate CR DRO on the perturbed models and Brier loss used in Appendix~\ref{app:flights-cvar}.
For the shape comparison, we fix $\epsilon=0.1$ and use $c\in\{1.25,1.5,2,4\}$; for the radius comparison, we fix $c=2$ and use $\epsilon\in\{0,0.01,0.1,1\}$.
Within each repetition, samples and models are shared, and we vary the evaluation-time reweighting constraint to examine the sensitivity of the gap and bound.

We use the same generator $f_c(t)=(t^c-ct+c-1)/\{c(c-1)\}$ as in the main text.
Let $\widehat P_s$ be the empirical distribution of the evaluation sample $s$ and $Q$ the reweighted distribution.
We maximize the mean loss subject to $D_c(Q\|\widehat P_s)=\mathbb E_{\widehat P_s}f_c(dQ/d\widehat P_s)\le\epsilon$~\citep{Duchi2021}.
The limit as $c\downarrow1$ is the KL divergence $D_{\mathrm{KL}}(Q\|\widehat P_s)$, whereas $c=2$ corresponds to one half of the Pearson $\chi^2$ divergence.
Under this convention, radius $\epsilon$ corresponds to radius $2\epsilon$ for the Pearson $\chi^2$ divergence without the factor $1/2$.
The values $c=1.25,1.5$ belong to the Cressie--Read family with $1<c<2$, while $c=4$ gives a fourth-order density-ratio term.
Equal numerical radii do not define the same feasible reweightings when $c$ changes.

Since the loss values lie in $[0,1]$ and empirical CR DRO with $c>1$ is convex and $L_2$-Lipschitz in the loss vector, we use Eq.~\eqref{eq:dro_vt_bound_main} in Corollary~\ref{cor:dro_l2} for positive radii.
At $\epsilon=0$, the objective is the empirical mean, so we substitute the one-point sensitivity $\beta_n=1/n$ into the term of Theorem~\ref{thm:wang_general} that uses information about the full selector vector.
We estimate the gap by replacing MC with CR DRO in Eq.~\eqref{eq:flights-gap-estimate}, and use the same information estimates as in Appendix~\ref{app:flights-cvar} for all $(c,\epsilon)$.

\paragraph{Computing CR DRO.}
We evaluate CR DRO using the method of~\citet{Duchi2021}, as follows.
As for CVaR, write the loss values on the evaluation sample $s$ as $r_1,\ldots,r_n$, and let $c>1$.
For a fixed threshold $\eta<\max_i r_i$, the density-ratio weights have the analytic form~\citep{Duchi2021}
\[
q_i(\eta)
=\frac{n[r_i-\eta]_+^{1/(c-1)}}
{\displaystyle\sum_{j=1}^{n}[r_j-\eta]_+^{1/(c-1)}}.
\]
These weights satisfy $n^{-1}\sum_iq_i(\eta)=1$; they assign zero weight to loss values at or below the threshold and larger weights to larger loss values.

Let $Q_\eta$ assign probability $q_i(\eta)/n$ to each sample point; its density ratio with respect to $\widehat P_s$ is $q_i(\eta)$.
For a positive radius, when the worst-case value is below the maximum loss, the threshold $\eta^\star$ satisfies
\[
D_c(Q_{\eta^\star}\|\widehat P_s)
=\frac1n\sum_{i=1}^{n}f_c\!\left(q_i(\eta^\star)\right)
=\frac{n^{-1}\displaystyle\sum_{i=1}^{n}q_i(\eta^\star)^c-1}{c(c-1)}
=\epsilon.
\]
The divergence $D_c(Q_\eta\|\widehat P_s)$ is nondecreasing in $\eta$.
Thus, if $D_c(Q_\eta\|\widehat P_s)<\epsilon$, we narrow the search to larger thresholds; if $D_c(Q_\eta\|\widehat P_s)>\epsilon$, we narrow it to smaller thresholds.
We find $\eta^\star$ by bisecting the threshold interval at each candidate value $\eta$ and retaining the half containing a solution to the equality above.
Finally, we compute \(\Phi_{\mathrm{DRO},c,\epsilon}(f_W;s)
=\frac1n\sum_{i=1}^{n}q_i(\eta^\star)r_i\) separately on the training and validation samples.
At $\epsilon=0$, we use the mean loss; if reweighting only the observations attaining the maximum loss satisfies the divergence constraint, we use that maximum.

\begin{figure}[H]
    \centering
    \includegraphics[width=\linewidth]{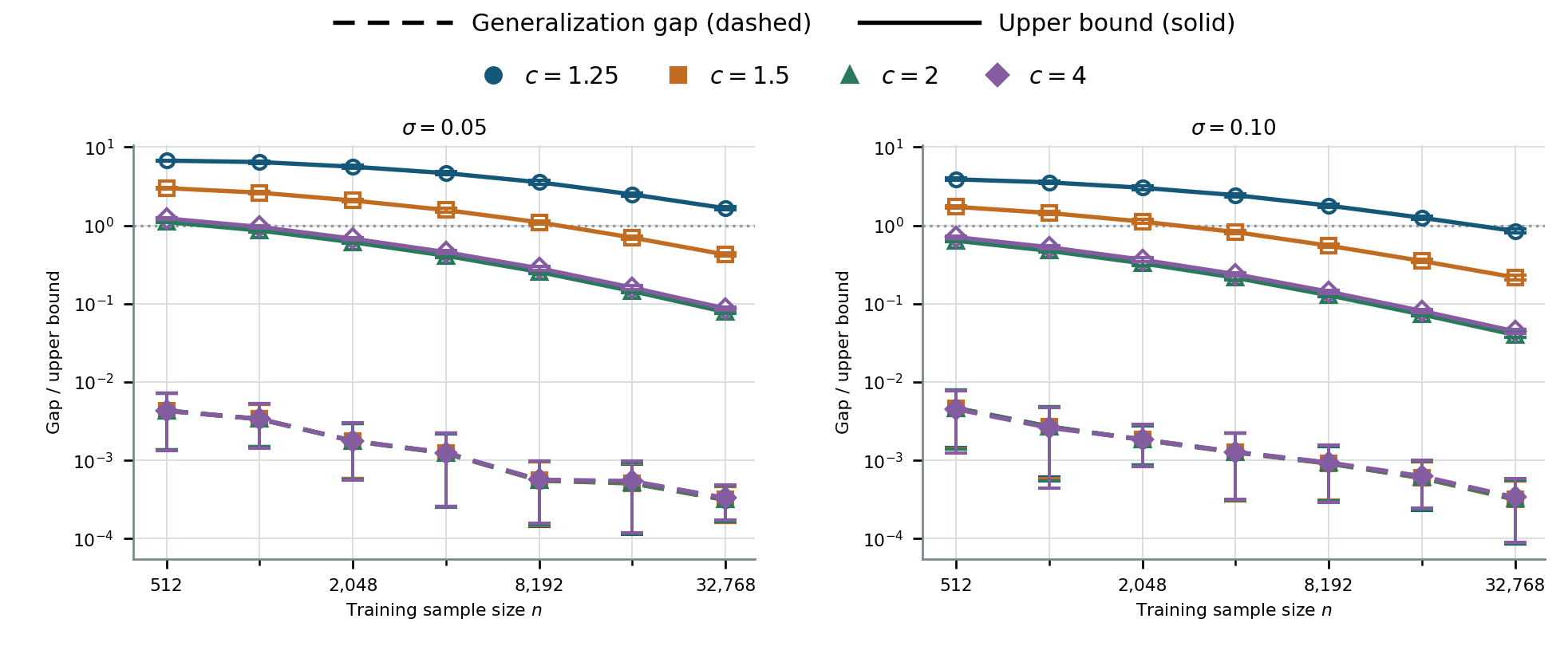}
    \caption{CR DRO generalization-gap and upper-bound estimates from Eq.~\eqref{eq:dro_vt_bound_main} in Corollary~\ref{cor:dro_l2}, at radius $\epsilon=0.1$. Colors identify $c$. Panels, line styles, markers, and error bars follow Figure~\ref{fig:flights-cvar-alpha}.}
    \label{fig:flights-dro-shape}
\end{figure}

In Corollary~\ref{cor:dro_l2}, the coefficient excluding the information factor scales as $n^{-(c-1)/c}$ for $1<c<2$ and as $n^{-1/2}$ for $c\ge2$.
In Figure~\ref{fig:flights-dro-shape}, the bounds for $1<c<2$ are larger and decrease more slowly as $c$ approaches one.
For $c=1.25$ and $\sigma=0.05$, the bound remains above one even at the largest sample size.
The cases $c=2,4$ share the same exponent, and their difference comes from the coefficient $\{1+c(c-1)\epsilon\}^{1/c}$; here, the bound at $c=4$ is slightly larger than that at $c=2$.
The sample-size dependence of the estimated information also contributes to that of the bound.

In contrast, the estimated gap curves remain close as $c$ varies, with differences small relative to their sample standard deviations.
In this setting, changing the divergence shape has a more pronounced effect on the bound's coefficient than on the observed gap.

\begin{figure}[H]
    \centering
    \includegraphics[width=\linewidth]{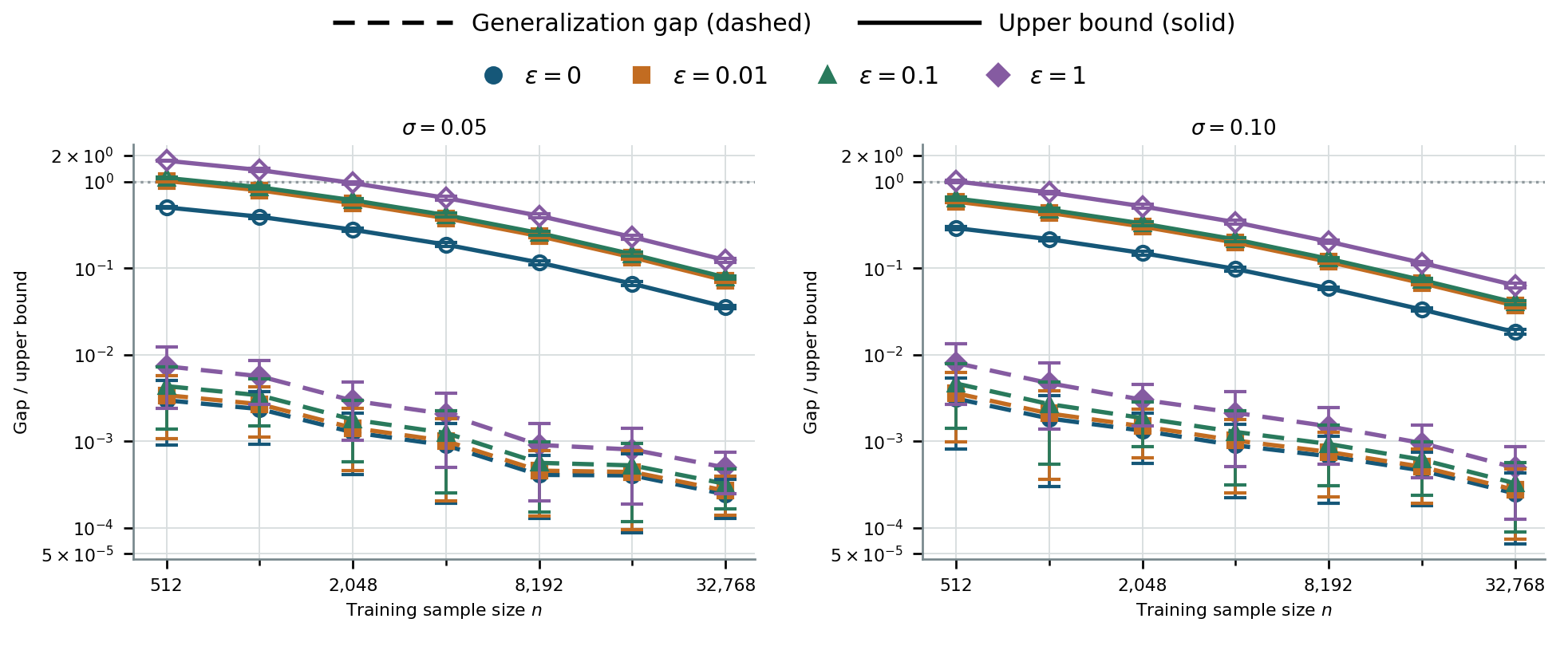}
    \caption{CR DRO radius comparison at $c=2$. Positive-radius bounds use Eq.~\eqref{eq:dro_vt_bound_main} in Corollary~\ref{cor:dro_l2}; the bound at $\epsilon=0$ uses the full-selector information term of Theorem~\ref{thm:wang_general} with $\beta_n=1/n$. Panels and plotting conventions follow Figure~\ref{fig:flights-cvar-alpha}.}
    \label{fig:flights-dro-radius}
\end{figure}

Increasing $\epsilon$ enlarges the feasible reweighting set, allowing larger loss values to receive greater weight in both training and validation evaluation.
In Figure~\ref{fig:flights-dro-radius}, the mean gap estimate increases with the radius.
For positive radii, the bound increases through $\kappa_{2,\epsilon}=\sqrt{1+2\epsilon}$ while retaining the same explicit exponent of $n$.
The empirical mean at zero radius uses the coefficient from Theorem~\ref{thm:wang_general}, so the difference between the zero- and positive-radius bounds also includes a change in the bound being used.
As for CVaR, the upper bounds are smaller at $\sigma=0.10$ than at $\sigma=0.05$, while the gaps are larger in many settings.

\section{CelebA/ViT: settings and estimation}
\label{app:celeba-settings}

This appendix describes the data, retraining protocol, information estimation,
and aggregation used in Figure~\ref{fig:bound}.

\subsection{Data, model, and evaluation objectives}

\paragraph{Dataset and groups.}
We predict the \texttt{Blond\_Hair} attribute in \texttt{CelebA}, with
\(Y=1\) denoting blond hair and \(Y=0\) denoting its absence.
The multicalibration group class contains 18 overlapping groups:
\texttt{Male}, \texttt{Female}, \texttt{Arched\_\allowbreak Eyebrows},
\texttt{Bangs}, \texttt{Big\_\allowbreak Lips}, \texttt{Chubby},
\texttt{Double\_\allowbreak Chin}, \texttt{Eyeglasses},
\texttt{High\_\allowbreak Cheekbones},
\texttt{Mouth\_\allowbreak Slightly\_\allowbreak Open},
\texttt{Oval\_\allowbreak Face}, \texttt{Pale\_\allowbreak Skin},
\texttt{Receding\_\allowbreak Hairline}, \texttt{Smiling},
\texttt{Straight\_\allowbreak Hair}, \texttt{Wavy\_\allowbreak Hair},
\texttt{Wearing\_\allowbreak Hat}, and \texttt{Young}.
The first two groups select examples with \texttt{Male} equal to 1 and 0,
respectively; each remaining group selects examples whose named attribute is 1.
These are single-attribute groups.
The dataset wrapper removes groups whose sizes are at most \(0.005\) of the
full dataset; all 18 groups are retained in this experiment.
The group collection is fixed throughout evaluation, with group indices
reindexed within each evaluation sample.

\paragraph{Model and training settings.}
We use a binary ViT classifier with images resized to \(224\times224\).
We initialize it from the pretrained
\texttt{google/vit-base-patch16-224-in21k} checkpoint.
Each classifier is trained for 10 epochs with cross-entropy loss and Adam,
using batch size 64, learning rate \(10^{-4}\), and weight decay \(0.01\).
We use the final-epoch checkpoint.

\paragraph{Evaluated objectives.}
We evaluate multicalibration (MC) and CVaR.
For MC, positive-class probabilities \(\hat p_i\) are converted into
two-class probabilities \((1-\hat p_i,\hat p_i)\), with one-hot labels.
Each class probability is discretized into \(B=10\) equal-width bins on
\([0,1]\), with a probability of 1 included in the last bin.
The evaluation routine forms a joint bin index
\(r=(r_0,r_1)\in\{0,\ldots,B-1\}^2\) from the two discretized probabilities
and computes calibration errors for joint-bin--group--class triples.
The reported statistic weights each occupied cell's absolute conditional
calibration error by its empirical cell fraction:
\[
\widehat\Phi_{\mathrm{MC}}(f;A)
=
\max_{\substack{r\in\{0,\ldots,B-1\}^2,\,g\in\mathcal G,\,k\in\{0,1\}\\
                 \text{cell }(r,g)\text{ occupied}}}
\frac{N_{r,g}(A)}{|A|}
\left|\widehat{\mathrm{calerr}}_{r,g,k}(f;A)\right|,
\]
where \(N_{r,g}(A)\) is the number of examples in the corresponding joint-bin--group
cell of evaluation sample \(A\).
The signed conditional error \(\widehat{\mathrm{calerr}}_{r,g,k}\) is
the mean of \(\mathbf 1\{y_i=k\}-p_{i,k}\) within that cell, where
\(p_{i,0}=1-\hat p_i\) and \(p_{i,1}=\hat p_i\).
The implementation also records an unweighted worst-cell diagnostic;
Figure~\ref{fig:bound} uses the moment-weighted statistic above.
For CVaR, we apply the empirical objective in
Eq.~\eqref{eq:cvar_empirical} to the absolute calibration losses
\(\ell_i=|\hat p_i-y_i|\), with \(\alpha=0.2\) and
\(\bar\alpha=1-\alpha\).

\paragraph{Implementation.}
We build on the public empirical multicalibration codebase
of~\citet{hansen2024multicalibration}, released under the MIT license.
We extend it with paired supersamples, branch retraining, mixed-path
evaluation, local increments, and mutual-information estimation for the
two objectives.

\subsection{Supersample construction and branch retraining}

\paragraph{Candidate pool and sample sizes.}
We first split the dataset into training, validation, and test parts with
fractions \(0.4\), \(0.4\), and \(0.2\), respectively.
The test split is held out from retraining and local information estimation.
The union of the training and validation indices forms the supersample
candidate pool, whose size \(N_{\max}\) is approximately \(80\%\) of the
full dataset, up to integer rounding.
For each sample-size percentage \(p\), we select \(p\%\) of this pool,
rounding down to an even total \(N\), with a minimum total of 20.

\begin{samepage}
\noindent The reported sample sizes are:
\begin{center}
\small
\begin{tabular}{lrrrrrr}
\toprule
\(p\) (\%) & 5 & 10 & 25 & 50 & 75 & 100 \\
\(N\) & 8104 & 16208 & 40520 & 81040 & 121560 & 162080 \\
\(n=N/2\) & 4052 & 8104 & 20260 & 40520 & 60780 & 81040 \\
\bottomrule
\end{tabular}
\end{center}
\end{samepage}
Each branch training sample and its complementary validation sample contain
\(n=N/2\) examples.
The horizontal axis of Figure~\ref{fig:bound} uses the total supersample
size \(N=2n\).

\paragraph{Supersamples and selected pairs.}
For each sample size and both objectives, we use \(J=3\) supersample
configurations, generated with seeds \(10\), \(15\), and \(20\).
Let \(j\in[J]\) index these configurations.
For each configuration \(j\), we sample \(N\) candidate indices without
replacement and randomly permute them into an \(n\times2\) paired array
\(\widetilde Z^{(j)}\).
A selector vector \(U^{(j)}\), with independent uniform binary components,
chooses one example from each pair for training and the other for validation.
We uniformly sample a subset \(T_j\subset[n]\) of \(m=10\) pair indices
for local information estimation.

\paragraph{Two branches and seven training seeds.}
For each selected pair \(t\in T_j\), we hold \(U_{-t}^{(j)}\) fixed and
consider both choices \(b\in\{0,1\}\) for \(U_t^{(j)}\).
Write \(U^{(j;t,b)}\) for the resulting selector vector.
Its branch training and validation samples are
\[
S_{\mathrm{tr}}^{(j)}(t,b)
=
\bigl(\widetilde Z^{(j)}_{i,U_i^{(j;t,b)}}\bigr)_{i=1}^n,
\qquad
S_{\mathrm{val}}^{(j)}(t,b)
=
\bigl(\widetilde Z^{(j)}_{i,1-U_i^{(j;t,b)}}\bigr)_{i=1}^n.
\]
For each branch, we train a fresh classifier using each of the \(R=7\)
model seeds \(s\in\{0,1,\ldots,6\}\), which initialize the Python, NumPy,
and PyTorch random states.
Denote the resulting predictor by \(f_{j,t,b,s}\).
The complementary branch sample is not used for checkpoint selection.
Thus, each selected pair yields \(2R=14\) branch-training runs, and the
ten selected pairs give \(m\times2\times R=140\) runs per configuration
and sample size.
The 14 predictors for each pair produce the 14 increment--selector
observations used in its information estimate, as described next.

\subsection{Local increments and nearest-neighbor information estimation}

\paragraph{Mixed-path evaluation.}
For each selected pair \(t\), define the two mixed-path samples
\[
S_{\mathrm{mix}}^{(j)}(t,a)
=
\bigl(
\widetilde Z^{(j)}_{1,U_1^{(j)}},\ldots,
\widetilde Z^{(j)}_{t-1,U_{t-1}^{(j)}},
\widetilde Z^{(j)}_{t,a},
\widetilde Z^{(j)}_{t+1,1-U_{t+1}^{(j)}},\ldots,
\widetilde Z^{(j)}_{n,1-U_n^{(j)}}
\bigr),
\quad a\in\{0,1\}.
\]
These are the empirical counterparts of \(S(t,a)\) in
Theorem~\ref{thm:wang_general}.
For each branch-trained predictor, we compute the signed increment
\[
\widehat D_{j,t,b,s}
=
\widehat\Phi\bigl(f_{j,t,b,s};S_{\mathrm{mix}}^{(j)}(t,0)\bigr)
-
\widehat\Phi\bigl(f_{j,t,b,s};S_{\mathrm{mix}}^{(j)}(t,1)\bigr).
\]
The predictor is fixed across these two evaluations, while the auxiliary
optimization defining \(\widehat\Phi\) is performed separately on each
evaluation sample.
The mixed-path samples used for this increment differ from the branch
training and validation samples used to measure the observed gap.

\paragraph{Information estimate from 14 observations.}
For each fixed configuration--pair context \((j,t)\), we use
\[
\bigl\{(\widehat D_{j,t,b,s},b):
       b\in\{0,1\},\ s\in\{0,1,\ldots,6\}\bigr\}
\]
to estimate the mutual information between the continuous signed increment
and the discrete selector.
There are 14 increment--selector observations: seven from each selector value.
The supersample and \(U_{-t}^{(j)}\) remain fixed within this estimate;
the seven model seeds vary the learner's randomness.
Following the nearest-neighbor approach
of~\citet{futami2024information}, we use a continuous--discrete
\(k\)-nearest-neighbor MI estimator with \(k=1\).
The implementation uses scikit-learn's \texttt{mutual\_info\_classif}
with a continuous increment feature, a discrete selector label, and
\texttt{random\_state=0}.
If the available samples are insufficient to estimate dependence, for
example if only one selector value is present, the implementation returns zero.
MI estimates are truncated below at zero before taking any square root.
We denote this nonnegative context-specific estimate by
\(\widehat I_{j,t}^{\mathrm{LOO}}\), reflecting the leave-one-out
conditioning discussed in Appendix~\ref{app:conditioning_information_relations}.
No absolute value is applied to \(\widehat D_{j,t,b,s}\) before MI estimation.
At each sample size, each objective therefore has \(Jm=30\)
context-specific MI estimates, each based on 14 observations.

\subsection{Aggregation of the quantities in Figure~\ref{fig:bound}}

\paragraph{Observed absolute validation--training gap.}
For each trained predictor, define the signed branch-level gap
\[
g_{j,t,b,s}
=
\widehat\Phi\bigl(f_{j,t,b,s};S_{\mathrm{val}}^{(j)}(t,b)\bigr)
-
\widehat\Phi\bigl(f_{j,t,b,s};S_{\mathrm{tr}}^{(j)}(t,b)\bigr).
\]
We first take the absolute value of each branch-level gap and then average
over model seeds, both selector values, and the ten selected pairs:
\[
G_{\Phi,j}
=
\frac{1}{2Rm}
\sum_{t\in T_j}\sum_{b=0}^1\sum_{s=0}^{R-1}|g_{j,t,b,s}|,
\qquad
\widehat G_\Phi
=
\frac{1}{J}\sum_{j=1}^J G_{\Phi,j}.
\]
The gap curve plots \(\widehat G_\Phi\).
Thus, the absolute value precedes all averaging.
Theorem~\ref{thm:wang_general} instead concerns the absolute value of the
expected signed gap; the plotted mean absolute gap is a diagnostic summary.

\paragraph{Local-information proxy.}
For each context, we first transform its MI estimate into
\(\beta_\Phi\sqrt{2\widehat I_{j,t}^{\mathrm{LOO}}}\), where
\(\beta_\Phi=2\) for MC and
\(\beta_\Phi=\bar\alpha^{-1}=1.25\) for CVaR with \(\alpha=0.2\).
We then average over the ten selected pairs within each configuration,
followed by an average over configurations:
\[
P_{\Phi,j}
=
\frac{\beta_\Phi}{m}
\sum_{t\in T_j}\sqrt{2\widehat I_{j,t}^{\mathrm{LOO}}},
\qquad
\widehat{\mathcal B}_\Phi
=
\frac{1}{J}\sum_{j=1}^J P_{\Phi,j}.
\]
The proxy curve plots \(\widehat{\mathcal B}_\Phi\).
The square root is therefore taken separately for each context before
either average.

\paragraph{Shaded bands and interpretation.}
Only the proxy curve has a shaded band.
At each sample size, its width is determined by one standard deviation
across the three supersample-level proxies, computed with divisor \(J\):
\[
s_\Phi
=
\left\{\frac{1}{J}\sum_{j=1}^J
\bigl(P_{\Phi,j}-\widehat{\mathcal B}_\Phi\bigr)^2\right\}^{1/2},
\qquad J=3.
\]
The lower and upper band endpoints are
\(\max\{10^{-12},\widehat{\mathcal B}_\Phi-s_\Phi\}\) and
\(\widehat{\mathcal B}_\Phi+s_\Phi\), respectively.
The mean and standard deviation are computed on the original scale before
displaying the curves on a logarithmic vertical axis.
The seven model seeds are used within each MI estimate; the band uses
supersample configurations as its replication units.
As discussed in Appendix~\ref{app:conditioning_information_relations},
population information comparisons do not certify this average of
context-specific square roots as an upper bound.
Finite-sample nearest-neighbor MI estimates can also be biased.
We therefore interpret the reported information quantity as a diagnostic proxy.

\end{document}